\pdfoutput=1
\documentclass[10pt]{article}
\usepackage[margin=0.9in]{geometry}
\usepackage{natbib}
\usepackage{titlesec}
\titlespacing*{\section}{0pt}{1.1ex plus .2ex minus .2ex}{0.7ex plus .1ex}
\titlespacing*{\subsection}{0pt}{0.9ex plus .2ex minus .2ex}{0.5ex plus .1ex}
\titlespacing*{\paragraph}{0pt}{0.5ex plus .1ex}{1em}
\usepackage[utf8]{inputenc}
\usepackage[T1]{fontenc}
\usepackage[hidelinks]{hyperref}
\usepackage{url}
\usepackage{booktabs}
\usepackage{float}
\usepackage{amsfonts}
\usepackage{amsmath}
\usepackage{amssymb}
\usepackage{nicefrac}
\usepackage{microtype}
\usepackage{graphicx}
\usepackage{multirow}
\usepackage{xcolor}
\ifdefined\TEMPLATESPACING
\makeatletter
\newcounter{subfigure}[figure]
\renewcommand{\thesubfigure}{\alph{subfigure}}
\renewcommand{\p@subfigure}{\thefigure}
\newenvironment{subfigure}[1]
  {\begin{minipage}[t]{#1}\centering
   \renewcommand{\caption}[1]{\par\refstepcounter{subfigure}%
     {\small(\thesubfigure)~##1}\par}}
  {\end{minipage}}
\newcommand{\captionof}[1]{\def\@captype{#1}\caption}
\makeatother
\else
\usepackage{subcaption}
\fi
\usepackage{tikz}
\usepackage{pgfplots}
\pgfplotsset{compat=1.17}
\usetikzlibrary{positioning,decorations.pathreplacing,calc,fit,arrows.meta,backgrounds}

\newif\ifartifactlink\artifactlinkfalse

\newif\ifanon\anonfalse
\newcommand{\anonauthorblock}{Anonymous Authors\\ Anonymous Institution\\
  \texttt{anon.email@domain.com}}
\newcommand{\applyanon}{%
  \ifanon
    \hypersetup{pdfauthor={Anonymous Submission}, pdftitle={Anonymous Submission}}%
  \fi}

\newif\ifwipnote\wipnotefalse

\makeatletter
\newif\ifwidefloats
\if@twocolumn\widefloatstrue\fi
\ifdefined\FORCEWIDEFLOATS\widefloatstrue\fi
\newenvironment{widefigure}
  {\ifwidefloats\begin{figure*}[t]\else\begin{figure}[tbp]\fi}
  {\ifwidefloats\end{figure*}\else\end{figure}\fi}
\newenvironment{widetable}
  {\ifwidefloats\begin{table*}[t]\else\begin{table}[tbp]\fi}
  {\ifwidefloats\end{table*}\else\end{table}\fi}
\makeatother

\graphicspath{{figs/}}
\providecommand{\arxivgpu}[1]{}
\providecommand{\gptosshead}{GPT-OSS-20B}

\newcommand{\Titer}{T_{\mathrm{iter}}}
\newcommand{\tdraft}{t_{\mathrm{draft}}}
\newcommand{\tverify}{t_{\mathrm{verify}}}

\newcommand{\AL}{\mathrm{AL}}
\newcommand{\TPOT}{\mathrm{TPOT}}
\newcommand{\dt}{\ensuremath{d_{\mathrm{t}}}}  %
\newcommand{\dd}{\ensuremath{d_{\mathrm{d}}}}  %
\newcommand{\vocab}{V}
\newcommand{\blk}{k}
\newcommand{\hd}{h_{\mathrm{d}}}               %
\newcommand{\htgt}{h_{\mathrm{t}}}             %
\newcommand{\hup}{h_{\mathrm{up}}}             %
\newcommand{\Wout}{W_{\mathrm{out}}}           %
\newcommand{\Wlm}{W_{\mathrm{lm}}}             %
\newcommand{\Wfold}{W_{\mathrm{fold}}}         %

\definecolor{armexact}{HTML}{4C566A}
\definecolor{armhalf}{HTML}{2E7DBE}
\definecolor{armqtr}{HTML}{D1793C}
\definecolor{accentgreen}{HTML}{4C8C4A}
\definecolor{accentgrey}{HTML}{9AA0A6}

\ifdefined\TEMPLATESPACING\else
\usepackage[font=small,labelfont=bf,skip=4pt]{caption}
\fi

\ifdefined\TEMPLATESPACING\else
\fi

\usepackage{placeins}

\newcommand{\sharedtablehere}[1]{\begingroup
  \renewenvironment{widetable}{\begin{table}[H]}{\end{table}}%
  #1\endgroup}

\newcommand{\headlinetitle}{Main result on B200 at temperature 0.}
\newcommand{\headlinedflash}{}
\newcommand{\headlineeagle}{}
\renewcommand{\arxivgpu}[1]{#1}
\newcommand{\familycaption}{\textbf{The two drafter families.} Both narrow the
whole residual stream and differ in how it reaches the vocabulary. DFlash uses
the target's head and folds the output adapter into it. EAGLE-3 trains its own
head over a pruned vocabulary. Pairs of values are Qwen3-8B / GPT-OSS-20B.}
\renewcommand{\gptosshead}{GPT-OSS}

\anonfalse

\title{NarrowSpec: Draft Hidden Width Need Not Match\\
the Target in Speculative Decoding}
\ifanon
  \author{\anonauthorblock}
\else
  \author{Alagappan Valliappan$^{\dagger}$ \quad Ankit More \quad Dheevatsa Mudigere
    \quad Jatin Mitra\\ NVIDIA}
\fi
\date{}

\begin{document}
\applyanon
\ifanon\else
  \hypersetup{pdftitle={NarrowSpec: Draft Hidden Width Need Not Match the Target in Speculative Decoding},
    pdfauthor={Alagappan Valliappan, Ankit More, Dheevatsa Mudigere, Jatin Mitra}}
\fi

\maketitle
\ifanon\else
  {\renewcommand{\thefootnote}{}%
   \footnotetext{$^{\dagger}$Corresponding author: \texttt{avalliappan@nvidia.com}}}
\fi

\begin{abstract}
Speculative decoding is widely used to speed up generation in large language
models (LLMs), with a small \emph{draft} model proposing tokens that one forward
pass of the \emph{target} model verifies. Draft models that read the target's
hidden states, such as DFlash and EAGLE-3, typically inherit the target's hidden
width. We show that the draft's whole residual stream can run at half or a
quarter of that width while largely preserving acceptance. For DFlash, which
borrows the target's vocabulary projection, the trained output adapter
\emph{folds} into it, giving one narrower projection with logits unchanged up to
floating-point rounding. We test two drafter families,
\textbf{DFlash} and \textbf{EAGLE-3}, on \textbf{two targets}, Qwen3-8B (NVFP4)
and GPT-OSS-20B (MXFP4), in SGLang on a \textbf{B200} and a
\textbf{DGX Spark (GB10)}. At batch size 1 and temperature 0, narrowing the
draft to half or a quarter width speeds up end-to-end decoding by
\textbf{1.06--1.12$\times$} on B200 and \textbf{1.13--1.44$\times$} on GB10.
The speedup holds at larger batch sizes. From batch 32 to the largest batch
each GPU fits, narrowing reaches up to \textbf{1.21$\times$} on both GPUs, and
no narrow draft is slower than full width.
\end{abstract}
\section{Introduction}
\label{sec:intro}

Speculative decoding (SD)~\citep{leviathan2023speculative,chen2023accelerating}
cuts the per-token latency of autoregressive generation. A small, fast draft
proposes candidate tokens, and the target verifies them all in one batched
forward pass and accepts the longest correct prefix. Let $\AL$ be the acceptance length, the
mean number of tokens committed per verify step, and let $\Titer$ be the cost
of one speculative iteration. The time per output token, a decode-only metric
that excludes prefill, is then
\begin{equation}
\TPOT \;=\; \frac{\Titer}{\AL}, \qquad
\Titer \;\approx\; \tverify + \tdraft .
\label{eq:tpot}
\end{equation}
Equation~\ref{eq:tpot} gives two ways to reduce the time per output token,
raising $\AL$ or lowering $\Titer$. Much of the literature has focused on
raising $\AL$, with drafts conditioned on the
target's hidden states~\citep{li2024eagle,li2024eagle2,li2025eagle3}, extra
prediction heads~\citep{cai2024medusa,gloeckle2024better,deepseekv3} or trees
of candidates~\citep{miao2024specinfer,chen2024sequoia}. While effective, these
techniques add work on the draft side, the verify side, or both. Alternatively, a
growing body of work lowers $\Titer$. It shortens what the draft attends to, drafts with
fewer layers, trims the vocabulary projection, makes verification sparser, or
stops drafting earlier (\S\ref{sec:related}). Each of these techniques optimizes
one component or one decision and leaves the rest of the iteration as is.

\textbf{Often overlooked in prior work is the draft's residual hidden width,
which is usually set equal to the target's hidden width.} \emph{Feature-conditioned} drafters, such as
DFlash, EAGLE-3 and MTP drafters, read hidden states tapped from the target and
typically run their entire residual stream at the target's width. We show that the
draft can run at half or even a quarter of this width while largely preserving
acceptance.

\paragraph{Narrowing the whole draft.}
Every block of the draft reads and writes its residual stream, so narrowing it shrinks every projection in the draft
at once. The narrow draft is trained from scratch at its own width, with
adapters wherever it meets target-width tensors and matching support in the
serving engine (\S\ref{sec:method}). A drafter that borrows the target's
vocabulary projection, as DFlash does, trains an output adapter back to
target width. At inference we \emph{fold} the two, once, into a single
narrower projection whose logits are unchanged up to floating-point rounding.
Prior work compresses only the vocabulary projection~\citep{slimspec2026} or
fixes a drafter's width without comparing alternatives~\citep{gemma4}
(\S\ref{sec:related}). We make width the object of study. For drafters
trained on a frozen target, we measure how much width can be removed, what
that costs in $\AL$ and what it buys end to end, for a block-parallel drafter
(DFlash) and an autoregressive one (EAGLE-3), on a dense and an MoE target.

\paragraph{What we find.}
For each draft family and target model, we train every width on identical
data and evaluate on the 880-prompt qualitative split of
SPEED-Bench~\citep{abramovich2026speedbench}, which spans 11 categories from
coding and math to RAG and roleplay. We serve the drafts on a B200 and repeat
the measurements on a DGX Spark (GB10) and, for DFlash on Qwen3-8B, an H100. Across both drafters and
both targets we find four things.

\begin{itemize}
\item \textbf{The draft's hidden width does not need to match the target's.} On
the 880-prompt panel, halving costs at most 3.0\% acceptance, and none in two of
the four drafter--target pairs, while quartering costs 2.1--5.4\%. At
temperature 0 and batch 1 on B200, halving gives \textbf{1.06--1.09$\times$} end to end
and quartering \textbf{1.06--1.12$\times$} (Table~\ref{tab:headline}).
\item \textbf{The gain is larger on lower-bandwidth GPUs.} On GB10 the four
pairs give \textbf{1.13--1.23$\times$} at half width and
\textbf{1.20--1.44$\times$} at quarter width, with acceptance costs within 1.8
points of B200's (Table~\ref{tab:gb10summary}). The quarter-width draft runs
1.43--1.75$\times$ faster than the full-width draft on B200, where its narrowed kernels
are too short to reach peak bandwidth, and 3.44--3.87$\times$ on GB10, where
they stay near it (\S\ref{sec:hw}). On an H100, which sits between the two in bandwidth, DFlash
on Qwen3-8B decodes \textbf{1.11$\times$} faster at both widths, against 1.09$\times$
and 1.06$\times$ on B200.
\item \textbf{The benefit persists across batch size and context length.} On
B200, from batch 1 to 512, half width decodes up to 1.16$\times$ faster end to
end and quarter width up to 1.21$\times$, and both stay at or above parity. On
GB10 the extra gain is largest at small batch, and from batch 32 to 256 it
narrows toward B200's. On B200, from 8k to 32k context, no $\Titer$ speedup
changes by more than 1.6\% (\S\ref{sec:scale}).
\item \textbf{The saving is spread across the draft, and folding lets a
borrowed head narrow exactly.} In three of the four pairs, the MLP and attention projections across the
draft's layers supply 50--60\% of the draft-body saving. Folding lets a narrow
DFlash draft read a head $2\times$ or $4\times$ smaller than the target's,
with the same logits up to rounding, and that head supplies 38--44\% of
DFlash's draft-body saving (\S\ref{sec:breakdown}).
\end{itemize}

\section{Method}
\label{sec:method}

Write $\dt$ and $\dd$ for the target's and the draft's hidden widths,
$\vocab$ for the vocabulary size, and $\htgt \in \mathbb{R}^{\dt}$ and
$\hd \in \mathbb{R}^{\dd}$ for their hidden states. A feature-conditioned
draft meets the target at two ends. Its input includes $\dt$-wide hidden
states tapped from the target, and after its layers it outputs a distribution
over the target's $\vocab$ tokens. The residual stream that runs through those
layers need not be $\dt$ wide. We set it to
$\dt/2$ or $\dt/4$. Every block of the draft shares the residual stream, so
narrowing it shrinks every projection that reads or writes it. That includes
the attention projections and MLP of every layer, the projection that brings
in the target's features, the projections that write the draft's KV cache,
and the projection to the vocabulary (Figure~\ref{fig:overview} in
App.~\ref{app:overview} expands one DFlash layer at full and half width). In both families the head
count, head dimension and MLP intermediate size do not change with width.
Because the head geometry is fixed, attention itself and the draft's KV cache
keep their size.

\subsection{The drafters we narrow}
\label{sec:drafters}

The two families of drafters we study differ in how they draft and in what they borrow
from the target (Table~\ref{tab:family}). \textbf{DFlash}~\citep{dflash2026}
is a block-parallel drafter (Figure~\ref{fig:path}). Its $\blk$ is the released drafters' block
size, 16 on Qwen3-8B and 8 on GPT-OSS-20B. Each iteration it drafts a block of
$\blk$ positions, the bonus token from the last verify followed by $\blk-1$
mask tokens, in one forward pass over five or eight stacked layers, so it
proposes $\blk-1$ tokens. It reads hidden
states from five target layers, which a context pre-pass projects into the
draft's KV cache once per iteration, and borrows the target's embedding and
vocabulary projection. On the other hand, \textbf{EAGLE-3}~\citep{li2025eagle3} is a
single decoder layer that runs autoregressively (Figure~\ref{fig:path-e3}). For EAGLE-3,
$\blk$ is the number of draft steps per iteration, each proposing one token, so
it proposes $\blk{=}7$ tokens. The first step reads the tokens the target just
accepted, including the bonus token, with their three tapped target layers
concatenated to $3\dt$. Only the bonus token's position continues the chain, and the
others only write the draft's KV cache. Steps 2--7 each draft one token per
request from the draft's own previous token and hidden state. At every step
the token's embedding and the hidden state are concatenated to $2\dd$ before
the decoder layer. EAGLE-3 borrows the target's token embedding and trains
its own vocabulary projection over a pruned vocabulary of 32{,}000 entries on
Qwen3-8B and 64{,}000 on GPT-OSS-20B, with a lookup table that maps each
draft token to its target id. Together, the two families cover parallel and
autoregressive drafting, and a borrowed and a self-trained vocabulary
projection, so the width result is tested on both sides of each choice.

\providecommand{\familycaption}{\textbf{Two families, one intervention.} Both narrow the whole residual
stream. They differ in how the narrow stream reaches the vocabulary. DFlash
borrows the target's full-vocabulary head and folds the output adapter into it.
EAGLE-3 trains its own head over a pruned vocabulary at draft width. Pairs
of values are Qwen3-8B / GPT-OSS-20B.}
\begin{widetable}
\centering
\small
\caption{\familycaption}
\label{tab:family}
\begin{tabular}{lll}
\toprule
 & \textbf{DFlash} & \textbf{EAGLE-3} \\
\midrule
Draft body        & 5 / 8 stacked layers & \textbf{one} layer, run $\blk{=}7$ times \\
Drafting          & block-parallel, one block per anchor  & autoregressive chain \\
LM head           & target's, $\vocab \times \dt$, borrowed frozen & \textbf{its own}, 32K / 64K $\times\,\dd$, trained \\
Head at narrow width & folded: $\Wfold = \Wlm\Wout$, $\vocab \times \dd$ & already at $\dd$, no head fold needed \\
Input adapter at serving & runtime GEMM, 4.2--7.2\,$\mu$s & \textbf{zero} --- folded into \texttt{embed\_tokens} \\
Output adapter    & $\dd \to \dt$, folded away at load & none \\
Target taps       & 5 layers & 3 layers \\
\bottomrule
\end{tabular}
\end{widetable}

\providecommand{\trapz}[4]{%
  \begin{scope}[on background layer]
    \path[#4] ($(#1.west)+(0,#2/2)$) -- ($(#1.east)+(0,#3/2)$)
      -- ($(#1.east)-(0,#3/2)$) -- ($(#1.west)-(0,#2/2)$) -- cycle;
  \end{scope}}
\begin{widefigure}
\centering
\resizebox{\textwidth}{!}{%
\begin{tikzpicture}[
  font=\small,
  box/.style={draw, rounded corners=2pt, align=center, inner sep=2.5pt, minimum height=10mm},
  tgt/.style={box, fill=armexact!12, draw=armexact, minimum height=14mm},
  nar/.style={box, fill=armhalf!14, draw=armhalf, thick},
  flat/.style={box, fill=accentgrey!18, draw=accentgrey!90!black},
  proj/.style={align=center, inner sep=1pt, minimum height=14mm, minimum width=19mm},
  narrow/.style={proj},
  widen/.style={proj},
  narp/.style={fill=armhalf!14, draw=armhalf, thick},
  gonep/.style={dashed, fill=white, draw=armqtr, text=armqtr, thick},
  tgtp/.style={dashed, fill=armexact!6, draw=armexact, thick},
  ar/.style={-{Latex[length=2mm]}, thick},
  dar/.style={-{Latex[length=2mm]}, dashed, thick, armqtr},
  lbl/.style={font=\footnotesize\itshape, text=black!65, align=left},
]

\node[tgt, minimum width=33mm] (verify) at (0,0)
  {\textbf{(1) Target verify}\\[1pt] one fwd over last accepted token\\$+$ $\blk\!-\!1$ proposals};
\node[tgt, right=14mm of verify] (taps)
  {5-layer hidden tap\\[1pt] $[\blk,\,5\dt]$};
\node[narrow, right=5mm of taps] (fc)
  {\texttt{fc}\\$5\dt\!\to\!\dd$};
\node[nar, right=5mm of fc] (kvproj)
  {context $K/V$ proj.\\[1pt] per draft layer};
\node[flat, right=5mm of kvproj] (kvcache)
  {draft KV cache\\[1pt] \emph{accepted prefix}};

\begin{scope}[on background layer]
  \node[draw=black!35, dashed, rounded corners=3pt, inner sep=4pt,
        fit=(taps)(fc)(kvproj)(kvcache)] (prepass) {};
\end{scope}
\node[lbl, below=0.5mm of prepass.south west, anchor=north west]
  {\textbf{(2) Context pre-pass}, once per iteration};

\node[tgt, minimum height=10mm, below=16mm of verify.south west, anchor=north west] (blockids)
  {input token ids\\[1pt] bonus $+$ $\blk\!-\!1$ masks};
\node[flat, minimum height=14mm, right=5mm of blockids] (embed)
  {target embedding\\[1pt] $[\vocab,\dt]$};
\node[narrow, right=5mm of embed] (inadapt)
  {input adapter\\$\dt\!\to\!\dd$};
\node[nar, right=5mm of inadapt] (layers)
  {$L$ draft layers\\[1pt] $+$ norm at $\dd$};
\node[widen, right=9mm of layers, minimum width=22mm] (head)
  {folded head\\$\dd\!\to\!\vocab$};

\begin{scope}[on background layer]
  \node[draw=black!35, dashed, rounded corners=3pt, inner sep=4pt,
        fit=(blockids)(embed)(inadapt)(layers)(head)] (draft) {};
\end{scope}
\node[lbl, below=0.5mm of draft.south west, anchor=north west]
  {\textbf{(3) Draft forward}, all $\blk$ positions in one pass};

\node[widen, text=armqtr, below=6mm of layers, font=\footnotesize] (outadapt)
  {output adapter\\$\dd\!\to\!\dt$};
\node[box, gonep, font=\footnotesize, minimum height=14mm] (lmhead)
  at (outadapt -| head) {target LM head\\$[\vocab,\dt]$};
\node[lbl, text=armqtr, anchor=east, align=right] at ($(outadapt.west)+(-3mm,-2mm)$)
  {output adapter maps $\dd$ back to $\dt$\\for the target's LM head};

\trapz{fc}{14mm}{7mm}{narp}
\trapz{inadapt}{14mm}{7mm}{narp}
\trapz{head}{7mm}{14mm}{narp}
\trapz{outadapt}{7mm}{14mm}{gonep}

\draw[ar] (verify) -- node[lbl, above] {hidden} (taps);
\draw[ar] (taps) -- (fc);
\draw[ar] (fc) -- (kvproj);
\draw[ar] (kvproj) -- (kvcache);
\draw[ar] (blockids.north |- verify.south) -- node[lbl, right] {bonus token} (blockids.north);
\draw[ar] (blockids) -- (embed);
\draw[ar] (embed) -- (inadapt);
\draw[ar] (inadapt) -- (layers);
\draw[ar] (layers) -- (head);
\draw[ar] (kvcache.south) |- node[lbl, left, pos=0.3] {context $K,V$} ([yshift=5mm]layers.north) -- (layers.north);
\draw[dar] (layers.south) -- (outadapt.north);
\draw[dar] (outadapt) -- (lmhead);
\coordinate (loopx) at ($(head.east -| kvcache.east)+(5mm,0)$);
\draw[dashed, thick, armqtr] (lmhead.east) -| (loopx);
\draw[ar] (head.east) -- (loopx) |- node[lbl, above, pos=0.75] {$\blk\!-\!1$ proposals, verified next iteration} ([yshift=6mm]verify.north) -- (verify.north);

\end{tikzpicture}}
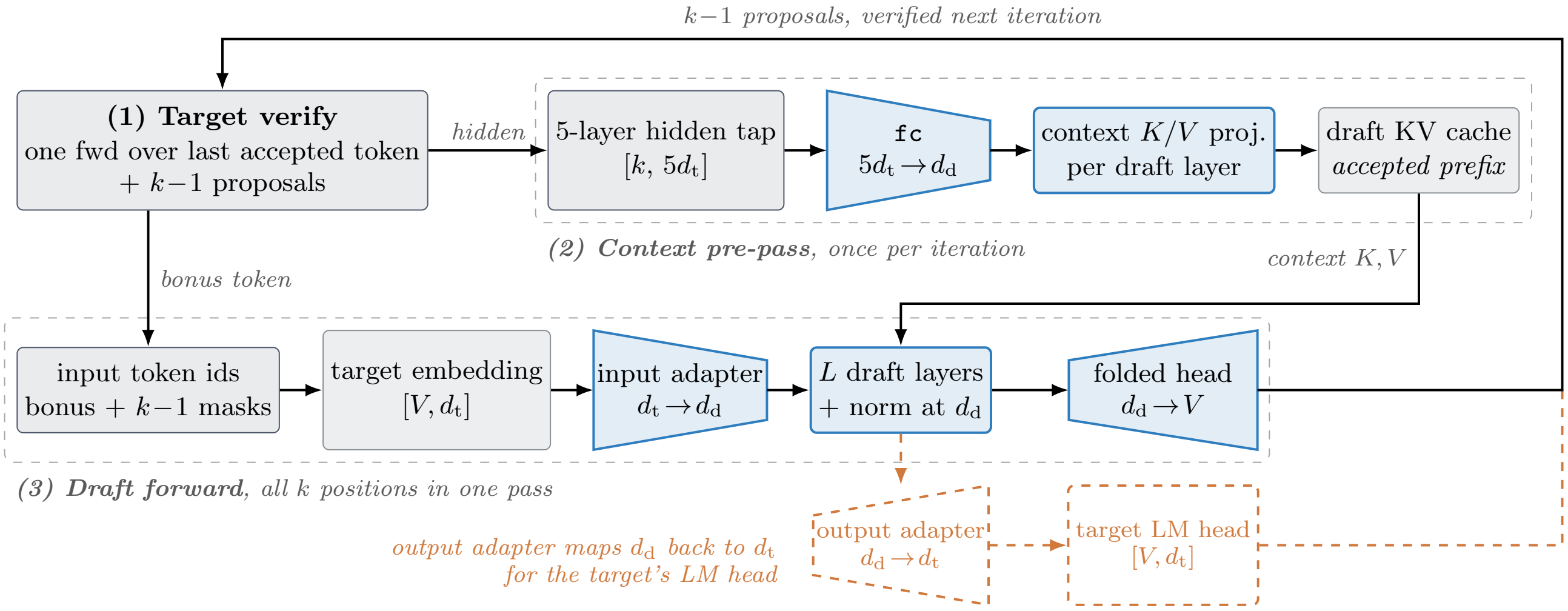
\caption{\textbf{One DFlash decode iteration and where draft width acts.} Blue
parts run at draft width $\dd$, grey parts at target width $\dt$ or independent
of width, and trapezoids are projections that change width. The draft has $L{=}5$ layers on Qwen3-8B and 8 on GPT-OSS-20B. Dashed orange marks
the training-only path. Training routes the draft through an output adapter to
the target's frozen LM head. At load for inference, the two are folded into one
narrow head.}
\label{fig:path}
\end{widefigure}

\subsection{Training a narrow draft}
\label{sec:train-narrow}

The narrow draft is trained from scratch at its own width, with an adapter
wherever it meets a target-width tensor (the trapezoids in
Figures~\ref{fig:path} and~\ref{fig:path-e3}). In both families the concat
projection \texttt{fc} maps the tapped target hidden states $\htgt$, from
$5\dt$ in DFlash and $3\dt$ in EAGLE-3, down to $\dd$ before anything else runs, so
every decoder layer reads a narrow input. DFlash's context $K/V$ projection
reads \texttt{fc}'s output and so narrows with no adapter of its own
(\S\ref{sec:projections}). Token embeddings come from the target's
$\vocab \times \dt$ table and pass through a $\dt \to \dd$ input adapter.

The families differ at the output. DFlash scores with the target's vocabulary
projection $\Wlm \in \mathbb{R}^{\vocab \times \dt}$, which expects a
$\dt$-wide input. Training therefore adds an output adapter
$\Wout \in \mathbb{R}^{\dt \times \dd}$ that maps the narrow state back up,
$\hup = \Wout\hd$, and computes logits $\Wlm\hup$ (the dashed path in
Figure~\ref{fig:path}) using the target's frozen LM head. EAGLE-3 trains its own vocabulary
projection at $\dd$, so it narrows with the rest of the draft and needs no output
adapter.

\subsection{Serving a narrow draft}
\label{sec:serve-narrow}

\begin{figure}[t]
\centering
\begin{tikzpicture}[
  font=\small,
  op/.style={draw, rounded corners=2pt, align=center, inner sep=4pt, minimum height=8mm},
  keep/.style={op, fill=armhalf!14, draw=armhalf, thick},
  drop/.style={op, fill=white, draw=armqtr, dashed, text=armqtr},
  tgt/.style={op, fill=accentgrey!18, draw=accentgrey!90!black},
  ar/.style={-{Latex[length=2mm]}, thick},
  lbl/.style={font=\scriptsize\itshape, text=black!65},
]
\node[lbl, anchor=west] (l1) at (0,0.95) {naive: reintroduces the full-width head};
\node[keep] (h1) at (0.6,0) {$\hd\,[\blk,\dd]$};
\node[drop, right=5mm of h1] (a1) {$\Wout$\\$\dd\!\to\!\dt$};
\node[tgt,  right=5mm of a1]  (e1) {$\Wlm$\\$\dt\!\to\!\vocab$};
\node[op, right=5mm of e1, fill=none, draw=none] (o1) {logits};
\draw[ar] (h1) -- (a1); \draw[ar] (a1) -- (e1); \draw[ar] (e1) -- (o1);

\node[lbl, anchor=west] (l2) at (0,-1.75) {folded: one GEMM, formed once at load, exact};
\node[keep] (h2) at (0.6,-2.5) {$\hd\,[\blk,\dd]$};
\node[keep, right=5mm of h2] (e2) {$\Wfold = \Wlm\Wout$\\$\dd\!\to\!\vocab$};
\node[op, right=5mm of e2, fill=none, draw=none] (o2) {logits};
\draw[ar] (h2) -- (e2); \draw[ar] (e2) -- (o2);

\draw[-{Latex[length=2mm]}, armqtr, thick]
  ($(a1.south)!0.5!(e1.south)+(0,-3mm)$) -- ++(0,-0.62cm)
  node[midway, right, font=\scriptsize, text=armqtr] {\ compose at load: $\Wlm\Wout$};
\end{tikzpicture}
\caption{\textbf{LM-head folding.} The output adapter $\Wout$ and the target head
$\Wlm$ are both linear, so $\Wfold = \Wlm\Wout$ is computed once at load and
serving runs one GEMM from draft width to the vocabulary. The composition is
algebraically exact, and in finite precision the logits differ only by rounding
(App.~\ref{app:artifact}).}
\label{fig:folding}
\end{figure}
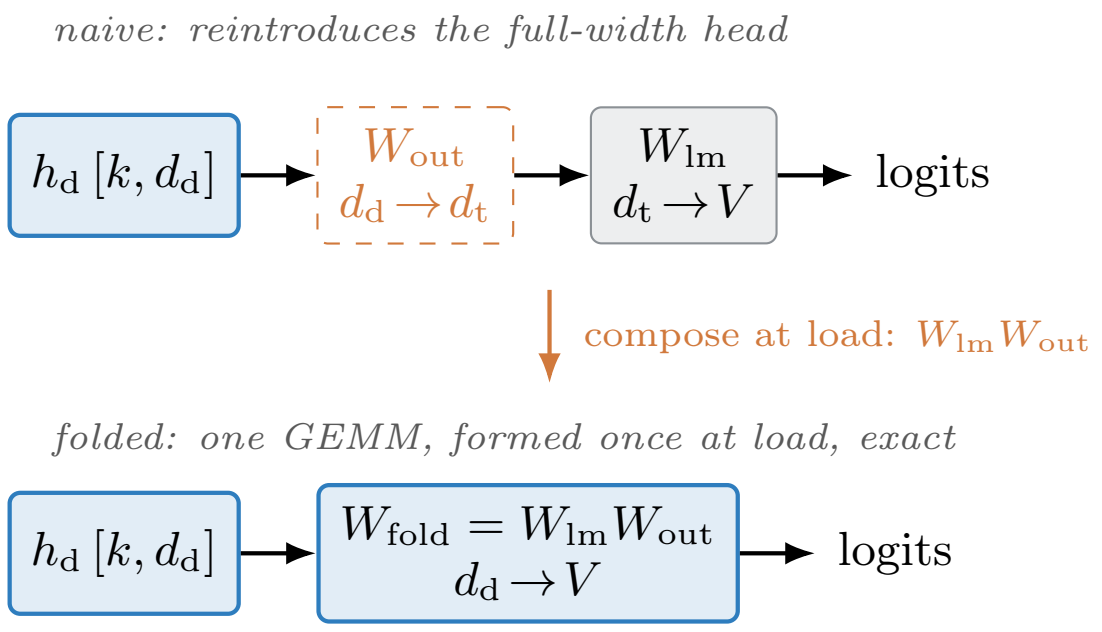

\paragraph{Folding the output adapter in DFlash.}
Served as trained, every DFlash iteration would run the output adapter $\Wout$
and then the target's full-width LM head $\Wlm$. Our narrow drafts run their final norm at $\dd$,
before $\Wout$, so nothing non-linear sits between the two maps, and at load we
combine them into one (Figure~\ref{fig:folding}),
\begin{equation}
\begin{aligned}
\Wlm(\Wout\hd) &\;=\; \Wfold\,\hd,\\
\Wfold &\;=\; \Wlm\Wout \;\in\; \mathbb{R}^{\vocab \times \dd},
\end{aligned}
\label{eq:fold}
\end{equation}
computed once. At serving time a single GEMM goes from draft width straight to
the vocabulary and the output adapter never runs. Both of our targets keep
their vocabulary projections in BF16. Folding changes the order of
floating-point operations and replaces the rounding of $\hup$ with a rounding
of $\Wfold$, so folded and unfolded logits need not be bitwise identical. On
our trained half-width Qwen3-8B adapter and real draft states, folded logits
are slightly closer to an FP64 reference than unfolded ones
(App.~\ref{app:artifact}). The width study uses BF16 heads.
The quantized drafts of App.~\ref{app:quant} are folded the same way, from
the BF16 head, and only then is $\Wfold$ quantized once to NVFP4. We do not
evaluate folding a target head that is already quantized. Since
$Q(\Wlm\Wout)\neq Q(\Wlm)\Wout$ in general, a quantized fold is not
guaranteed to match the unfolded quantized computation
(App.~\ref{app:foldprecision}). The folded $\vocab \times \dd$ head costs
resident memory alongside the target's own, but cuts the bytes the draft reads
per step. On Qwen3-8B the head GEMM and its argmax kernel together drop from
218.5 to 65.4\,$\mu$s at quarter width (Table~\ref{tab:l2}), and across both
targets it supplies 38--44\% of DFlash's draft-body saving
(App.~\ref{app:l3full}).

\providecommand{\trapz}[4]{%
  \begin{scope}[on background layer]
    \path[#4] ($(#1.west)+(0,#2/2)$) -- ($(#1.east)+(0,#3/2)$)
      -- ($(#1.east)-(0,#3/2)$) -- ($(#1.west)-(0,#2/2)$) -- cycle;
  \end{scope}}
\begin{widefigure}
\centering
\begin{tikzpicture}[
  font=\small,
  box/.style={draw, rounded corners=2pt, align=center, inner sep=2.5pt, minimum height=10mm},
  tgt/.style={box, fill=armexact!12, draw=armexact, minimum height=14mm},
  nar/.style={box, fill=armhalf!14, draw=armhalf, thick},
  flat/.style={box, fill=accentgrey!18, draw=accentgrey!90!black},
  proj/.style={align=center, inner sep=1pt, minimum height=14mm, minimum width=19mm},
  narrow/.style={proj},
  widen/.style={proj},
  narp/.style={fill=armhalf!14, draw=armhalf, thick},
  gonep/.style={dashed, fill=white, draw=armqtr, text=armqtr, thick},
  tgtd/.style={box, dashed, fill=armexact!6, draw=armexact, thick},
  ar/.style={-{Latex[length=2mm]}, thick},
  dar/.style={-{Latex[length=2mm]}, dashed, thick, armqtr},
  lbl/.style={font=\footnotesize\itshape, text=black!65, align=left},
]

\node[tgt, minimum width=33mm] (verify) at (0,0)
  {\textbf{(1) Target verify}\\[1pt] one fwd over last accepted token\\$+$ $\blk$ proposals};
\node[tgt, right=21mm of verify] (taps)
  {3-layer hidden tap\\[1pt] $3\dt$ per token};
\node[narrow, right=8mm of taps] (fc)
  {\texttt{fc}\\$3\dt\!\to\!\dd$};

\begin{scope}[on background layer]
  \node[draw=black!35, dashed, rounded corners=3pt, inner sep=4pt,
        fit=(taps)(fc)] (featproj) {};
\end{scope}
\node[lbl, above=0.5mm of featproj.north west, anchor=south west]
  {\textbf{(2) Feature projection}, once per iteration};

\node[flat, minimum width=12mm, below=25mm of verify.south west, anchor=north west] (tok)
  {input\\token ids};
\node[nar, right=12mm of tok] (embed)
  {embedding\\[1pt] $[\vocab,\dd]$};
\node[nar, right=14mm of embed] (cat)
  {concat\\[1pt] $2\dd$};
\node[nar, right=12mm of cat] (layer)
  {decoder layer\\[1pt] at $\dd$};
\node[widen, right=12mm of layer, minimum width=25mm] (head)
  {own head\\$\dd\!\to\!32\text{k}/64\text{k}$};
\node[flat, minimum height=6mm, below=3mm of layer, xshift=8mm] (kv)
  {draft KV cache};

\coordinate (loopy) at ([yshift=-2mm]kv.south);
\node[circle, draw=armhalf, fill=armhalf!14, thick, inner sep=1.6pt]
  (mux) at ([xshift=3mm, yshift=7mm]cat.north) {};
\coordinate (split) at ($(layer.east)!0.5!(head.west)$);
\begin{scope}[on background layer]
  \node[draw=black!35, dashed, rounded corners=3pt, inner sep=4pt,
        fit=(tok)(embed)(cat)(layer)(head)(kv)(loopy)(mux)] (chain) {};
\end{scope}
\node[lbl, below=0.5mm of chain.south west, anchor=north west]
  {\textbf{(3) Draft forward}, one layer run $\blk$ times in sequence};

\node[box, gonep, anchor=south west, font=\footnotesize, minimum height=14mm] (temb)
  at ([yshift=5mm]tok.north west) {target embedding\\$[\vocab,\dt]$};
\node[narrow, text=armqtr, right=3mm of temb, font=\footnotesize, minimum width=15mm] (embadapt)
  {adapter\\$\dt\!\to\!\dd$};
\node[lbl, text=armqtr, anchor=south west] at ([yshift=0.5mm]temb.west |- embadapt.north)
  {target embedding, then adapter down to $\dd$};

\trapz{fc}{14mm}{7mm}{narp}
\trapz{head}{7mm}{14mm}{narp}
\trapz{embadapt}{14mm}{7mm}{gonep}

\draw[ar] (verify) -- node[lbl, above] {hidden states} (taps);
\draw[ar] (taps) -- (fc);
\draw[ar] (fc.south) -- ([yshift=6mm]fc.south |- mux.north) -| node[lbl, below, pos=0.3] {tapped features (step 1)} (mux.north);
\draw[ar] (mux) -- ([xshift=3mm]cat.north);
\draw[ar] (verify.west) -- ++(-4mm,0) |- node[lbl, left, pos=0.25] {accepted\\tokens\\(step 1)} (tok.west);
\draw[ar] (tok) -- (embed);
\draw[ar] (embed) -- (cat);
\draw[ar] (cat) -- (layer);
\draw[ar] (layer) -- (head);
\fill (split) circle (1.2pt);
\draw[ar] (split) |- node[lbl, below, pos=0.75] {own hidden state (steps 2--$\blk$)} (mux.east);
\draw[ar, <->] ([xshift=8mm]layer.south) -- (kv.north);
\draw[ar] (head.south) |- (loopy) -| (tok.south);
\node[lbl, anchor=south] at (embed |- loopy) {own next token (steps 2--$\blk$)};
\draw[dar] (tok.north) -- (tok.north |- temb.south);
\draw[dar] (temb) -- (embadapt);
\draw[dar] (embadapt.east) -| ([xshift=-3mm]cat.north);
\draw[ar] (head.east) -- ++(5mm,0) |- node[lbl, above, pos=0.75] {$\blk$ proposals, verified next iteration} ([yshift=7.5mm]verify.north) -- (verify.north);

\end{tikzpicture}
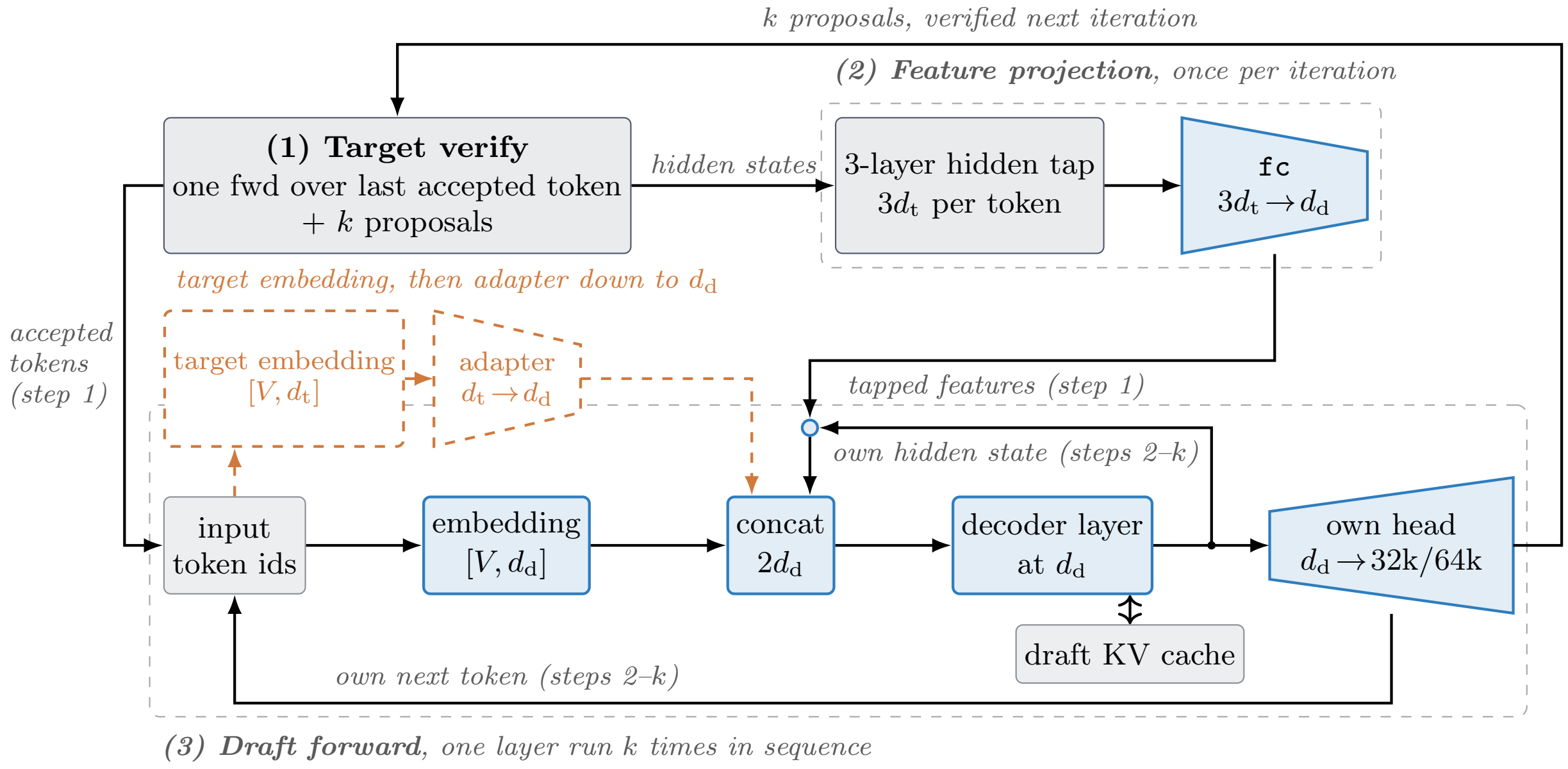
\caption{\textbf{One EAGLE-3 decode iteration and where draft width acts.} Colours
and shapes as in Figure~\ref{fig:path}. One decoder layer runs $\blk$ times. Step
1 reads the tokens just accepted, with their three tapped target layers through
\texttt{fc}. Later steps read the draft's own previous token and hidden state.
Training (dashed orange) embeds tokens with the target's table and a $\dt\!\to\!\dd$ adapter,
and export folds the two into the draft's own narrow table.}
\label{fig:path-e3}
\end{widefigure}

\paragraph{Folding the input adapter in EAGLE-3.}
EAGLE-3 multiplies its trained input adapter into the target's embedding
table at export, so the served draft owns a $\vocab \times \dd$ table and runs
no adapter GEMM. This fold is optional. The table takes about 0.6\,GB in BF16
at half width on either target, so a memory-constrained deployment can keep
the target's embedding instead and run the input adapter at each of the seven draft
steps. It uses the same $\dt \times \dd$ weight as DFlash's
4.2--7.2\,$\mu$s input adapter, but on steps 2--7 it embeds one token per
request rather than a block of $\blk$. DFlash could fold its input adapter
the same way, at the cost of its own $\vocab \times \dd$ table, but it runs
that adapter once per iteration rather than at seven sequential steps, so we
keep the target's embedding there and pay the 4.2--7.2\,$\mu$s.

\paragraph{Draft latency per iteration is fixed by width and $\blk$.}
Every iteration the draft proposes a fixed number of tokens, $\blk-1$ in DFlash
and $\blk$ in EAGLE-3, \textbf{not} the number the target accepts. DFlash drafts a full block every iteration, so its forward
pass processes $B\blk$ token positions ($B$ requests, $\blk$ positions each)
\emph{whatever the acceptance}. Acceptance changes how many tokens are added
to the draft's context, not the size of the block it drafts. EAGLE-3's first
draft forward pass (step 1 in Figure~\ref{fig:path-e3}) runs \texttt{fc} and
the decoder layer over the tokens just accepted, so in principle it covers
$\AL$ positions per request, but SGLang pads it to 8 under CUDA graphs, and
every iteration runs all seven passes. In both families the draft's latency per iteration therefore
depends, to first order, on width and $\blk$ and not on acceptance, which lets Eq.~\ref{eq:tpot} split
cleanly into a $\Titer$ set by width and an $\AL$ set by training.

\subsection{Where width enters the serving path}
\label{sec:projections}

Narrowing changes four paths in DFlash (Figure~\ref{fig:path}).
\textbf{The input adapter} projects the target-table embeddings of the anchor
and mask tokens from $\dt$ to $\dd$. It is the only added work, 4.2--7.2\,$\mu$s
per iteration against 212--351\,$\mu$s saved elsewhere
(Table~\ref{tab:l3full}), and the full-width draft has no input adapter.
\textbf{The concat projection \texttt{fc}} exists in every variant and maps the
five tapped target hidden states from $5\dt$ to $\dd$, so every later
operation reads a narrow input. \textbf{The output adapter} maps $\dd$ back to
$\dt$ in front of the target's head. Folding absorbs it at load, so it never
runs at serving time. \textbf{The context $K/V$ projections} read
\texttt{fc}'s output, so they narrow with no adapter of their own. They run once
per iteration and write the draft's KV cache, whose entries keep their size
because head count and head dimension do not change.
In EAGLE-3 width enters only at \texttt{fc} and the folded embedding table,
and its own head and every decoder projection run at $\dd$
(Figure~\ref{fig:path-e3}).

\section{Experimental setup}
\label{sec:setup}

\paragraph{Variants.} We train DFlash and EAGLE-3 drafts for two targets,
Qwen3-8B and GPT-OSS-20B, and call each drafter and target pair a
\emph{cell}, four in total. Each cell has three variants, \textbf{full width} (draft
width equal to target width, the baseline), \textbf{half} and
\textbf{quarter}. The narrow drafts set their residual width,
\texttt{hidden\_size}, to $\dt/2$ or $\dt/4$, which narrows every matrix that
reads or writes the residual stream. They also add the adapters of
\S\ref{sec:method}. Both drafters map the target's token embeddings to draft
width, which EAGLE-3 folds into its own embedding table, and DFlash adds an
output adapter that is folded into the borrowed head when the draft is loaded
for inference. Depth,
intermediate size, head geometry, vocabulary, data, order and seed stay fixed.

\begin{widetable}
\centering
\footnotesize
\setlength{\tabcolsep}{4pt}
\caption{Targets, variants and budgets. Both targets quantize the body and keep a BF16
\texttt{lm\_head}. Within a cell the narrow drafts change \texttt{hidden\_size}
and add width adapters (\S\ref{sec:method}). Data, training-token budget and
architecture settings other than residual width are matched, and learning rate
and, for EAGLE-3, initialization are selected per width (Table~\ref{tab:lr}). Trainable parameters include the adapters and the EAGLE-3 head,
but not the frozen target embedding or any fold. Each corpus regenerates every assistant response with the target at
temperature 0.}
\label{tab:setup}
\begin{tabular}{lll}
\toprule
 & \textbf{Qwen3-8B} & \textbf{GPT-OSS-20B} \\
\midrule
Body / head precision   & NVFP4 / \textbf{BF16 dense} & MXFP4 MoE / \textbf{BF16 dense} \\
Target hidden $\dt$     & 4{,}096 & 2{,}880 \\
Vocabulary $\vocab$     & 151{,}936 & 201{,}088 \\
Corpus samples / tokens per epoch & 799{,}936 / 587M & 788{,}544 / 1.04B \\
Generation cap          & 512 & 2{,}048 \\
\midrule
\multicolumn{3}{l}{\emph{DFlash draft} --- borrows the target's head, folds its output adapter at load}\\
Widths                  & 4{,}096 / 2{,}048 / 1{,}024 & 2{,}880 / 1{,}440 / 720 \\
Layers / intermediate   & 5 / 12{,}288 & 8 / 7{,}680 \\
Block size $\blk$       & 16 & 8 \\
Trainable parameters    & 1048.6M / 541.1M / 270.5M & 784.8M / 400.7M / 200.4M \\
Epochs; reported       & 6; ep.\ 5, 2.94B tokens & 6; ep.\ 6, 6.26B tokens \\
Global batch            & 16 & 64 \\
Reference draft         & \texttt{z-lab/Qwen3-8B-DFlash-b16} & \texttt{z-lab/gpt-oss-20b-DFlash} \\
\midrule
\multicolumn{3}{l}{\emph{EAGLE-3 draft} --- one layer $\times$ 7 steps, trains its own head at draft width}\\
Widths                  & 4{,}096 / 2{,}048 / 1{,}024 & 2{,}880 / 1{,}440 / 720 \\
Layers / intermediate   & \textbf{1} / 12{,}288 & \textbf{1} / 2{,}880 \\
Heads (Q / KV), \texttt{head\_dim} & 32 / 8, 128 & 64 / 8, 64 \\
$\blk$ (\texttt{ttt\_length}) & 7 & 7 \\
Draft vocabulary        & 32{,}000 of 151{,}936 & 64{,}000 of 201{,}088 \\
Trainable parameters    & 399.5M / 208.2M / 104.1M & 275.4M / 141.8M / 70.9M \\
Epochs; reported       & 3; ep.\ 3, 1.76B tokens & 3; ep.\ 3, 3.13B tokens \\
Global batch            & 64 & 64 \\
Reference draft         & \begin{tabular}[t]{@{}l@{}}\texttt{RedHatAI/}\\\texttt{Qwen3-8B-speculator.eagle3}\end{tabular} & \begin{tabular}[t]{@{}l@{}}\texttt{RedHatAI/}\\\texttt{gpt-oss-20b-speculator.eagle3}\end{tabular} \\
\bottomrule
\end{tabular}
\end{widetable}

\paragraph{Draft training data.}
Prompts come from Nemotron-Post-Training-v2 (English) and CodeAlpaca, but
we keep only the user turns. Every assistant response is regenerated
by the target itself at temperature 0, so the drafter learns the target's own
outputs. This gives about 800K training samples per target
(Table~\ref{tab:setup}). Evaluation
prompts are excluded from the training data, and all variants of a cell train on
the same processed dataset.

\paragraph{Training.} Every variant, including full width, is trained from
scratch with SpecForge~\citep{specforge2025} on four B200 GPUs, DFlash for six
epochs and EAGLE-3 for three, at the best of the learning rates we tested for
it. For EAGLE-3 that rate doubles with each halving of width, and its
initialization scale is selected per width too (App.~\ref{app:train}). Within a cell, every variant sees the same
data in the same order, for the same number of tokens and epochs, under the
same schedule shape. The variants are therefore compared at a matched data budget,
not at convergence, and the narrow drafts reach that budget with less training
compute. The budget is enough to train competitive drafters. The \emph{stock}
rows of Table~\ref{tab:headline} evaluate the released full-width
checkpoints on the same 880-prompt panel described below. At temperature 0 the
\texttt{z-lab} DFlash checkpoints are within 1.6\% of our DFlash full-width drafts'
acceptance length on both targets, and the \texttt{RedHatAI} EAGLE-3
checkpoints trail our EAGLE-3 full-width drafts by 20.6\% on Qwen3-8B and 7.9\% on
GPT-OSS-20B. At temperature 1 \texttt{z-lab} stays within 2.0\% of ours, while
\texttt{RedHatAI} pulls 1.4\% ahead on GPT-OSS-20B (Table~\ref{tab:stock}).
Table~\ref{tab:setup} lists the full budgets.

\paragraph{Inference and metrics.} We serve on one NVIDIA B200 (SM100, 148
SMs, 8\,TB/s HBM3e) with $\mathrm{TP}{=}1$, running SGLang~\citep{zheng2024sglang} 0.5.18
nightly with paged KV management~\citep{kwon2023vllm}, CUDA graphs and overlap
scheduling, and radix cache off. We serve the same drafts, targets and contracts
on one NVIDIA GB10, the GPU of the DGX Spark (SM121, 128\,GB of LPDDR5x shared
with the CPU, 273\,GB/s). GB10 keeps every B200 setting it can run. Only the
target attention changes, because TRT-LLM MHA runs only on SM100, and SGLang gets
less memory because the operating system shares it (Table~\ref{tab:gb10setup}).
Acceptance length is \textbf{pooled}, meaning total completion tokens divided
by total verify steps across the whole panel, so a 20-token reply counts for
less than a 4{,}000-token one.
Section~\ref{sec:hw} also repeats DFlash on Qwen3-8B on an H100.

\paragraph{Evaluation contract.}
We use the qualitative split of SPEED-Bench~\citep{abramovich2026speedbench},
11 categories of 80 prompts each, 880 in total. We run at temperatures 0 and 1
with seed 7, up to 4{,}096 new tokens, a 32{,}768-token context and batch
size 1. The context is long enough that no multi-turn prompt overflows in any variant, so every variant completes
all 880 prompts. We also run eight standard benchmarks (GSM8K, MATH-500,
AIME25, HumanEval, MBPP, LiveCodeBench, MT-Bench and Alpaca), 80 prompts each
(30 for AIME25), with a 4{,}096-token context and up to 1{,}024 new tokens.
\section{Results}
\label{sec:results}

\providecommand{\headlinetitle}{Main result at temperature 0.}
\providecommand{\headlinedflash}{ --- block-parallel, borrows the target's full-vocabulary head}
\providecommand{\headlineeagle}{ --- one layer $\times$ 7 steps, trains its own head over a pruned draft vocabulary}
\begin{widetable}
\centering
\footnotesize
\setlength{\tabcolsep}{4pt}
\caption{\textbf{\headlinetitle} 880 SPEED-Bench qualitative
prompts, batch 1, one B200. $\AL$ is completion tokens over verify steps.
$\Titer$ and $\TPOT$ exclude each turn's first decode step, which includes
prefill (App.~\ref{app:profiling}). $\Delta\AL$ is relative to our full-width
draft, and \emph{stock} rows are released full-width drafters, shown as
training references. Bold marks each cell's best $\AL$ and $\TPOT$ speedup.
Temperature 1 is in App.~\ref{app:t1}.}
\label{tab:headline}
\begin{tabular}{llrrrrrr}
\toprule
Target & Draft & $\AL$ & $\Delta\AL$ & $\Titer$ (ms) & $\TPOT$ (ms) & $\Titer$ speedup & \textbf{\boldmath$\TPOT$ speedup} \\
\midrule
\multicolumn{8}{l}{\textbf{DFlash}\headlinedflash}\\
\midrule
\multirow{4}{*}{Qwen3-8B}
 & full 4{,}096          & 4.298 & ---       & 3.585 & 0.834 & 1.00$\times$ & 1.00$\times$ \\
 & half 2{,}048    & \textbf{4.334} & $+0.84\%$ & 3.313 & 0.764 & 1.08$\times$ & \textbf{1.09$\times$} \\
 & quarter 1{,}024 & 4.064 & $-5.44\%$ & 3.197 & 0.787 & 1.12$\times$ & 1.06$\times$ \\
 & \emph{stock} \texttt{z-lab} & \emph{4.304} & \emph{$+0.14\%$} & \emph{3.584} & \emph{0.833} & \emph{1.00$\times$} & \emph{1.00$\times$} \\
\midrule
\multirow{4}{*}{GPT-OSS-20B}
 & full 2{,}880          & \textbf{4.376} & ---       & 3.342 & 0.765 & 1.00$\times$ & 1.00$\times$ \\
 & half 1{,}440    & 4.309 & $-1.53\%$ & 3.115 & 0.724 & 1.07$\times$ & 1.06$\times$ \\
 & quarter 720     & 4.220 & $-3.56\%$ & 3.002 & 0.712 & 1.11$\times$ & \textbf{1.07$\times$} \\
 & \emph{stock} \texttt{z-lab} & \emph{4.309} & \emph{$-1.53\%$} & \emph{3.339} & \emph{0.776} & \emph{1.00$\times$} & \emph{0.99$\times$} \\
\midrule
\multicolumn{8}{l}{\textbf{EAGLE-3}\headlineeagle}\\
\midrule
\multirow{4}{*}{Qwen3-8B}
 & full 4{,}096          & \textbf{3.583} & ---       & 4.236 & 1.181 & 1.00$\times$ & 1.00$\times$ \\
 & half 2{,}048           & 3.476 & $-2.99\%$ & 3.796 & 1.092 & 1.12$\times$ & 1.08$\times$ \\
 & quarter 1{,}024        & 3.450 & $-3.71\%$ & 3.652 & 1.058 & 1.16$\times$ & \textbf{1.12$\times$} \\
 & \emph{stock} \texttt{RedHatAI} & \emph{2.846} & \emph{$-20.6\%$} & \emph{4.209} & \emph{1.478} & \emph{1.01$\times$} & \emph{0.80$\times$} \\
\midrule
\multirow{4}{*}{GPT-OSS-20B}
 & full 2{,}880          & 3.705 & ---       & 3.651 & 0.987 & 1.00$\times$ & 1.00$\times$ \\
 & half 1{,}440           & \textbf{3.783} & $+2.11\%$ & 3.414 & 0.902 & 1.07$\times$ & \textbf{1.09$\times$} \\
 & quarter 720            & 3.627 & $-2.11\%$ & 3.275 & 0.902 & 1.12$\times$ & \textbf{1.09$\times$} \\
 & \emph{stock} \texttt{RedHatAI} & \emph{3.413} & \emph{$-7.88\%$} & \emph{3.654} & \emph{1.072} & \emph{1.00$\times$} & \emph{0.92$\times$} \\
\bottomrule
\end{tabular}
\end{widetable}

\subsection{Headline}

On B200, narrowing the draft's residual width pays off in all four cells. Halving gives
\textbf{1.06--1.09$\times$} end to end at temperature 0, and quartering gives
\textbf{1.06--1.12$\times$}. On $\Titer$ alone every cell gets faster as the draft narrows, by up to 1.16$\times$. What separates the cells
is acceptance.

\textbf{Halving preserves acceptance closely on the primary panel.} It costs at most 3.0\% acceptance in any cell, and in two of the four cells acceptance
does not drop. There the half-width draft lands 0.84\% (DFlash-Qwen) and 2.11\%
(EAGLE-3-GPT-OSS) above full width. Each variant is a single training run with one
seed, so we read these as parity rather than a gain. Quartering costs 2.1--5.4\%.

\textbf{Both narrow widths win, and which one wins most depends on the cell.}
Halving is fastest on DFlash-Qwen, and quartering on DFlash-GPT-OSS and EAGLE-3-Qwen. On
EAGLE-3-GPT-OSS the two tie to three decimals for different reasons, with
quarter having the cheaper iteration and half the better acceptance
(Table~\ref{tab:headline}).

\textbf{On DGX Spark (GB10) the same drafts gain more.} Halving gives
\textbf{1.13--1.23$\times$} end to end at temperature 0 and quartering
\textbf{1.20--1.44$\times$}, and quarter width is the fastest variant in every
cell (Table~\ref{tab:gb10summary}, App.~\ref{app:gb10}). Narrowing costs
acceptance within 1.8 points of what it costs on B200, so the extra gain comes
from a faster iteration (\S\ref{sec:hw}).

\begin{table}[H]
\centering
\footnotesize
\setlength{\tabcolsep}{4pt}
\caption{\textbf{Main result on DGX Spark (GB10) at temperature 0}, batch 1,
same drafts and panel as Table~\ref{tab:headline}. $\Delta\AL$ is relative to
the full-width draft. Full detail is in Table~\ref{tab:gb10headline}.}
\label{tab:gb10summary}
\begin{tabular}{llrrrr}
\toprule
 & & \multicolumn{2}{c}{$\Delta\AL$} & \multicolumn{2}{c}{$\TPOT$ speedup} \\
\cmidrule(lr){3-4}\cmidrule(lr){5-6}
Family & Target & half & quarter & half & quarter \\
\midrule
DFlash  & Qwen3-8B    & $-0.17\%$ & $-6.17\%$ & 1.20$\times$ & \textbf{1.26$\times$} \\
DFlash  & GPT-OSS-20B & $-2.02\%$ & $-3.89\%$ & 1.13$\times$ & \textbf{1.20$\times$} \\
EAGLE-3 & Qwen3-8B    & $-3.02\%$ & $-2.81\%$ & 1.21$\times$ & \textbf{1.44$\times$} \\
EAGLE-3 & GPT-OSS-20B & $+0.33\%$ & $-2.65\%$ & 1.23$\times$ & \textbf{1.34$\times$} \\
\bottomrule
\end{tabular}
\end{table}

\subsection{Outside the panel's prompt distribution}

\begin{widefigure}
\centering
\begin{subfigure}{0.49\linewidth}
\centering
\begin{tikzpicture}
\begin{axis}[
  title={DFlash, Qwen3-8B}, title style={font=\scriptsize, at={(0,1)}, anchor=south west, yshift=-4pt},
  ybar, area legend, width=1.04\linewidth, height=3.4cm, bar width=2.9pt, ymin=0, ymax=8.6, ytick={0,2,4,6,8},
  ylabel={acceptance length},
  symbolic x coords={gsm8k,math500,aime25,humaneval,mbpp,livecodebench,mtbench,alpaca},
  xtick=data, xticklabels={}, x tick label style={rotate=35, anchor=east, font=\scriptsize},
  tick label style={font=\scriptsize}, label style={font=\scriptsize},
  ymajorgrids, grid style={black!12}, axis lines*=left, enlarge x limits=0.07,
  legend style={font=\scriptsize, at={(1,1)}, anchor=south east, inner sep=0pt, legend columns=3, draw=none, fill=none},
]
\addplot+[fill=armexact,draw=armexact!60!black] coordinates {(gsm8k,5.751) (math500,6.761) (aime25,5.744) (humaneval,6.846) (mbpp,5.809) (livecodebench,4.866) (mtbench,3.278) (alpaca,2.931)};
\addplot+[fill=armhalf,draw=armhalf!60!black] coordinates {(gsm8k,5.801) (math500,6.778) (aime25,5.763) (humaneval,6.746) (mbpp,5.833) (livecodebench,4.868) (mtbench,3.270) (alpaca,2.914)};
\addplot+[fill=armqtr,draw=armqtr!60!black] coordinates {(gsm8k,5.657) (math500,6.708) (aime25,5.705) (humaneval,6.477) (mbpp,5.715) (livecodebench,4.802) (mtbench,3.207) (alpaca,2.825)};
\legend{full, half, quarter}
\end{axis}
\end{tikzpicture}
\end{subfigure}
\hfill
\begin{subfigure}{0.49\linewidth}
\centering
\begin{tikzpicture}
\begin{axis}[
  title={DFlash, GPT-OSS-20B}, title style={font=\scriptsize, at={(0,1)}, anchor=south west, yshift=-4pt},
  ybar, area legend, width=1.04\linewidth, height=3.4cm, bar width=2.9pt, ymin=0, ymax=5.0, ytick={0,1,2,3,4,5},
  ylabel={acceptance length},
  symbolic x coords={gsm8k,math500,aime25,humaneval,mbpp,livecodebench,mtbench,alpaca},
  xtick=data, xticklabels={}, x tick label style={rotate=35, anchor=east, font=\scriptsize},
  tick label style={font=\scriptsize}, label style={font=\scriptsize},
  ymajorgrids, grid style={black!12}, axis lines*=left, enlarge x limits=0.07,
]
\addplot+[fill=armexact,draw=armexact!60!black] coordinates {(gsm8k,4.470) (math500,4.524) (aime25,3.851) (humaneval,4.328) (mbpp,4.299) (livecodebench,3.471) (mtbench,3.732) (alpaca,3.854)};
\addplot+[fill=armhalf,draw=armhalf!60!black] coordinates {(gsm8k,4.413) (math500,4.466) (aime25,3.783) (humaneval,4.262) (mbpp,4.267) (livecodebench,3.404) (mtbench,3.668) (alpaca,3.777)};
\addplot+[fill=armqtr,draw=armqtr!60!black] coordinates {(gsm8k,4.300) (math500,4.371) (aime25,3.669) (humaneval,4.135) (mbpp,4.154) (livecodebench,3.362) (mtbench,3.573) (alpaca,3.692)};
\end{axis}
\end{tikzpicture}
\end{subfigure}
\\[-0.2em]
\begin{subfigure}{0.49\linewidth}
\centering
\begin{tikzpicture}
\begin{axis}[
  title={EAGLE-3, Qwen3-8B}, title style={font=\scriptsize, at={(0,1)}, anchor=south west, yshift=-4pt},
  ybar, area legend, width=1.04\linewidth, height=3.4cm, bar width=2.9pt, ymin=0, ymax=8.6, ytick={0,2,4,6,8},
  ylabel={acceptance length},
  symbolic x coords={gsm8k,math500,aime25,humaneval,mbpp,livecodebench,mtbench,alpaca},
  xtick=data, x tick label style={rotate=35, anchor=east, font=\scriptsize},
  tick label style={font=\scriptsize}, label style={font=\scriptsize},
  ymajorgrids, grid style={black!12}, axis lines*=left, enlarge x limits=0.07,
]
\addplot+[fill=armexact,draw=armexact!60!black] coordinates {(gsm8k,4.797) (math500,5.576) (aime25,5.003) (humaneval,5.471) (mbpp,4.989) (livecodebench,4.635) (mtbench,3.221) (alpaca,2.995)};
\addplot+[fill=armhalf,draw=armhalf!60!black] coordinates {(gsm8k,4.511) (math500,5.232) (aime25,4.708) (humaneval,5.174) (mbpp,4.720) (livecodebench,4.434) (mtbench,3.048) (alpaca,2.815)};
\addplot+[fill=armqtr,draw=armqtr!60!black] coordinates {(gsm8k,4.543) (math500,5.264) (aime25,4.717) (humaneval,5.200) (mbpp,4.783) (livecodebench,4.448) (mtbench,3.001) (alpaca,2.772)};
\end{axis}
\end{tikzpicture}
\end{subfigure}
\hfill
\begin{subfigure}{0.49\linewidth}
\centering
\begin{tikzpicture}
\begin{axis}[
  title={EAGLE-3, GPT-OSS-20B}, title style={font=\scriptsize, at={(0,1)}, anchor=south west, yshift=-4pt},
  ybar, area legend, width=1.04\linewidth, height=3.4cm, bar width=2.9pt, ymin=0, ymax=5.0, ytick={0,1,2,3,4,5},
  ylabel={acceptance length},
  symbolic x coords={gsm8k,math500,aime25,humaneval,mbpp,livecodebench,mtbench,alpaca},
  xtick=data, x tick label style={rotate=35, anchor=east, font=\scriptsize},
  tick label style={font=\scriptsize}, label style={font=\scriptsize},
  ymajorgrids, grid style={black!12}, axis lines*=left, enlarge x limits=0.07,
]
\addplot+[fill=armexact,draw=armexact!60!black] coordinates {(gsm8k,3.810) (math500,4.052) (aime25,3.382) (humaneval,3.862) (mbpp,3.888) (livecodebench,3.164) (mtbench,3.359) (alpaca,3.498)};
\addplot+[fill=armhalf,draw=armhalf!60!black] coordinates {(gsm8k,3.760) (math500,4.002) (aime25,3.361) (humaneval,3.783) (mbpp,3.821) (livecodebench,3.095) (mtbench,3.300) (alpaca,3.446)};
\addplot+[fill=armqtr,draw=armqtr!60!black] coordinates {(gsm8k,3.739) (math500,3.920) (aime25,3.320) (humaneval,3.725) (mbpp,3.764) (livecodebench,3.034) (mtbench,3.231) (alpaca,3.376)};
\end{axis}
\end{tikzpicture}
\end{subfigure}
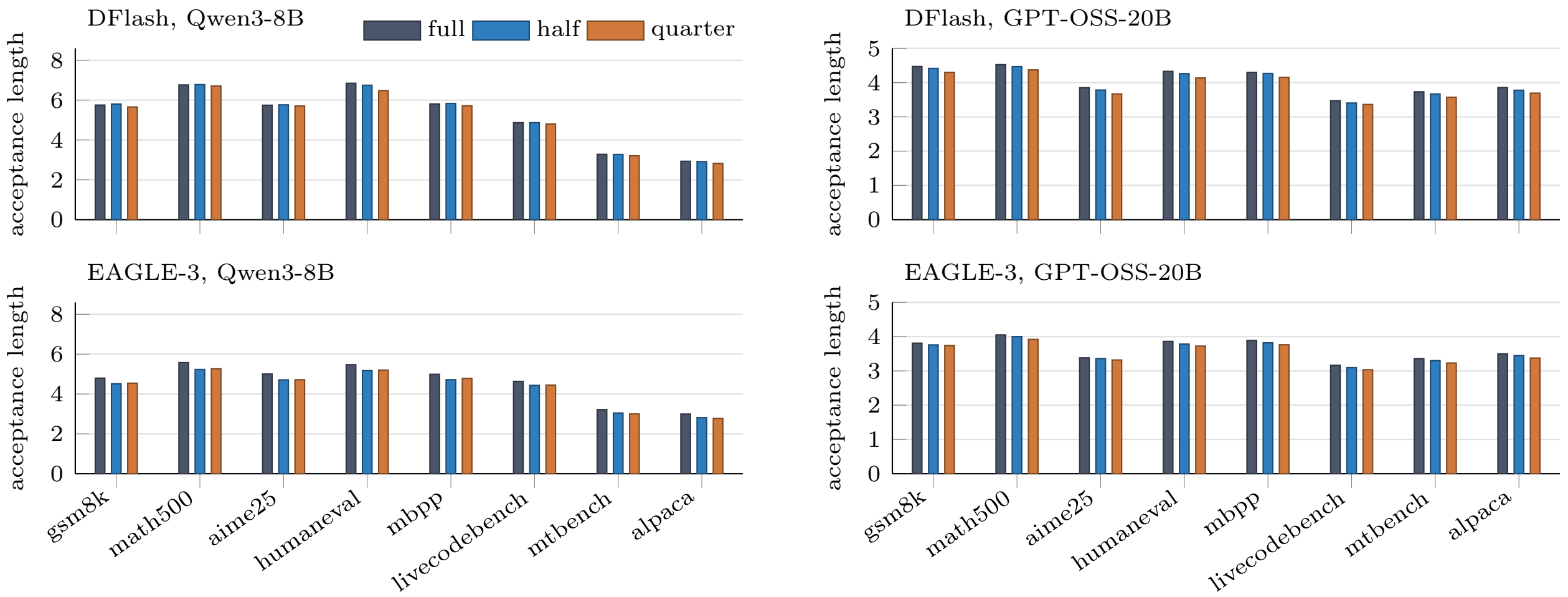
\caption{\textbf{Narrowing keeps acceptance close to full width on eight standard
benchmarks in three of four cells.}
Acceptance length at temperature 0\arxivgpu{ on B200} (setup in \S\ref{sec:setup}). Temperature 1
is in
Table~\ref{tab:bench8t1}.}
\label{fig:bench8}
\end{widefigure}

We also evaluated every variant on eight standard math, code and chat
benchmarks (Figure~\ref{fig:bench8}), with acceptance and timing from the same run.
\textbf{Narrowing keeps acceptance close to full width in three of four cells.}
In those three the half-width draft's acceptance length is within 2.2\% of full width on
every benchmark, and on DFlash-Qwen it is above full width on 5 of 8. On
EAGLE-3-Qwen halving costs 4.3--6.2\% acceptance on every benchmark, somewhat
more than its 3.0\% on the panel. Quartering never costs more than 7.4\%
acceptance on any benchmark. \textbf{The speedup carries over too.} Pooled
over the suite at temperature 0, halving speeds up decoding by
1.06--1.09$\times$ and quartering by 1.06--1.10$\times$, similar to the
880-prompt panel.

\section{Top-down latency attribution}
\label{sec:breakdown}

We break the speedup down from the top. First we split the iteration into
draft and verify, then the draft into decoder layers and the vocabulary
projection, and finally the draft's saving into individual operations. We run
all three on all four cells so they line up. Every profile uses the production
settings and is checked against an unprofiled run of the same prompt
(App.~\ref{app:profiling}). The ratios agree to 0.2--0.9\%, which is
why we can use the proportions in this section to explain the absolute
numbers in \S\ref{sec:results}.

\subsection{Iteration and draft --- the draft shrinks}

\begin{widetable}
\centering
\footnotesize
\setlength{\tabcolsep}{4pt}
\caption{\textbf{The draft shrinks and verify does not.} Per-iteration
stage times\arxivgpu{ on B200 at batch 1} under the production settings (App.~\ref{app:profiling}).}
\label{tab:l1}
\begin{tabular}{llrrrr}
\toprule
Family / target & Draft & $\tdraft$ (ms) & vs full & $\tverify$ (ms) & sum, vs full \\
\midrule
\multirow{3}{*}{DFlash, Qwen3-8B}
 & full 4{,}096          & 0.906 & ---                & 2.578 & 3.484 \\
 & half 2{,}048    & 0.634 & \textbf{\textminus30.0\%} & 2.576 ($-0.09\%$) & 3.210 (1.09$\times$) \\
 & quarter 1{,}024 & 0.517 & \textbf{\textminus43.0\%} & 2.590 ($+0.44\%$) & 3.106 (1.12$\times$) \\
\midrule
\multirow{3}{*}{DFlash, GPT-OSS-20B}
 & full 2{,}880       & 0.956 & ---                & 2.216 & 3.171 \\
 & half 1{,}440 & 0.724 & \textbf{\textminus24.3\%} & 2.227 ($+0.53\%$) & 2.951 (1.08$\times$) \\
 & quarter 720  & 0.641 & \textbf{\textminus32.9\%} & 2.202 ($-0.62\%$) & 2.842 (1.12$\times$) \\
\midrule
\multirow{3}{*}{EAGLE-3, Qwen3-8B}
 & full 4{,}096   & 1.439 & ---                & 2.735 & 4.174 \\
 & half 2{,}048    & 0.989 & \textbf{\textminus31.3\%} & 2.718 ($-0.61\%$) & 3.707 (1.13$\times$) \\
 & quarter 1{,}024 & 0.827 & \textbf{\textminus42.5\%} & 2.719 ($-0.57\%$) & 3.546 (1.18$\times$) \\
\midrule
\multirow{3}{*}{EAGLE-3, GPT-OSS-20B}
 & full 2{,}880 & 1.134 & ---                & 2.598 & 3.732 \\
 & half 1{,}440  & 0.911 & \textbf{\textminus19.7\%} & 2.622 ($+0.92\%$) & 3.533 (1.06$\times$) \\
 & quarter 720   & 0.793 & \textbf{\textminus30.1\%} & 2.581 ($-0.65\%$) & 3.374 (1.11$\times$) \\
\bottomrule
\end{tabular}
\end{widetable}

The draft runs 1.25--1.46$\times$ faster at half width and 1.43--1.75$\times$
faster at quarter width, while verify stays within 1\% in every cell.
\textbf{Verify staying flat across three variants provides a useful internal
consistency check.} It holds across two drafter families, a dense target and
an MoE target. Within the draft, the decoder layers and the vocabulary
projection both shrink in every cell. In
three of four cells the decoder layers account for more of the absolute
saving. At quarter width they save 181--394\,$\mu$s, against 121--209\,$\mu$s
for the vocabulary projection and its argmax (full minus quarter in
Table~\ref{tab:l2}), on DFlash-Qwen, DFlash-GPT-OSS and EAGLE-3-Qwen.
EAGLE-3-GPT-OSS, whose single layer is small, is the exception.

\begin{widetable}
\centering
\footnotesize
\setlength{\tabcolsep}{4pt}
\caption{\textbf{Both the decoder layers and the head shrink.} $\mu$s per
iteration\arxivgpu{ on B200}. The draft total is $\tdraft$ from Table~\ref{tab:l1}, so its ratios
match that table. \emph{Rest} is the part of $\tdraft$ outside the layers and
the head. For DFlash it is the context pre-pass, block preparation, the final
norm and the input adapter, plus inter-kernel gaps on Qwen3-8B, and for EAGLE-3 it is the carry-in GEMM
(App.~\ref{app:l3full}).}
\label{tab:l2}
\begin{tabular}{llrrrr}
\toprule
Family / target & Draft & draft total $\tdraft$ & decoder layers & head $+$ argmax & rest \\
\midrule
\multirow{3}{*}{DFlash, Qwen3-8B}
 & full 4{,}096   & 906.1 & 527.8 & 218.5 & 159.8 \\
 & half     & 634.1 (1.43$\times$) & 373.6 (1.41$\times$) & 111.6 (1.96$\times$) & 148.9 \\
 & quarter  & 516.6 (1.75$\times$) & 323.4 (1.63$\times$) &  65.4 (3.34$\times$) & 127.8 \\
\midrule
\multirow{3}{*}{DFlash, GPT-OSS-20B}
 & full 2{,}880   & 955.5 & 613.4 & 179.0 & 163.1 \\
 & half     & 723.5 (1.32$\times$) & 478.4 (1.28$\times$) &  99.3 (1.80$\times$) & 145.8 \\
 & quarter  & 640.7 (1.49$\times$) & 432.0 (1.42$\times$) &  58.5 (3.06$\times$) & 150.2 \\
\midrule
\multirow{3}{*}{EAGLE-3, Qwen3-8B}
 & full 4{,}096   & 1439.1 & 1078.3 & 342.1 & 18.7 \\
 & half            &  989.0 (1.46$\times$) &  768.4 (1.40$\times$) & 210.0 (1.63$\times$) & 10.6 \\
 & quarter         &  827.1 (1.74$\times$) &  684.1 (1.58$\times$) & 132.7 (2.58$\times$) & 10.3 \\
\midrule
\multirow{3}{*}{EAGLE-3, GPT-OSS-20B}
 & full 2{,}880   & 1134.3 & 682.7 & 441.2 & 10.5 \\
 & half            &  911.0 (1.25$\times$) & 639.1 (1.07$\times$) & 264.9 (1.67$\times$) & 7.0 \\
 & quarter         &  792.6 (1.43$\times$) & 612.8 (1.11$\times$) & 173.0 (2.55$\times$) & 6.9 \\
\bottomrule
\end{tabular}
\end{widetable}

\subsection{Operations --- where the saving comes from}

\begin{widetable}
\centering
\footnotesize
\setlength{\tabcolsep}{3pt}
\caption{\textbf{Where the draft-side saving comes from.} $\mu$s per
iteration\arxivgpu{ on B200}, ratio to full width in brackets. \emph{Rest of body} is
attention, norms, activation, sampling, bookkeeping and the DFlash input adapter.
\emph{Outside} covers the DFlash context pre-pass (with inter-kernel gaps on
Qwen3-8B) and the EAGLE-3 carry-in GEMM, so each row sums to $\tdraft$
(Table~\ref{tab:l1}). Per-operation detail is in
App.~\ref{app:l3full}.}
\label{tab:l3}
\begin{tabular}{llrrrrrr}
\toprule
Family / target & Draft & attn.\ proj. & MLP & head & rest of body & outside & $\tdraft$ \\
\midrule
\multirow{3}{*}{DFlash, Qwen3-8B}
 & full 4{,}096 & 96.3 & 271.4 & 206.2 & 199.4 & 132.9 & 906.1 \\
 & half & 70.1 (1.37$\times$) & 148.5 (1.83$\times$) & 99.4 (2.07$\times$) & 203.5 (0.98$\times$) & 112.6 & 634.1 (1.43$\times$) \\
 & quarter & 53.4 (1.80$\times$) & 103.2 (2.63$\times$) & 53.3 (3.87$\times$) & 212.7 (0.94$\times$) & 94.0 & 516.6 (1.75$\times$) \\
\midrule
\multirow{3}{*}{DFlash, GPT-OSS-20B}
 & full 2{,}880 & 113.7 & 211.6 & 166.5 & 325.3 & 138.4 & 955.5 \\
 & half & 85.9 (1.32$\times$) & 131.8 (1.61$\times$) & 86.9 (1.92$\times$) & 300.7 (1.08$\times$) & 118.2 & 723.5 (1.32$\times$) \\
 & quarter & 74.8 (1.52$\times$) & 101.7 (2.08$\times$) & 46.1 (3.61$\times$) & 297.8 (1.09$\times$) & 120.2 & 640.7 (1.49$\times$) \\
\midrule
\multirow{3}{*}{EAGLE-3, Qwen3-8B}
 & full 4{,}096 & 193.1 & 366.9 & 304.9 & 555.5 & 18.7 & 1439.1 \\
 & half & 121.7 (1.59$\times$) & 207.1 (1.77$\times$) & 174.2 (1.75$\times$) & 475.4 (1.17$\times$) & 10.6 & 989.0 (1.46$\times$) \\
 & quarter & 86.2 (2.24$\times$) & 149.6 (2.45$\times$) & 96.4 (3.16$\times$) & 484.6 (1.15$\times$) & 10.3 & 827.1 (1.74$\times$) \\
\midrule
\multirow{3}{*}{EAGLE-3, GPT-OSS-20B}
 & full 2{,}880 & 138.6 & 100.6 & 403.2 & 481.4 & 10.5 & 1134.3 \\
 & half & 99.0 (1.40$\times$) & 85.1 (1.18$\times$) & 226.0 (1.78$\times$) & 493.9 (0.97$\times$) & 7.0 & 911.0 (1.25$\times$) \\
 & quarter & 74.7 (1.86$\times$) & 61.9 (1.63$\times$) & 130.7 (3.08$\times$) & 518.4 (0.93$\times$) & 6.9 & 792.6 (1.43$\times$) \\
\bottomrule
\end{tabular}
\end{widetable}

\textbf{The saving is spread across the draft.} In three of four cells the MLP
projections give 36--49\% of the draft-body saving, the attention projections 10--18\% and the
vocabulary projection 30--44\% (App.~\ref{app:l3full}). On DFlash the context
pre-pass, which these totals leave out, shrinks too (Table~\ref{tab:l3}, outside).
The split follows the draft's shapes. EAGLE-3 on GPT-OSS-20B pairs a 64{,}000-entry draft vocabulary with an
\texttt{intermediate\_size} of only 2{,}880, and there the vocabulary
projection supplies 81\%. At quarter width the vocabulary projection runs
3.08--3.87$\times$ faster and FC1 1.96--3.23$\times$, while the attention
output projection and FC2, which write the residual stream, reach only
1.25--1.92$\times$. Attention, whose head count, head dimension and block size
do not change, stays at 0.96--1.04$\times$. EAGLE-3-GPT-OSS's FC2 is slower
at half width than at full width because it switches to a different kernel.

\FloatBarrier
\subsection{Width against MLP-only and head-only cuts}
\label{sec:whywidth}

A narrower draft trades capacity for latency, and so would a smaller MLP or a
cheaper head. Under the measured operation costs
(Table~\ref{tab:l3}), even eliminating the head or the whole MLP cannot match
quarter width's draft-latency reduction in three of four cells. On DFlash-Qwen
they would save 206 and 271\,$\mu$s against 390\,$\mu$s, so no head-only
method, low-rank~\citep{slimspec2026} or trimmed, and no MLP-only cut can save
more draft time. EAGLE-3-GPT-OSS, whose head dominates, is the exception.
\begin{widetable}
\centering
\footnotesize
\setlength{\tabcolsep}{4pt}
\caption{\textbf{Acceptance an MLP-only draft would need.} Ablation. The
MLP-only draft keeps full width and cuts only the MLP's intermediate size by
the width factor. It is an untrained slice of the full-width draft, timed
beside our trained drafts at $B{=}1$\arxivgpu{ on B200} in a separate run with the
Table~\ref{tab:batch} server settings\arxivgpu{ but a 2{,}048-token context, on one of its prompts}. Speedups are $\TPOT$ over full width
from that run's $\Titer$ with the $\AL$ of Table~\ref{tab:headline}, so they
differ slightly from Tables~\ref{tab:headline} and~\ref{tab:batch}. $\AL$ needed is
what the MLP-only draft needs to tie our narrow draft. $\AL$ changes are relative to
full width, so every positive entry means the MLP-only draft must accept more
than the full-width draft it was cut from.}
\label{tab:breakeven}
\begin{tabular}{llrrrrrr}
\toprule
 & & \multicolumn{3}{c}{half width vs MLP $\div 2$} & \multicolumn{3}{c}{quarter width vs MLP $\div 4$} \\
\cmidrule(lr){3-5}\cmidrule(lr){6-8}
Family & Target & $\TPOT$ speedup & draft $\Delta\AL$ & $\AL$ needed & $\TPOT$ speedup & draft $\Delta\AL$ & $\AL$ needed \\
\midrule
DFlash  & Qwen3-8B    & 1.09$\times$ & $+0.8\%$ & $+4.9\%$  & 1.07$\times$ & $-5.4\%$ & $+0.7\%$ \\
DFlash  & GPT-OSS-20B & 1.07$\times$ & $-1.5\%$ & $+2.1\%$  & 1.08$\times$ & $-3.6\%$ & $+3.0\%$ \\
EAGLE-3 & Qwen3-8B    & 1.08$\times$ & $-3.0\%$ & $+4.8\%$  & 1.14$\times$ & $-3.7\%$ & $+7.5\%$ \\
EAGLE-3 & GPT-OSS-20B & 1.08$\times$ & $+2.1\%$ & $+10.6\%$ & 1.06$\times$ & $-2.1\%$ & $+5.1\%$ \\
\bottomrule
\end{tabular}
\end{widetable}
 We also ran the MLP alternative as an ablation, because the MLP
is the largest block in three of four cells and gives the largest saving at
half width there. The MLP-only draft keeps full width and cuts only the MLP's
intermediate size by our width factor. Its weights are untrained slices of the
full-width draft, so we use it for latency only. Fixed kernel shapes do not
guarantee that a trained draft sees the same runtime costs, so these timings
are an estimate. \textbf{To tie our narrow drafts, an MLP-only draft would need a
higher $\AL$ than the full-width draft it was cut from.} Timed beside our
trained drafts at batch 1, it would need 2.1--10.6\% more $\AL$ than full width
to tie our half-width draft, and 0.7--7.5\% more to tie our quarter-width draft
(Table~\ref{tab:breakeven}). Against our fastest draft in each cell the hurdles
rise to 3.6--10.9\% and 3.0--7.9\%. The cut saves as much MLP time as
narrowing does, 120 against 122\,$\mu$s on DFlash-Qwen at half width, but nothing in
the head or the attention projections, so it captures only 41--55\% of our
variants' draft-body saving. On EAGLE-3-GPT-OSS, where the MLP is already near the
small-GEMM floor, it is no faster than full width within about 2\% noise. Its timings come
from one run of all five drafts on one node. A trained smaller MLP could
accept more than its slice, but it would still have to beat full
width by these margins. A trained MLP-only or head-only comparison is
future work.

\subsection{On GB10}

The same split on GB10 (Tables~\ref{tab:gb10l1} and~\ref{tab:gb10l2},
App.~\ref{app:gb10}) gives the full-width draft 28.9--45.4\% of the iteration,
against 26.0--34.5\% on B200, and verify again moves by at most 1.8\% with
width. The quarter-width draft runs 3.44--3.87$\times$ faster than full width,
against 1.43--1.75$\times$ on B200, because the MLP and head times roughly
halve with each halving of width while attention stays at 0.08--0.19\,ms.

\section{Batch size and context length}
\label{sec:scale}

So far every number is at batch 1. We also swept batch size from 1 to 512,
measuring $\AL$ and $\Titer$ in the same runs, and context length from 8k to
32k, which measures latency only. The batch sweep serves 512 SPEED-Bench
throughput prompts rather than the 880-prompt panel (App.~\ref{app:scope}), and
its acceptance costs differ from Table~\ref{tab:headline}'s by up to 4.5
points, so Table~\ref{tab:headline} remains our acceptance result. Within the
sweep each variant's $\AL$ ratio moves by at most 0.8 points across $B$, so
narrowing costs the same acceptance at every batch size and the $\TPOT$
speedup follows the $\Titer$ speedup. Quarter-width DFlash-Qwen, for example,
is 1.21$\times$ faster per iteration at $B{=}512$ and accepts 1.5\% fewer
tokens, so its $\TPOT$ is 1.19$\times$ faster.

\begin{widetable}
\centering
\footnotesize
\setlength{\tabcolsep}{4pt}
\caption{\textbf{End-to-end speedup over full width across batch size\arxivgpu{ on B200}.} $\TPOT$ speedup on
512 SPEED-Bench throughput prompts, computed from the full-batch $\Titer$ and
whole-run $\AL$ of the same run (App.~\ref{app:scope}). Draft share is full-width draft time over draft plus
verify\arxivgpu{, from a separate Nsight capture of about ten full-batch steps on
the first $B$ of the same prompts}. $\Titer$ speedups, $\AL$ ratios and stage times are in
Table~\ref{tab:batchstages}.}
\label{tab:batch}
\begin{tabular}{lllrrrrrr}
\toprule
Family & Target & Row & $B{=}1$ & $32$ & $64$ & $128$ & $256$ & $512$ \\
\midrule
\multirow{6}{*}{DFlash} & \multirow{3}{*}{Qwen3-8B} & half & 1.09$\times$ & 1.12$\times$ & 1.14$\times$ & 1.16$\times$ & 1.13$\times$ & 1.13$\times$ \\
 &  & quarter & 1.13$\times$ & 1.17$\times$ & 1.16$\times$ & 1.21$\times$ & 1.18$\times$ & 1.19$\times$ \\
 &  & draft share & 25.4\% & 26.5\% & 28.8\% & 29.9\% & 30.2\% & 31.4\% \\
 & \multirow{3}{*}{GPT-OSS-20B} & half & 1.05$\times$ & 1.05$\times$ & 1.05$\times$ & 1.09$\times$ & 1.07$\times$ & 1.09$\times$ \\
 &  & quarter & 1.07$\times$ & 1.04$\times$ & 1.06$\times$ & 1.09$\times$ & 1.09$\times$ & 1.09$\times$ \\
 &  & draft share & 30.7\% & 25.0\% & 28.7\% & 32.5\% & 32.3\% & 35.7\% \\
\midrule
\multirow{6}{*}{EAGLE-3} & \multirow{3}{*}{Qwen3-8B} & half & 1.08$\times$ & 1.05$\times$ & 1.04$\times$ & 1.02$\times$ & 1.03$\times$ & 1.02$\times$ \\
 &  & quarter & 1.12$\times$ & 1.07$\times$ & 1.05$\times$ & 1.04$\times$ & 1.05$\times$ & 1.03$\times$ \\
 &  & draft share & 34.2\% & 30.4\% & 27.7\% & 25.2\% & 23.8\% & 24.4\% \\
 & \multirow{3}{*}{GPT-OSS-20B} & half & 1.07$\times$ & 1.02$\times$ & 1.00$\times$ & 1.05$\times$ & 1.03$\times$ & 1.01$\times$ \\
 &  & quarter & 1.10$\times$ & 1.04$\times$ & 1.01$\times$ & 1.04$\times$ & 1.03$\times$ & 1.04$\times$ \\
 &  & draft share & 31.2\% & 24.8\% & 24.4\% & 24.0\% & 22.6\% & 24.6\% \\
\bottomrule
\end{tabular}
\end{widetable}

\textbf{Narrowing speeds up the iteration at every batch size we measured.} From
$B{=}1$ to 512, half width speeds up the iteration by 1.02--1.15$\times$ and
quarter width by 1.05--1.23$\times$ (Table~\ref{tab:batchstages}), and
end-to-end decoding by up to 1.16$\times$ and 1.21$\times$, with the smallest
gain at parity (Table~\ref{tab:batch}). The trend with batch goes both ways, though. The gain grows on DFlash and shrinks or holds on EAGLE-3. With batch, the
draft's share of the iteration and the narrow draft's own speedup generally
hold or grow on DFlash, while on EAGLE-3 the share falls or holds and the
speedup falls.

\textbf{On DFlash the draft's share grows at large batch.} The draft runs
$\blk$ positions per request, the bonus token and $\blk-1$ masks, so a batch of $B$ gives its GEMMs $\blk B$ token
positions, the same number the verify pass checks. On Qwen3-8B both stages are
dense, so every MLP GEMM in the draft and in verify sees the same $\blk B$
tokens, though their shapes, precision and kernels differ. The share rises
from 25.4\% at $B{=}1$ to 31.4\% at 512, with its largest step, 2.3 points,
between $B{=}32$ and 64. That step is consistent with the two stages
saturating at different batch sizes. NVFP4 runs 4$\times$ the math of BF16 per
second but, with an FP8 scale per 16 weights, reads only 3.56$\times$ fewer
bytes per weight, so the target's NVFP4 GEMMs should saturate at a slightly
larger batch than the draft's BF16 ones, and their differing shapes and
kernels may also contribute.
The draft-level speedup rises too, from 1.39$\times$ at $B{=}1$ to
1.58$\times$ at $B{=}128$ at half width and from 1.71$\times$ to 2.16$\times$
at quarter, before easing to 1.48$\times$ and 1.99$\times$ at 512, so
quarter's iteration speedup peaks at 1.23$\times$ at $B{=}128$. On GPT-OSS-20B the target's MLP is a mixture of experts that routes each
verify token to 4 of its 32 experts. At $B{=}1$ the 8 verify tokens touch only
part of each layer's experts, and by $B{=}32$ its 256 tokens touch essentially
all of them. Over that range verify time doubles while draft time grows
1.6$\times$, so the draft's share dips from 30.7\% to 25.0\%, plausibly
because verify reads more of its expert weights. Past $B{=}32$ every expert is
already read, and under balanced routing each MXFP4 expert GEMM sees about $B$
tokens and stays weight-bound, so verify grows slowly. The draft's dense MLP
sees all $8B$ tokens and grows faster, and the share rises 11 points from
$B{=}32$ to 512.

\textbf{On EAGLE-3 the draft's share falls.} The verify pass reads the last
accepted token and the seven proposals, so it checks $8B$ token positions. The
draft's first step also covers 8 positions per request, but each of its other
six steps drafts one token per request, so most of its GEMMs see only $B$
tokens. On Qwen3-8B both stages are dense, so verify's MLP GEMMs see $8B$
tokens and turn compute-bound early in the sweep, while most of the draft's GEMMs
stay weight-bound and barely grow with $B$. The share falls from 34.2\% at
$B{=}1$ to 23.8\% at $B{=}256$ and then levels off, and the speedup drifts
from 1.11$\times$ to 1.05$\times$ at half width and from 1.18$\times$ to
1.08$\times$ at quarter. On GPT-OSS-20B, as on DFlash, the 8 verify
tokens at $B{=}1$ touch only part of each layer's experts and the 256 at
$B{=}32$ touch essentially all of them. The share drops from 31.2\% to 24.8\%
over that range, plausibly because verify reads more of its expert weights.
Past $B{=}32$ each expert GEMM sees about $B$ tokens under balanced routing, as
most of the draft's GEMMs do, so both stages stay weight-bound and grow
together, and the share holds at 23--25\% and the speedup at 1.02--1.07$\times$
at half width and 1.05--1.08$\times$ at quarter. With acceptance costs of about 2\% at half width and 3\% at quarter, both narrow drafts then decode from parity to 1.05$\times$ faster than full width.

\begin{widefigure}
\centering
\begin{subfigure}{0.49\linewidth}
\centering
\begin{tikzpicture}
\begin{axis}[width=1.0\linewidth, height=4.6cm, /pgf/number format/1000 sep={},
  xmin=0, xmax=1000, xtick distance=200, ymin=0, ymax=26,
  xlabel={per-user tokens/s}, ylabel={per-GPU tokens/s ($\times 10^3$)},
  tick label style={font=\scriptsize}, label style={font=\scriptsize},
  grid=major, grid style={black!12}, axis lines*=left, clip mode=individual,
  mark size=1.6pt, line width=0.9pt,
  legend style={font=\scriptsize, at={(0.97,0.97)}, anchor=north east, draw=none, fill=white, fill opacity=0.8, text opacity=1}]
\addplot[armexact, mark=*] coordinates {(820.9,0.82) (395.5,12.66) (239.0,15.30) (131.5,16.83) (69.0,17.66) (35.1,17.97)};
\addplot[armhalf, mark=square*] coordinates {(891.5,0.89) (444.4,14.22) (273.0,17.47) (151.9,19.44) (78.2,20.02) (39.6,20.26)};
\addplot[armqtr, mark=triangle*] coordinates {(925.1,0.93) (464.4,14.86) (276.4,17.69) (159.5,20.42) (81.3,20.82) (41.7,21.33)};
\node[font=\tiny, anchor=south west, xshift=-3pt, yshift=3pt] at (axis cs:41.7,21.33) {$B{=}512$};
\node[font=\tiny, anchor=south, yshift=3pt] at (axis cs:925.1,0.93) {$B{=}1$};
\legend{full, half, quarter}
\end{axis}
\end{tikzpicture}
\caption{DFlash, Qwen3-8B.}
\label{fig:pareto-dq}
\end{subfigure}
\hfill
\begin{subfigure}{0.49\linewidth}
\centering
\begin{tikzpicture}
\begin{axis}[width=1.0\linewidth, height=4.6cm, /pgf/number format/1000 sep={},
  xmin=0, xmax=1200, xtick distance=200, ymin=0, ymax=50,
  xlabel={per-user tokens/s}, ylabel={per-GPU tokens/s ($\times 10^3$)},
  tick label style={font=\scriptsize}, label style={font=\scriptsize},
  grid=major, grid style={black!12}, axis lines*=left, clip mode=individual,
  mark size=1.6pt, line width=0.9pt]
\addplot[armexact, mark=*] coordinates {(1020.2,1.02) (530.9,16.99) (385.7,24.69) (246.1,31.50) (137.8,35.27) (74.7,38.25)};
\addplot[armhalf, mark=square*] coordinates {(1073.0,1.07) (558.3,17.87) (405.5,25.95) (268.6,34.38) (148.0,37.88) (81.2,41.55)};
\addplot[armqtr, mark=triangle*] coordinates {(1089.6,1.09) (554.4,17.74) (408.8,26.17) (269.2,34.46) (149.8,38.34) (81.4,41.65)};
\node[font=\tiny, anchor=south west, xshift=-3pt, yshift=3pt] at (axis cs:81.4,41.65) {$B{=}512$};
\node[font=\tiny, anchor=south, yshift=3pt] at (axis cs:1089.6,1.09) {$B{=}1$};
\end{axis}
\end{tikzpicture}
\caption{DFlash, GPT-OSS-20B.}
\label{fig:pareto-dg}
\end{subfigure}
\\[0.4em]
\begin{subfigure}{0.49\linewidth}
\centering
\begin{tikzpicture}
\begin{axis}[width=1.0\linewidth, height=4.6cm, /pgf/number format/1000 sep={},
  xmin=0, xmax=800, xtick distance=200, ymin=0, ymax=39,
  xlabel={per-user tokens/s}, ylabel={per-GPU tokens/s ($\times 10^3$)},
  tick label style={font=\scriptsize}, label style={font=\scriptsize},
  grid=major, grid style={black!12}, axis lines*=left, clip mode=individual,
  mark size=1.6pt, line width=0.9pt]
\addplot[armexact, mark=*] coordinates {(649.4,0.65) (451.5,14.45) (328.4,21.02) (208.3,26.66) (116.5,29.83) (61.7,31.58)};
\addplot[armhalf, mark=square*] coordinates {(701.7,0.70) (472.8,15.13) (340.8,21.81) (212.9,27.25) (119.6,30.61) (62.8,32.16)};
\addplot[armqtr, mark=triangle*] coordinates {(729.5,0.73) (482.9,15.45) (343.3,21.97) (215.6,27.60) (122.8,31.44) (63.5,32.51)};
\node[font=\tiny, anchor=south west, xshift=-3pt, yshift=3pt] at (axis cs:63.5,32.51) {$B{=}512$};
\node[font=\tiny, anchor=south, yshift=3pt] at (axis cs:729.5,0.73) {$B{=}1$};
\end{axis}
\end{tikzpicture}
\caption{EAGLE-3, Qwen3-8B.}
\label{fig:pareto-eq}
\end{subfigure}
\hfill
\begin{subfigure}{0.49\linewidth}
\centering
\begin{tikzpicture}
\begin{axis}[width=1.0\linewidth, height=4.6cm, /pgf/number format/1000 sep={},
  xmin=0, xmax=1000, xtick distance=200, ymin=0, ymax=47,
  xlabel={per-user tokens/s}, ylabel={per-GPU tokens/s ($\times 10^3$)},
  tick label style={font=\scriptsize}, label style={font=\scriptsize},
  grid=major, grid style={black!12}, axis lines*=left, clip mode=individual,
  mark size=1.6pt, line width=0.9pt]
\addplot[armexact, mark=*] coordinates {(796.6,0.80) (421.5,13.49) (332.2,21.26) (226.5,28.99) (131.2,33.59) (73.7,37.72)};
\addplot[armhalf, mark=square*] coordinates {(852.2,0.85) (431.4,13.81) (333.1,21.32) (237.0,30.33) (134.7,34.47) (74.4,38.11)};
\addplot[armqtr, mark=triangle*] coordinates {(879.2,0.88) (437.4,14.00) (336.8,21.56) (234.4,30.00) (134.7,34.48) (76.5,39.18)};
\node[font=\tiny, anchor=south west, xshift=-3pt, yshift=3pt] at (axis cs:76.5,39.18) {$B{=}512$};
\node[font=\tiny, anchor=south, yshift=3pt] at (axis cs:879.2,0.88) {$B{=}1$};
\end{axis}
\end{tikzpicture}
\caption{EAGLE-3, GPT-OSS-20B.}
\label{fig:pareto-eg}
\end{subfigure}
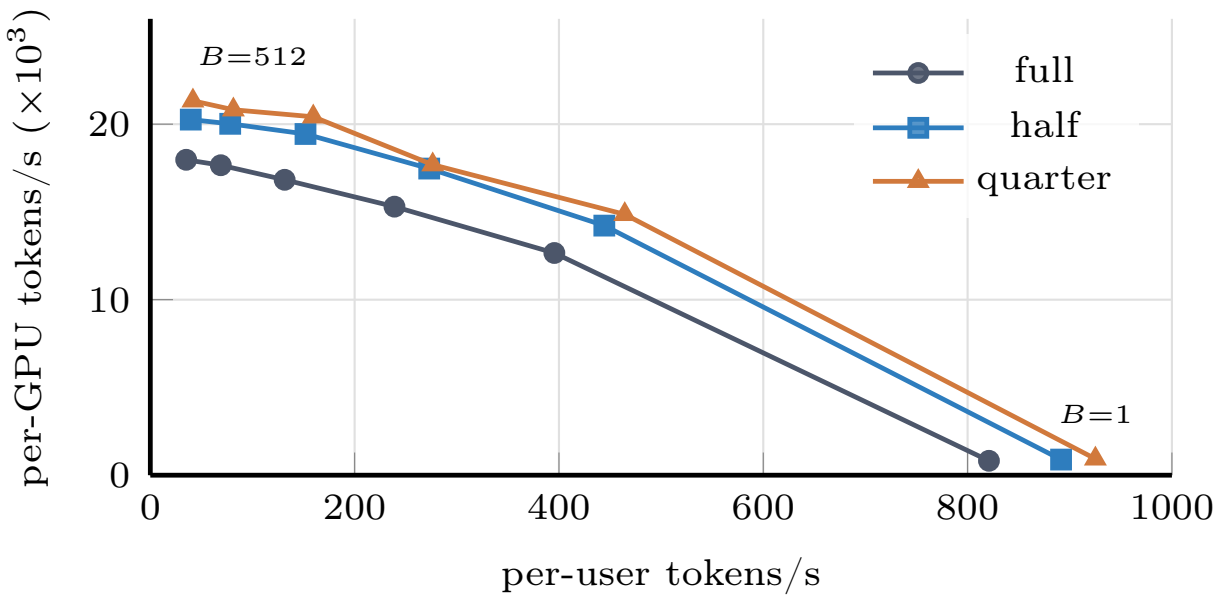
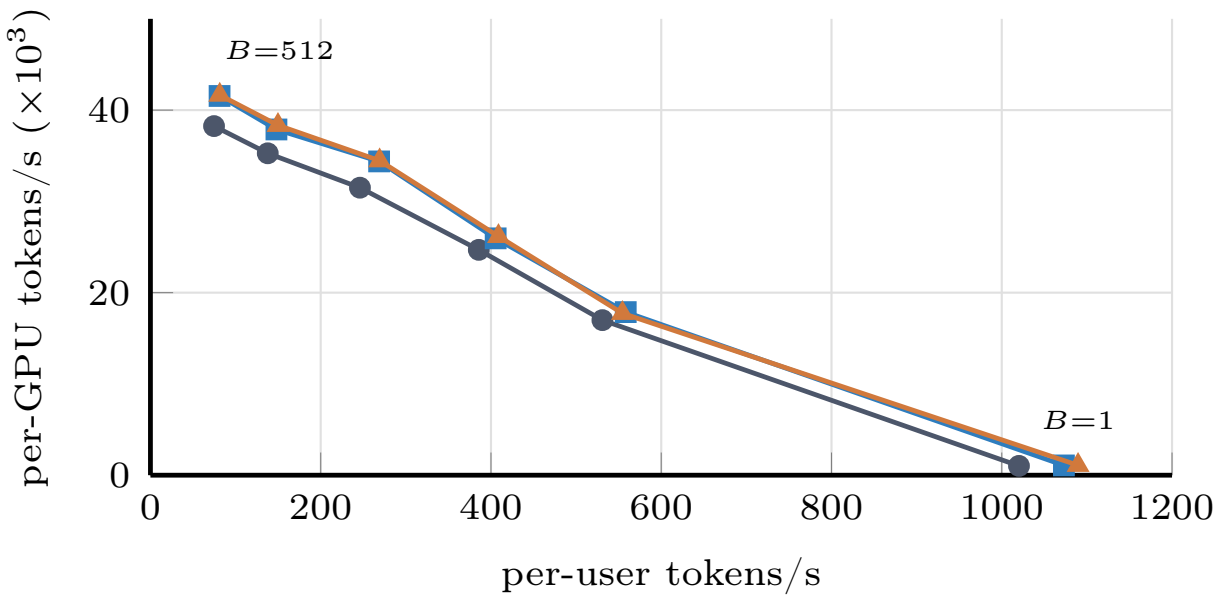
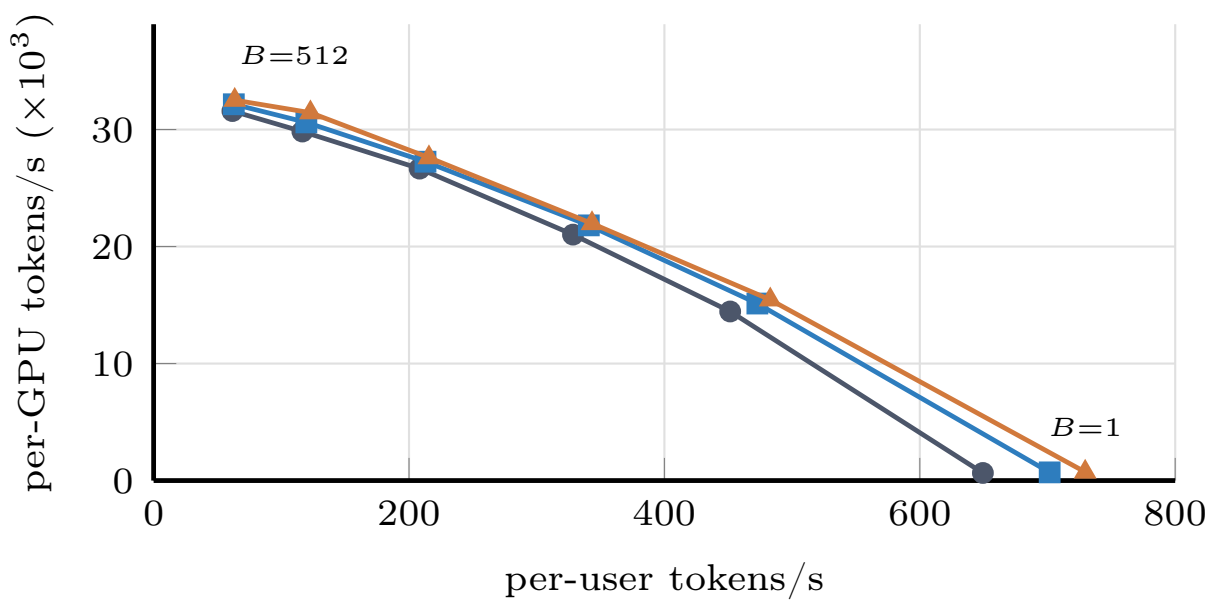
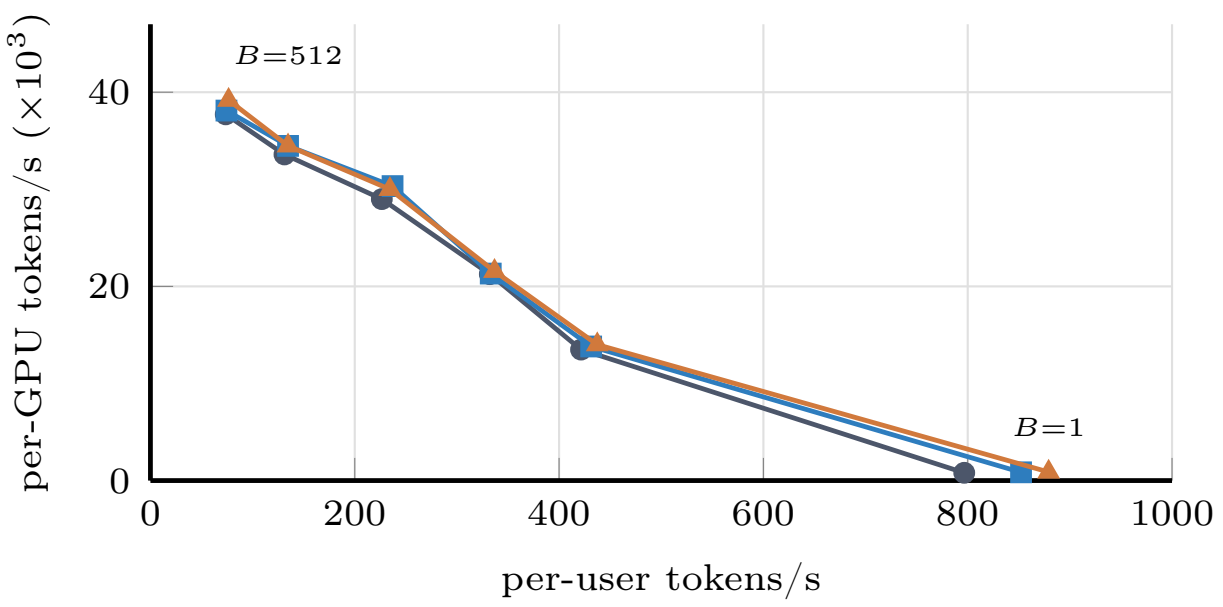
\caption{\textbf{Throughput across batch size\arxivgpu{ on B200}.} Each line is one
width, and its points are $B{=}1$, 32, 64, 128, 256 and 512 from right to left.
Per-user tokens/s is $1000/\TPOT$ with $\TPOT$ in ms, and per-GPU tokens/s is
$B$ times that. $\TPOT$ is that of Table~\ref{tab:batch}, full-batch
$\Titer$ over whole-run $\AL$ from the same run, on 512 SPEED-Bench throughput prompts.}
\label{fig:pareto}
\end{widefigure}

\textbf{Narrowing's throughput--latency Pareto frontier lies on or beyond full
width's at every batch size.}
Figure~\ref{fig:pareto} plots each variant's per-user and per-GPU throughput
across the sweep. In all four cells both narrow drafts sit at or beyond full
width at every batch size. The smallest gain, EAGLE-3-GPT-OSS at half width and
$B{=}64$ (1.00$\times$), is parity within its $\AL$ interval
(Table~\ref{tab:batchstages}). At a matched $B$ both axes improve by the same
factor, the $\TPOT$ speedup.

On GB10 the sweep runs to $B{=}256$ on Qwen3-8B, which does not fit at
$B{=}512$ there, and DFlash-Qwen gains 1.12--1.19$\times$ from $B{=}32$ to
256, close to B200 (Table~\ref{tab:gb10batch}, App.~\ref{app:gb10}).

\textbf{Context length barely moves the $\Titer$ benefit.} From 8k to 32k at
batch 1, no cell's $\Titer$ speedup changes by more than 1.6\% (Table~\ref{tab:ctx}). Longer
context adds KV-attention work to both stages, and attention is the one draft
operation that width does not shrink, so the benefit thins out only slowly.
On GPT-OSS-20B half of the target's layers use 128-token sliding-window
attention, so full-width $\Titer$ grows only 2.7--4.0\% over that range, against
8.1--8.7\% on Qwen3-8B. Appendix~\ref{app:scope} gives the full stage split
for every batch size and the sweep setup.

\section{Why the gain differs across GPUs}
\label{sec:hw}

The same drafts gain 1.06--1.12$\times$ on B200 and 1.13--1.44$\times$ on
GB10, and DFlash-Qwen gains 1.11$\times$ on H100. Verify barely moves with
width on any of them, so narrowing pays through the draft alone, by an amount
set by the draft's share of the iteration and by how much faster the narrow
draft runs. The draft's weights shrink in proportion to width, so a draft whose
time followed its bytes would run 2$\times$ faster at half width and
4$\times$ at quarter. How close a draft comes depends on how much of peak
bandwidth its GEMMs keep as they shrink (\S\ref{sec:gemmbw}). The quarter-width
draft runs 1.43--1.75$\times$ faster than full width on B200 and 3.44--3.87$\times$
on GB10. The draft's share is also higher on GB10 in three of four cells, but
DFlash-GPT-OSS has nearly the same share on both GPUs, 30.1\% and 28.9\%, and
still gains 1.07$\times$ on B200 and 1.20$\times$ on GB10 from its faster
draft alone.

\subsection{Draft GEMM bandwidth}
\label{sec:gemmbw}

Table~\ref{tab:opbw} (App.~\ref{app:gb10}) gives each DFlash-Qwen draft GEMM's
time per kernel and its BF16 weight bytes over that time, as a share of peak
bandwidth.
The vocabulary projection, at 1{,}245\,MB the largest GEMM, keeps 73--78\% of
peak at every width on B200 and 84--85\% on GB10. The smaller projections lose
bandwidth as they shrink. On B200, from full to quarter width, FC1 falls from
70\% to 56\%, FC2 from 69\% to 33\%, QKV from 60\% to 30\% and the output
projection from 48\% to 19\%. At quarter width each of their kernels runs
5--11\,$\mu$s, short enough that a fixed per-kernel cost is a large part of
it, while the vocabulary projection still runs 53\,$\mu$s. On GB10 the same
kernels run 46--226\,$\mu$s at quarter width and stay at 79--84\% of peak at
every width, and only the quarter-width output projection, at 8\,MB per kernel
the smallest GEMM, falls to 67\%. Over all five GEMMs,
B200 goes from 69\% of peak at full width to 47\% at quarter, and GB10 from
83\% to 81\%.

\subsection{H100}

We add DFlash on Qwen3-8B on one H100 (3.35\,TB/s HBM3) with the same drafts,
SGLang build and 880-prompt contract at temperature 0 and batch 1. NVFP4 does
not run on Hopper, so the target is the FP8 release of Qwen3-8B with FA3
attention and a BF16 KV cache, which changes the serving configuration as well
as the GPU (App.~\ref{app:h100}). Half width stays at acceptance parity and
quarter width costs 4.42\%, and both narrow drafts decode 1.11$\times$ faster
end to end (Table~\ref{tab:h100}), against 1.09$\times$ and 1.06$\times$ on
B200. The full-width draft takes 23.8\% of the iteration, and the
quarter-width draft runs 2.37$\times$ faster (Table~\ref{tab:h100l3}), between
B200's 1.75$\times$ and GB10's 3.57$\times$ on this cell.

\section{Related work}
\label{sec:related}

\paragraph{Draft families.}
Drafts range from separate small models~\citep{leviathan2023speculative} and
extra heads on the target~\citep{stern2018blockwise,cai2024medusa} to
feature-level autoregression (EAGLE)~\citep{li2024eagle,li2024eagle2,li2025eagle3},
block-parallel drafting~\citep{dflash2026} and built-in multi-token prediction
heads~\citep{gloeckle2024better,deepseekv3}. Work on these families mostly
changes how the draft predicts, to raise $\AL$. We keep each family's design
and change only its width.

\paragraph{Making each decode step cheaper.}
MagicDec~\citep{chen2024magicdec} observes that at long context and large
batch the bottleneck moves to loading the KV cache and drafts with a
fixed-size KV. LongSpec~\citep{yang2026longspec},
SpecExtend~\citep{cha2026specextend} and windowed
drafting~\citep{valliappan2026windowedmtp} likewise bound what the draft
attends to, and self-speculative methods draft with a subset of the target's
layers~\citep{zhang2024draftverify,elhoushi2024layerskip}. These shorten the
draft's context or depth. We change its \emph{feature} dimension, which is
independent of both.

\paragraph{Relation to vocabulary-projection and draft-shape work.}
FR-Spec~\citep{frspec2025} and VocabTrim~\citep{vocabtrim2025} shrink the
vocabulary projection to a fixed, frequency-ranked subset, and
NanoSpec~\citep{nanospec2026} to a per-request active set. These cut the
vocabulary dimension and narrowing cuts the hidden one, so the two are
potentially complementary. SlimSpec~\citep{slimspec2026} also acts on the
hidden dimension, through a low-rank bottleneck, but only in the projection,
while the draft's layers stay at full width (\S\ref{sec:whywidth}).
Narrowing shrinks the projection and every draft layer at once. EAGLE-3 already comes with a pruned draft vocabulary
(32{,}000 of 151{,}936 entries on Qwen3-8B, 64{,}000 of 201{,}088 on
GPT-OSS-20B), and narrowing its width still shrinks that projection further
(Table~\ref{tab:l2}). Folding, which lets DFlash's borrowed head narrow too, is a one-time merge at
load after the narrow draft is trained. Gemma~4~\citep{gemma4} recently shipped a co-trained
MTP drafter whose transformer is 256 or 1{,}024 wide, but its report fixes
that width without comparing alternatives or measuring what the choice costs
or buys.

\section{Discussion}
\label{sec:limits}

\paragraph{What a practitioner should take from this.}
These results suggest that target-width drafters can be overprovisioned for
the objective of minimizing decode latency at a matched training-data budget.
Draft hidden width is a design choice to make before training. Narrowing keeps
the draft's architecture and adds only width adapters at its boundary, plus
matching support in the serving engine. Which width to pick depends on the
GPU. On a datacenter GPU such as B200, halving is the default worth trying
first. On our panel it gives 1.06--1.09$\times$ end to end and costs no
acceptance in two of four cells, and 1.5\% and 3.0\% in the other two. On a lower-bandwidth GPU such as
GB10, quarter width is the fastest variant in every cell we test, at
1.20--1.44$\times$. Two things are worth checking. First, make
sure the vocabulary projection narrows with the body, by folding it if it is
borrowed (a draft that trains its own needs nothing). Second, tune the
learning rate and initialization at the new width instead of reusing the
full-width draft's. How much narrowing
pays depends on the draft's share of the iteration and on how close the narrow
draft's kernels stay to peak bandwidth (\S\ref{sec:hw}).

\paragraph{Composing with other methods.}
Width is a separate design axis from the other ways to make a decode step
cheaper and is potentially complementary to them. A narrow draft could also be
windowed or given a pruned vocabulary. We measure only the latter, in our
EAGLE-3 drafts.

\paragraph{Quantizing the draft.}
Quantization can complement narrowing, but its gain shrinks as the draft narrows. Both cut the
bytes the draft reads and the math it does, quantization by storing each
weight and activation in fewer bits and narrowing by shrinking the matrices
themselves. On
DFlash-Qwen at batch 1, NVFP4 W4A4 post-training quantization of the
full-width draft gives 1.07$\times$ end to end but loses 2.4\% $\AL$.
Halving the BF16 draft gives more, 1.09$\times$, and keeps $\AL$. The two compose at half
width, where the quantized draft is fastest at 1.11$\times$. At quarter width
quantization speeds up the iteration by only 1.02$\times$ and loses 3.4\%
$\AL$, so it is 0.99$\times$ against BF16 (App.~\ref{app:quant}). Other pairs,
large batches and quantization-aware training remain future work.

\paragraph{Limitations.}
Each variant is one training run with one seed at a matched data budget, not
at convergence, so other budgets could change what narrowing costs in
acceptance. We test two drafter families on two targets, and targets where the
draft is a small share of the iteration stand to gain less. Our
context sweep stops at 32k, and because narrowing does not shrink KV attention
we expect the gain to shrink at much longer contexts. The MLP-only ablation
uses untrained slices and compares latency only.

\section{Conclusion}

Full target-width residual streams are unnecessary for latency-efficient
feature-conditioned drafting in our evaluated settings. Narrowing the residual width
shrinks projections throughout the draft at once, and folding keeps a borrowed
vocabulary projection exact up to floating-point rounding. Across DFlash and
EAGLE-3 on two targets, narrowing the draft to half or a quarter of the
target's width speeds up end-to-end decoding by 1.06--1.12$\times$ on a B200
and by 1.13--1.44$\times$ on a DGX Spark (GB10), at temperature 0 on the 880-prompt
panel, for an acceptance cost of at most 3.0\% and 5.4\% on B200,
respectively. The speedup holds at larger batch sizes. From batch 32 to
the largest batch each GPU fits, narrowing reaches up to 1.21$\times$ on both
B200 and GB10, and no narrow draft is slower than full width. The gain grows with the
draft's share of the iteration and is largest where the draft's kernels stay
near peak bandwidth as they narrow. Anyone training a new feature-conditioned
drafter should pick its residual width on its own merits, and for the GPU it
will run on, rather than inherit it from the target.

\bibliography{refs}

\clearpage
\appendix
\section*{Appendix}
\suppressfloats[t]

\section{One DFlash draft layer at full and half width}
\label{app:overview}

Figure~\ref{fig:overview} expands one DFlash draft layer at full width and at half
width. Every projection reads or writes the residual stream, so all of them
shrink with it, while attention and the MLP's intermediate size keep theirs.
\begingroup
\providecommand{\moRes}[3]{%
  \fill[armhalf!18] (#1-0.075,#2-#3/2) rectangle (#1+0.075,#2+#3/2);
  \draw[armhalf, thin] (#1-0.075,#2-#3/2) rectangle (#1+0.075,#2+#3/2);}
\providecommand{\moNrm}[3]{%
  \fill[armhalf!55] (#1-0.025,#2-#3/2) rectangle (#1+0.025,#2+#3/2);}
\providecommand{\moInner}[4]{%
  \fill[accentgrey!22] (#1-#4/2,#2-#3/2) rectangle (#1+#4/2,#2+#3/2);
  \draw[accentgrey!90!black, thin] (#1-#4/2,#2-#3/2) rectangle (#1+#4/2,#2+#3/2);}
\providecommand{\moPC}{armhalf}
\providecommand{\moProj}[5]{%
  \fill[\moPC!16] (#1-#5,#2-#3/2) -- (#1+#5,#2-#4/2) -- (#1+#5,#2+#4/2) -- (#1-#5,#2+#3/2) -- cycle;
  \draw[\moPC, thin] (#1-#5,#2-#3/2) -- (#1+#5,#2-#4/2) -- (#1+#5,#2+#4/2) -- (#1-#5,#2+#3/2) -- cycle;}
\providecommand{\moPlus}[2]{\node[circle, draw, black!70, inner sep=0.4pt, font=\scriptsize] at (#1,#2) {$+$};}

\providecommand{\moRow}[4]{%
  \moRes{0}{#1}{#2}
  \moNrm{0.5}{#1}{#2}
  \moProj{1.35}{#1}{#2}{1.2}{0.32}
  \moInner{2.85}{#1}{1.2}{1.42}
  \node[font=\scriptsize, text=black!70, align=center] at (2.85,#1)
    {\textbf{Attention}};
  \moProj{4.35}{#1}{1.0}{#2}{0.32}
  \moPlus{5.1}{#1}
  \moNrm{5.6}{#1}{#2}
  \moProj{6.45}{#1}{#2}{1.4}{0.32}
  \moInner{7.75}{#1}{1.4}{0.15}
  \moProj{9.05}{#1}{1.4}{#2}{0.32}
  \moPlus{9.8}{#1}
  \moNrm{10.5}{#1}{#2}
  \moProj{11.4}{#1}{#2}{1.6}{0.45}
  \moInner{12.5}{#1}{1.6}{0.15}
  \foreach \a/\b in {0.08/0.47, 0.53/1.01, 1.69/2.12, 3.58/4.01, 4.69/4.98, 5.22/5.57, 5.63/6.11, 6.79/7.66, 7.84/8.71, 9.39/9.68, 9.92/10.47, 10.53/10.93, 11.87/12.41}
    \draw[ar] (\a,#1) -- (\b,#1);
  \draw[ar, rounded corners=2pt] (0,#1+#2/2) -- (0,#1+0.95) -| (5.1,#1+0.12);
  \draw[ar, rounded corners=2pt] (5.35,#1) -- (5.35,#1+0.95) -| (9.8,#1+0.12);
  \node[lb] at (0,#1-0.82)     {$X$\\$[\blk,#3]$};
  \node[lb] at (1.35,#1-0.82)  {\textbf{QKV}\\$#3{\to}d_{\mathrm{qkv}}$};
  \node[lb] at (4.35,#1-0.82)  {\textbf{O}\\$d_{\mathrm{attn}}{\to}#3$};
  \node[lb] at (6.45,#1-0.82)  {\textbf{gate/up}\\$#3{\to}2d_{\mathrm{ff}}$};
  \node[lb] at (7.75,#1-0.82)  {\textbf{SwiGLU}\\$2d_{\mathrm{ff}}{\to}d_{\mathrm{ff}}$};
  \node[lb] at (9.05,#1-0.82)  {\textbf{MLP down}\\$d_{\mathrm{ff}}{\to}#3$};
  \node[la] at (10.5,#1+#2/2+0.05) {norm};
  \node[lb] at (11.4,#1-0.9)   {#4};
  \node[la] at (12.5,#1+0.85)  {logits};
}

\begin{widefigure}
\centering
\resizebox{0.92\textwidth}{!}{%
\begin{tikzpicture}[
  x=1.2cm,
  font=\footnotesize,
  ar/.style={-{Stealth[length=1.5mm]}, thin, black!70},
  lb/.style={font=\scriptsize, text=black!70, align=center, anchor=north, inner sep=1pt},
  la/.style={font=\scriptsize, text=black!70, align=center, anchor=south, inner sep=1pt},
  rw/.style={font=\footnotesize\bfseries, anchor=east, align=right},
]
\draw[decorate, decoration={brace, amplitude=3pt}, black!55]
  (-0.15,1.1) -- node[above=2pt, font=\scriptsize, text=black!65] {$\times L$ draft layers in both rows} (10.0,1.1);
\renewcommand{\moPC}{armexact}
\moRow{0}{1.0}{\dt}{\textbf{vocab head}\\$[\vocab,\dt]$}
\node[rw, text=armexact] at (-1.7,0) {Full width\\$d=\dt$};

\renewcommand{\moPC}{armhalf}
\moRow{-2.75}{0.5}{\dd}{\textbf{vocab head}\\$[\vocab,\dd]$}
\node[rw, text=armhalf] at (-1.7,-2.75) {Half width\\$d=\dd=\dt/2$};
\moProj{-1.2}{-2.75}{1.0}{0.5}{0.26}
\draw[ar] (-0.92,-2.75) -- (-0.1,-2.75);
\node[lb] at (-1.2,-3.57) {\textbf{input adapter}\\$\dt{\to}\dd$};
\end{tikzpicture}}
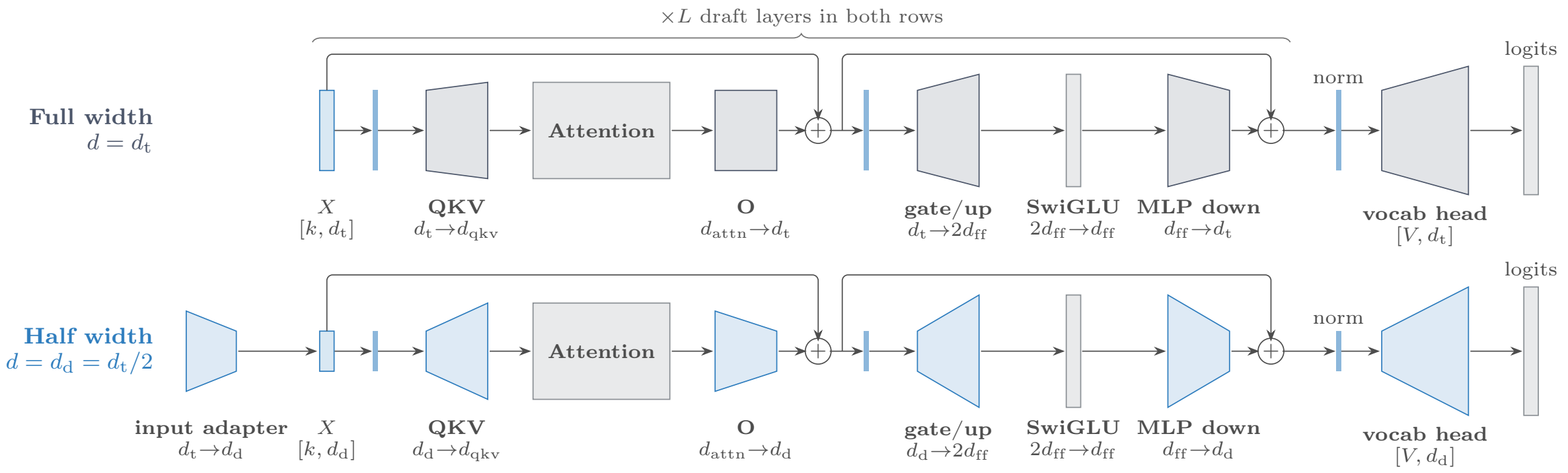
\caption{\textbf{One \arxivgpu{DFlash }draft layer at full and half width.} Blue tensors are the
residual stream at width $d$, drawn 2:1. Trapezoids are projections, and every
one of them reads or writes the residual, so all shrink with $d$. Grey parts
(attention over $d_{\mathrm{attn}}$, the MLP's $d_{\mathrm{ff}}$) keep their
size. The input adapter maps the target's $\dt$-wide inputs into the narrow
stream.\arxivgpu{ EAGLE-3 differs in three places. It has one layer that runs $\blk$
times, its QKV projection reads the $2d$ concatenation of token embedding and
hidden state, and it scores with its own head over a pruned vocabulary
(Figure~\ref{fig:path-e3}).}}
\label{fig:overview}
\end{widefigure}
\endgroup

\section{Folding: numerical check and precision}
\label{app:artifact}

We check Eq.~\ref{eq:fold} on our trained half-width Qwen3-8B output adapter
and the Qwen3-8B BF16 head, using 3{,}405 real post-norm draft hidden states
from 29 panel prompts that cover every block position the head scores. We
compare two BF16 paths with an FP64 reference. All three start from the same
stored BF16 draft states $\hd$, adapter $\Wout$ and head $\Wlm$, and
converting BF16 to FP64 is exact, so the three read identical inputs and
differ only in where they round.
\textbf{The FP64 reference} multiplies $\hd$ by $\Wout$ in FP64, keeps the
product $\hup$ in FP64, multiplies it by $\Wlm$ in FP64 and keeps the logits in
FP64. Nothing is rounded to BF16.
\textbf{The unfolded path} is what serving would run without folding. The
adapter GEMM multiplies $\hd$ by $\Wout$, accumulating in FP32, and writes
$\hup$ in BF16, which is the first rounding. It happens for every token at
every draft step. The head GEMM then multiplies this $\hup$ by $\Wlm$, again
accumulating in FP32, and writes the logits in BF16 before the draft picks its
proposal by argmax, which is the second.
\textbf{The folded path} is what we serve. At load, SGLang multiplies $\Wlm$
by $\Wout$ once with \texttt{torch.mm}, accumulating in FP32, and stores
$\Wfold$ in BF16, which is the first rounding. It happens once, to a weight,
not to each token's activation. At every draft step one GEMM multiplies $\hd$
by $\Wfold$, accumulating in FP32, and writes the logits in BF16 before the
draft picks its proposal by argmax, which is the second.
Both paths therefore round twice relative to the reference, and both share
the final rounding of the logits. Either rounding can flip the argmax where the
top two logits nearly tie.
With $Z$ a path's logits and $Z_{64}$ the reference, both over all positions
and the full 151{,}936-entry vocabulary, normalized RMS is
$\lVert Z - Z_{64}\rVert_F / \lVert Z_{64}\rVert_F$, and top-1 $=$ FP64 is
the share of positions whose argmax matches the reference. Folding trades the
per-token rounding of $\hup$ for a one-time rounding of $\Wfold$, and here
the trade leaves the folded path slightly closer to the reference
(Table~\ref{tab:foldcheck}).

\begin{center}
\begin{minipage}{\linewidth}
\centering
\captionof{table}{Folded and unfolded BF16 logits against an FP64 reference on 3{,}405 real draft states, trained half-width Qwen3-8B adapter.}
\label{tab:foldcheck}
\footnotesize
\setlength{\tabcolsep}{4pt}
\begin{tabular}{lrrr}
\toprule
BF16 path & max $|$err$|$ & norm.\ RMS & top-1 $=$ FP64 \\
\midrule
folded (served) & 0.251 & 0.20\% & 97.2\% \\
unfolded        & 0.270 & 0.21\% & 97.0\% \\
\bottomrule
\end{tabular}
\end{minipage}
\end{center}

Both paths miss the FP64 argmax at about 3\% of positions, and every miss is
a near-tie. Define the margin as the reference's top-1 minus top-2 logit. The
largest margin at any miss is 0.10 logits for the folded path and 0.12 for the
unfolded one, while the median margin over all positions is 0.51.
The two paths agree at 99.1\% of positions, so folding changes the draft's
proposals no more than BF16 rounding already does.
Serving the same half- and quarter-width DFlash-Qwen checkpoints both ways on
the 880-prompt panel at temperature 0, in two node-paired passes, confirms this
end to end. Every output was token-identical in these paired temperature-0
runs, and $\AL$ matches within 0.03\%, the residue of the near-tie
positions where the two heads propose different tokens. The unfolded path is 3.1\% and 4.7\% slower per iteration because it reads
the full $\vocab \times \dt$ head at every draft step. Folding cannot change
the output distribution at any temperature. Our drafts propose their argmax, so
verification scores each proposal with the target's own probabilities alone,
and folding can change only which token is proposed.

\subsection{Precision and memory}
\label{app:foldprecision}

Both targets keep \texttt{lm\_head} in BF16, so every fold in this paper is
computed from a BF16 head. The width study serves the BF16 fold, and
App.~\ref{app:quant} also quantizes it to NVFP4 after folding. Folding drops the per-token BF16 rounding of $\hup$ and adds a one-time BF16 rounding of $\Wfold$ at load. It
removes the adapter outright because nothing in DFlash but the head reads
$\hup$.

We do not evaluate folding an already quantized target head. To first order,
a 4-bit fold should be built from the unquantized head in FP32 and quantized
once, which keeps the single 4-bit weight rounding the unfolded head already has, and
this is what App.~\ref{app:quant} does. Folding a shipped 4-bit head would add
a second one. Either way $Q_4(\Wlm\Wout)$ is a different matrix from
$Q_4(\Wlm)\,\Wout$, and $\Wlm\Wout$ may quantize worse than $\Wlm$.
App.~\ref{app:quant} measures the end-to-end effect of the quantized fold but
does not compare it with the unfolded quantized computation.

Because the target keeps its own head for verification, the folded head is
extra resident weight. It replaces a $\dt \times \dd$ adapter with a
$\vocab \times \dd$ head, 0.017 against 0.622\,GB in BF16 on half-width
DFlash-Qwen (Table~\ref{tab:draftmem}), or 0.005 against 0.175\,GB when
App.~\ref{app:quant} stores it in NVFP4.
In exchange the draft no longer reads the $\vocab \times \dt$ head every
iteration. On DFlash-Qwen in BF16 the full-width draft reads the 1.24\,GB
target head once per iteration, while the folded half- and quarter-width
heads are 0.62 and 0.31\,GB, so each iteration reads 0.62 and 0.93\,GB less.
On the half-width draft the head GEMM drops from 206.2 to 99.4\,$\mu$s, about
6\,TB/s in both cases, so the GEMM is bound by these bytes. The price is the
extra 0.61\,GB of resident weight, about 0.3\% of one B200's memory.

Table~\ref{tab:draftmem} gives the net served weights of every draft, counting
only what each adds to the target. On DFlash narrowing always lowers it, to
0.80--0.87$\times$ of full width at half width and 0.40--0.44$\times$ at quarter,
because the folded head is smaller than the layers it removes. On EAGLE-3 the
folded embedding table makes the half-width draft larger than full width,
1.28$\times$ on Qwen3-8B and 1.55$\times$ on GPT-OSS-20B, because full width
reads the target's embedding directly and has a small layer stack to shrink.
Quarter width is smaller again, 0.64--0.78$\times$. Serving the narrow EAGLE-3
drafts unfolded avoids the table and brings half width to 0.42 and 0.28\,GB,
0.52$\times$ of full width on both targets. The price is an input adapter GEMM
at each of the seven draft steps, with the same $\dt \times \dd$ weight that
takes 4.2--7.2\,$\mu$s per call in DFlash (\S\ref{sec:method}).

\begin{widetable}
\centering
\footnotesize
\setlength{\tabcolsep}{4pt}
\caption{\textbf{Served draft weights per variant}, in GB ($10^9$ bytes) at BF16,
from the checkpoint tensor shapes. Only weights the draft adds to the target
are counted. \emph{Shared} and \emph{borrowed} mark the target's own embedding
and head, which cost the draft nothing. DFlash keeps its input adapter and replaces its output adapter with the
folded $\vocab \times \dd$ head. EAGLE-3 folds its input adapter into a
$\vocab \times \dd$ embedding table and keeps its own head. The unfolded total
serves the narrow draft with the target's head (DFlash) or embedding (EAGLE-3)
and the unfolded adapter instead.}
\label{tab:draftmem}
\begin{tabular}{lllrrrrrr}
\toprule
Family & Target & Draft & Layers $+$ \texttt{fc} & Input side & Output side & \textbf{Total} & vs full & Unfolded total \\
\midrule
\multirow{6}{*}{DFlash} & \multirow{3}{*}{Qwen3-8B}
   & full 4{,}096    & 2.097 & ---           & \emph{borrowed} & \textbf{2.097} & ---          & 2.097 \\
 & & half 2{,}048    & 1.049 & 0.017         & 0.622           & \textbf{1.688} & 0.80$\times$ & 1.082 \\
 & & quarter 1{,}024 & 0.524 & 0.008         & 0.311           & \textbf{0.844} & 0.40$\times$ & 0.541 \\
\cmidrule{2-9}
 & \multirow{3}{*}{GPT-OSS-20B}
   & full 2{,}880    & 1.570 & ---           & \emph{borrowed} & \textbf{1.570} & ---          & 1.570 \\
 & & half 1{,}440    & 0.785 & 0.008         & 0.579           & \textbf{1.372} & 0.87$\times$ & 0.801 \\
 & & quarter 720     & 0.392 & 0.004         & 0.290           & \textbf{0.686} & 0.44$\times$ & 0.401 \\
\midrule
\multirow{6}{*}{EAGLE-3} & \multirow{3}{*}{Qwen3-8B}
   & full 4{,}096    & 0.537 & \emph{shared} & 0.262 & \textbf{0.799} & ---            & 0.799 \\
 & & half 2{,}048    & 0.268 & 0.622         & 0.131 & \textbf{1.022} & 1.28$\times$   & 0.416 \\
 & & quarter 1{,}024 & 0.134 & 0.311         & 0.066 & \textbf{0.511} & 0.64$\times$   & 0.208 \\
\cmidrule{2-9}
 & \multirow{3}{*}{GPT-OSS-20B}
   & full 2{,}880    & 0.182 & \emph{shared} & 0.369 & \textbf{0.551} & ---            & 0.551 \\
 & & half 1{,}440    & 0.091 & 0.579         & 0.184 & \textbf{0.855} & 1.55$\times$   & 0.284 \\
 & & quarter 720     & 0.046 & 0.290         & 0.092 & \textbf{0.427} & 0.78$\times$   & 0.142 \\
\bottomrule
\end{tabular}
\end{widetable}

\paragraph{Multi-token prediction drafters.} Several recent open models ship
built-in multi-token prediction (MTP) layers as their drafter, among them
Qwen3.8-27B~\citep{qwen38} and Nemotron~3 Super~\citep{nemotron3super}. In
both checkpoints the MTP layer has no embedding or vocabulary projection of
its own. A projection maps the target's token embedding and hidden state,
concatenated, to the layer's width, and the layer's output goes through the
target's full-vocabulary head. Drafting is autoregressive, one MTP pass per
proposal. These drafters therefore combine the two cases we study. Like DFlash
they borrow the full-vocabulary head, so folding would narrow it, and like
EAGLE-3 they draft one token per step, so at small batch they read the whole
$\vocab \times \dt$ head $\blk$ times per iteration, 2.5\,GB per step in BF16
on Qwen3.8-27B. Their input projection already reads the embedding, so it
narrows with no extra adapter, as DFlash's \texttt{fc} does. We therefore
expect the vocabulary projection's share of draft time, and with it the saving
from narrowing, to be larger than in either drafter we measure. MTP layers are
usually trained with the target. Training
and evaluating narrow MTP drafters is future work.

\section{Training details}
\label{app:train}

\paragraph{Reported checkpoints.} All variants of a cell report the same epoch.
DFlash-Qwen reports epoch 5 of 6, where held-out loss on corpus samples never
trained on is lowest for every variant, and where half width's lead over full width
is smaller than at epoch 6. Every other cell reports its final checkpoint,
epoch 6 for DFlash-GPT-OSS and epoch 3 for EAGLE-3.
The $\AL$ ordering of the widths at the reported checkpoint holds at every epoch in three
cells. On DFlash-Qwen quarter width is last at every epoch and half width leads
full width from epoch 4 onwards (Table~\ref{tab:ordering}).

\begin{center}
\begin{minipage}{\linewidth}
\centering
\captionof{table}{Reported epoch and the $\AL$ ordering of the widths at every epoch.}
\label{tab:ordering}
\footnotesize
\setlength{\tabcolsep}{3pt}
\begin{tabular}{lcl}
\toprule
Cell & Reported & $\AL$ ordering at every epoch \\
\midrule
DFlash-Qwen     & ep.\ 5 of 6 & qtr last; half $>$ full from ep.\ 4 \\
DFlash-GPT-OSS  & ep.\ 6 of 6 & full $>$ half $>$ qtr \\
EAGLE-3-Qwen    & ep.\ 3 of 3 & full $>$ half $>$ qtr \\
EAGLE-3-GPT-OSS & ep.\ 3 of 3 & half $>$ full $>$ qtr \\
\bottomrule
\end{tabular}
\end{minipage}
\end{center}

\sharedtablehere{\begin{widetable}
\centering
\footnotesize
\caption{\textbf{Learning rate and initialization per variant.}}
\label{tab:lr}
\begin{tabular}{lllll}
\toprule
Family & Target & Draft & \textbf{Learning rate} & \textbf{\texttt{initializer\_range}} \\
\midrule
\emph{DFlash} & \emph{both} & \emph{every width} & \emph{6e-4} & \emph{default} \\
\midrule
\multirow{3}{*}{EAGLE-3} & \multirow{3}{*}{Qwen3-8B}
 & full 4{,}096   & \textbf{2e-4} & \textbf{0.005} \\
& & half 2{,}048    & \textbf{4e-4} & \textbf{0.01} \\
& & quarter 1{,}024 & \textbf{8e-4} & \textbf{0.02} \\
\midrule
\multirow{3}{*}{EAGLE-3} & \multirow{3}{*}{GPT-OSS-20B}
 & full 2{,}880 & \textbf{3e-4} & \textbf{0.007} \\
& & half 1{,}440  & \textbf{6e-4} & \textbf{0.02} \\
& & quarter 720   & \textbf{1.2e-3} & \textbf{0.028} \\
\bottomrule
\end{tabular}
\end{widetable}
}

\paragraph{Optimizer, batch and learning rate.} Every variant uses AdamW, a cosine
schedule with 4\% warmup, gradient clipping at 1.0 and sequences of up to
3{,}072 tokens. DFlash uses a global batch of 16 on Qwen3-8B (micro-batch 4 per
GPU) and 64 on GPT-OSS-20B (micro-batch 8, two accumulation steps). EAGLE-3
uses a global batch of 64 on both targets (micro-batch 4, four accumulation
steps). DFlash runs a learning rate of 6e-4 at every width. EAGLE-3's learning rate doubles
with each halving of width, and \texttt{initializer\_range} scales with it
except at half width on GPT-OSS-20B, which keeps 0.02 (Table~\ref{tab:lr}).
Sweeping the gradient-clipping norm moved acceptance by at most 0.008, against
a 0.045 width gap.

\paragraph{GPT-OSS-20B.} Its chat template inserts the current date into the
system prompt, so the same weights score up to 0.8\% different $\AL$ on
different days. Every reported comparison across width variants on this target
uses variants served on the same day.

\section{Temperature 1, per-benchmark speedups, categories and released drafters}
\label{app:t1}

At temperature 0 greedy verification accepts only the target's argmax, so
every variant emits the target's greedy output up to numerical nondeterminism. At
temperature 1 rejection sampling preserves the target's distribution, but each
variant draws a different sample from it, because which tokens are accepted
depends on the draft's proposals. Acceptance differences therefore also carry
sampling noise, and we read Tables~\ref{tab:t1}
and~\ref{tab:bench8t1} as robustness checks. Table~\ref{tab:bench8tpot} gives
the per-benchmark $\TPOT$ speedups behind \S\ref{sec:results}.
Tables~\ref{tab:percat0} and~\ref{tab:percat1} give $\AL$ on each
SPEED-Bench category of our panel.

\sharedtablehere{\begin{widetable}
\centering
\footnotesize
\setlength{\tabcolsep}{4pt}
\caption{\textbf{Temperature 1, all four cells\arxivgpu{, on B200}.} Same panel and contract as
Table~\ref{tab:headline}. Bold marks each cell's best $\TPOT$ speedup. No
$\TPOT$ speedup moves by more than 0.03$\times$ from its temperature-0 value. Each variant samples its own text, so
read acceptance here as a robustness check (App.~\ref{app:t1}).}
\label{tab:t1}
\begin{tabular}{llrrrr}
\toprule
Family / target & Draft & $\AL$ & $\Titer$ (ms) & $\TPOT$ (ms) & \textbf{\boldmath$\TPOT$ speedup} \\
\midrule
\multirow{3}{*}{DFlash, Qwen3-8B}
 & full 4{,}096 & 2.818 & 3.724 & 1.326 & 1.00$\times$ \\
 & half    & 2.825 & 3.459 & 1.229 & 1.08$\times$ \\
 & quarter & 2.743 & 3.341 & 1.222 & \textbf{1.09$\times$} \\
\midrule
\multirow{3}{*}{DFlash, GPT-OSS-20B}
 & full 2{,}880 & 2.285 & 3.462 & 1.520 & 1.00$\times$ \\
 & half    & 2.266 & 3.258 & 1.442 & 1.05$\times$ \\
 & quarter & 2.219 & 3.144 & 1.421 & \textbf{1.07$\times$} \\
\midrule
\multirow{3}{*}{EAGLE-3, Qwen3-8B}
 & full 4{,}096 & 2.650 & 4.296 & 1.620 & 1.00$\times$ \\
 & half           & 2.571 & 3.863 & 1.503 & 1.08$\times$ \\
 & quarter        & 2.560 & 3.693 & 1.442 & \textbf{1.12$\times$} \\
\midrule
\multirow{3}{*}{EAGLE-3, GPT-OSS-20B}
 & full 2{,}880 & 2.085 & 3.743 & 1.801 & 1.00$\times$ \\
 & half           & 2.103 & 3.509 & 1.670 & 1.08$\times$ \\
 & quarter        & 2.044 & 3.365 & 1.647 & \textbf{1.09$\times$} \\
\bottomrule
\end{tabular}
\end{widetable}
}
\sharedtablehere{\begin{widetable}
\centering
\footnotesize
\setlength{\tabcolsep}{3pt}
\caption{\textbf{Per-category acceptance at temperature 0\arxivgpu{ on B200}.} Each of the 11
SPEED-Bench categories of 80 prompts in our 880-prompt panel. \emph{Full} is the
full-width draft's $\AL$, and half and quarter are $\Delta\AL$ in \% relative to it.
The pooled row matches Table~\ref{tab:headline}.}
\label{tab:percat0}
\begin{tabular}{lrrr@{\hskip 1em}rrr@{\hskip 1em}rrr@{\hskip 1em}rrr}
\toprule
 & \multicolumn{6}{c}{\textbf{DFlash}} & \multicolumn{6}{c}{\textbf{EAGLE-3}} \\
\cmidrule(lr){2-7}\cmidrule(lr){8-13}
 & \multicolumn{3}{c}{Qwen3-8B} & \multicolumn{3}{c}{GPT-OSS-20B} & \multicolumn{3}{c}{Qwen3-8B} & \multicolumn{3}{c}{GPT-OSS-20B} \\
\cmidrule(lr){2-4}\cmidrule(lr){5-7}\cmidrule(lr){8-10}\cmidrule(lr){11-13}
 & full & half & qtr & full & half & qtr & full & half & qtr & full & half & qtr \\
Category & $\AL$ & $\Delta$\% & $\Delta$\% & $\AL$ & $\Delta$\% & $\Delta$\% & $\AL$ & $\Delta$\% & $\Delta$\% & $\AL$ & $\Delta$\% & $\Delta$\% \\
\midrule
coding         & 6.123 & $+0.9$ & $-9.6$ & 4.129 & $-2.1$ & $-4.2$ & 4.907 & $-4.2$ & $-4.4$ & 3.636 & $-0.2$ & $-3.8$ \\
math           & 6.174 & $+1.5$ & $-4.6$ & 3.951 & $-1.9$ & $-4.0$ & 4.613 & $-3.3$ & $-3.0$ & 3.369 & $+2.6$ & $-1.6$ \\
stem           & 5.317 & $+1.0$ & $-4.9$ & 4.132 & $-2.2$ & $-4.3$ & 4.252 & $-2.6$ & $-3.5$ & 3.514 & $+1.8$ & $-1.6$ \\
humanities     & 4.934 & $+1.3$ & $-4.7$ & 4.864 & $-1.0$ & $-2.9$ & 4.038 & $-2.7$ & $-3.8$ & 4.044 & $+3.7$ & $+0.2$ \\
multilingual   & 4.389 & $+1.7$ & $-10.7$ & 4.335 & $-1.9$ & $-3.5$ & 2.343 & $-2.0$ & $-1.6$ & 3.164 & $-0.2$ & $-1.4$ \\
reasoning      & 4.194 & $+0.9$ & $-2.8$ & 4.522 & $-0.3$ & $-1.9$ & 3.755 & $-3.5$ & $-4.4$ & 3.744 & $+3.5$ & $-1.7$ \\
rag            & 3.939 & $-0.1$ & $-14.8$ & 4.691 & $-3.0$ & $-4.1$ & 3.498 & $-4.4$ & $-4.5$ & 3.821 & $+2.9$ & $-2.7$ \\
qa             & 3.567 & $-0.3$ & $-1.3$ & 5.620 & $-1.0$ & $-2.2$ & 3.496 & $-4.9$ & $-5.9$ & 4.946 & $+1.8$ & $-0.2$ \\
writing        & 2.979 & $+0.6$ & $-5.1$ & 3.891 & $-1.6$ & $-4.3$ & 2.820 & $-2.5$ & $-4.2$ & 3.334 & $+2.1$ & $-3.6$ \\
summarization  & 2.783 & $+1.1$ & $-0.2$ & 3.062 & $-1.2$ & $-3.8$ & 2.582 & $-2.2$ & $-4.0$ & 2.537 & $-0.9$ & $-1.5$ \\
roleplay       & 2.496 & $-0.2$ & $-2.5$ & 4.641 & $-1.2$ & $-3.7$ & 3.260 & $-2.7$ & $-3.2$ & 4.147 & $+1.3$ & $-5.6$ \\
\midrule
\textbf{pooled}& 4.298 & $+0.8$ & $-5.4$ & 4.376 & $-1.5$ & $-3.6$ & 3.583 & $-3.0$ & $-3.7$ & 3.705 & $+2.1$ & $-2.1$ \\
\bottomrule
\end{tabular}
\end{widetable}
}
\sharedtablehere{\begin{widetable}
\centering
\footnotesize
\setlength{\tabcolsep}{4pt}
\caption{\textbf{Per-category $\TPOT$ speedup over full width at temperature 0\arxivgpu{ on B200}},
from the same runs as Tables~\ref{tab:headline} and~\ref{tab:percat0}. The
pooled row matches Table~\ref{tab:headline}.}
\label{tab:percattpot}
\begin{tabular}{lrr@{\hskip 1em}rr@{\hskip 1em}rr@{\hskip 1em}rr}
\toprule
 & \multicolumn{4}{c}{\textbf{DFlash}} & \multicolumn{4}{c}{\textbf{EAGLE-3}} \\
\cmidrule(lr){2-5}\cmidrule(lr){6-9}
 & \multicolumn{2}{c}{Qwen3-8B} & \multicolumn{2}{c}{\gptosshead} & \multicolumn{2}{c}{Qwen3-8B} & \multicolumn{2}{c}{\gptosshead} \\
\cmidrule(lr){2-3}\cmidrule(lr){4-5}\cmidrule(lr){6-7}\cmidrule(lr){8-9}
Category & half & qtr & half & qtr & half & qtr & half & qtr \\
\midrule
coding         & 1.09 & 1.02 & 1.05 & 1.05 & 1.07 & 1.12 & 1.07 & 1.08 \\
math           & 1.10 & 1.07 & 1.05 & 1.07 & 1.07 & 1.09 & 1.10 & 1.11 \\
stem           & 1.09 & 1.06 & 1.05 & 1.07 & 1.09 & 1.14 & 1.08 & 1.10 \\
humanities     & 1.09 & 1.07 & 1.06 & 1.09 & 1.08 & 1.13 & 1.11 & 1.12 \\
multilingual   & 1.10 & 1.00 & 1.04 & 1.05 & 1.07 & 1.10 & 1.08 & 1.10 \\
reasoning      & 1.09 & 1.09 & 1.08 & 1.10 & 1.09 & 1.13 & 1.10 & 1.09 \\
rag            & 1.08 & 0.95 & 1.03 & 1.06 & 1.06 & 1.11 & 1.10 & 1.07 \\
qa             & 1.08 & 1.11 & 1.06 & 1.09 & 1.08 & 1.11 & 1.10 & 1.11 \\
writing        & 1.09 & 1.07 & 1.06 & 1.07 & 1.10 & 1.11 & 1.10 & 1.09 \\
summarization  & 1.09 & 1.12 & 1.07 & 1.07 & 1.08 & 1.11 & 1.07 & 1.12 \\
roleplay       & 1.07 & 1.10 & 1.06 & 1.07 & 1.09 & 1.13 & 1.09 & 1.05 \\
\midrule
\textbf{pooled}& 1.09 & 1.06 & 1.06 & 1.07 & 1.08 & 1.12 & 1.09 & 1.09 \\
\bottomrule
\end{tabular}
\end{widetable}
}
\sharedtablehere{\begin{widetable}
\centering
\footnotesize
\setlength{\tabcolsep}{3pt}
\caption{\textbf{Per-category acceptance at temperature 1\arxivgpu{ on B200}.} Same panel and
layout as Table~\ref{tab:percat0}. The pooled row matches Table~\ref{tab:t1}.
Each variant samples its own text, so per-category changes also carry sampling
noise (App.~\ref{app:t1}).}
\label{tab:percat1}
\begin{tabular}{lrrr@{\hskip 1em}rrr@{\hskip 1em}rrr@{\hskip 1em}rrr}
\toprule
 & \multicolumn{6}{c}{\textbf{DFlash}} & \multicolumn{6}{c}{\textbf{EAGLE-3}} \\
\cmidrule(lr){2-7}\cmidrule(lr){8-13}
 & \multicolumn{3}{c}{Qwen3-8B} & \multicolumn{3}{c}{GPT-OSS-20B} & \multicolumn{3}{c}{Qwen3-8B} & \multicolumn{3}{c}{GPT-OSS-20B} \\
\cmidrule(lr){2-4}\cmidrule(lr){5-7}\cmidrule(lr){8-10}\cmidrule(lr){11-13}
 & full & half & qtr & full & half & qtr & full & half & qtr & full & half & qtr \\
Category & $\AL$ & $\Delta$\% & $\Delta$\% & $\AL$ & $\Delta$\% & $\Delta$\% & $\AL$ & $\Delta$\% & $\Delta$\% & $\AL$ & $\Delta$\% & $\Delta$\% \\
\midrule
coding         & 3.939 & $-1.7$ & $-2.4$ & 2.887 & $-2.4$ & $-3.5$ & 3.719 & $-5.3$ & $-6.6$ & 2.639 & $-2.0$ & $-3.4$ \\
math           & 3.245 & $+0.2$ & $-2.6$ & 2.479 & $-0.8$ & $-1.7$ & 2.994 & $-1.0$ & $-1.8$ & 2.295 & $0.0$ & $-2.2$ \\
stem           & 2.890 & $-1.9$ & $-3.0$ & 2.302 & $-1.3$ & $-2.9$ & 2.684 & $0.0$ & $-4.0$ & 2.100 & $+1.0$ & $-1.1$ \\
humanities     & 2.704 & $+2.1$ & $-2.6$ & 2.138 & $-2.2$ & $-3.2$ & 2.570 & $-3.0$ & $-0.1$ & 1.939 & $+1.2$ & $-1.0$ \\
multilingual   & 2.713 & $+5.9$ & $+4.5$ & 2.815 & $-1.8$ & $-1.4$ & 2.151 & $-11.3$ & $-0.7$ & 2.296 & $+0.8$ & $-1.7$ \\
reasoning      & 3.204 & $-1.1$ & $-1.6$ & 2.409 & $+0.6$ & $-1.7$ & 3.003 & $-4.4$ & $-3.3$ & 2.203 & $+1.3$ & $-4.5$ \\
rag            & 3.255 & $+1.4$ & $-11.9$ & 2.443 & $+1.8$ & $+0.9$ & 3.017 & $-6.8$ & $-4.3$ & 2.204 & $-0.9$ & $-1.7$ \\
qa             & 2.469 & $-1.2$ & $-3.2$ & 2.057 & $-0.4$ & $-3.8$ & 2.420 & $-4.9$ & $-5.5$ & 1.936 & $-0.4$ & $-1.0$ \\
writing        & 2.331 & $0.0$ & $-3.6$ & 2.040 & $-0.3$ & $-3.2$ & 2.269 & $-3.5$ & $-4.4$ & 1.870 & $+3.1$ & $-1.3$ \\
summarization  & 2.657 & $+0.6$ & $-1.8$ & 2.683 & $-4.0$ & $-7.2$ & 2.500 & $-3.9$ & $-4.2$ & 2.241 & $-0.9$ & $-1.5$ \\
roleplay       & 2.016 & $-0.2$ & $-0.9$ & 1.811 & $0.0$ & $-2.7$ & 2.096 & $+14.7$ & $-5.8$ & 1.691 & $+1.8$ & $-1.7$ \\
\midrule
\textbf{pooled}& 2.818 & $+0.2$ & $-2.7$ & 2.285 & $-0.8$ & $-2.9$ & 2.650 & $-3.0$ & $-3.4$ & 2.085 & $+0.9$ & $-2.0$ \\
\bottomrule
\end{tabular}
\end{widetable}
}
\sharedtablehere{\begin{widetable}
\centering
\footnotesize
\setlength{\tabcolsep}{3.5pt}
\caption{Eight-benchmark acceptance length at temperature 1\arxivgpu{ on B200}, setup as in
Figure~\ref{fig:bench8}. Pooled, half width is within 1.6\% of full width in
three cells and costs 3.8\% on EAGLE-3-Qwen, and quarter width costs
0.4--3.8\%. Single benchmarks are noisy under sampling.}
\label{tab:bench8t1}
\begin{tabular}{lrrr@{\hskip 1.0em}rrr@{\hskip 1.0em}rrr@{\hskip 1.0em}rrr}
\toprule
& \multicolumn{6}{c}{\textbf{DFlash}} & \multicolumn{6}{c}{\textbf{EAGLE-3}} \\
\cmidrule(lr){2-7}\cmidrule(lr){8-13}
& \multicolumn{3}{c}{Qwen3-8B} & \multicolumn{3}{c}{GPT-OSS-20B} & \multicolumn{3}{c}{Qwen3-8B} & \multicolumn{3}{c}{GPT-OSS-20B} \\
\cmidrule(lr){2-4}\cmidrule(lr){5-7}\cmidrule(lr){8-10}\cmidrule(lr){11-13}
Benchmark & full & half & qtr & full & half & qtr & full & half & qtr & full & half & qtr \\
\midrule
gsm8k         & 4.840 & 5.034 & 4.841 & 3.072 & 3.214 & 3.117 & 4.265 & 4.037 & 4.099 & 2.862 & 2.851 & 2.788 \\
math500       & 5.003 & 5.033 & 5.010 & 3.210 & 3.213 & 3.176 & 4.529 & 4.292 & 4.249 & 3.032 & 3.022 & 2.984 \\
aime25        & 3.769 & 3.651 & 3.758 & 2.782 & 2.670 & 2.721 & 3.509 & 3.511 & 3.591 & 2.568 & 2.505 & 2.535 \\
humaneval     & 6.000 & 6.018 & 5.811 & 2.961 & 2.991 & 2.940 & 4.984 & 4.799 & 4.820 & 2.811 & 2.755 & 2.753 \\
mbpp          & 4.513 & 4.779 & 4.523 & 3.057 & 3.065 & 3.076 & 4.069 & 3.875 & 3.934 & 2.936 & 2.910 & 2.825 \\
livecodebench & 3.497 & 3.558 & 3.515 & 2.471 & 2.499 & 2.448 & 3.490 & 3.418 & 3.393 & 2.320 & 2.274 & 2.267 \\
mtbench       & 2.868 & 2.901 & 2.865 & 2.354 & 2.297 & 2.262 & 2.857 & 2.674 & 2.676 & 2.239 & 2.190 & 2.150 \\
alpaca        & 2.630 & 2.586 & 2.559 & 2.278 & 2.285 & 2.237 & 2.666 & 2.537 & 2.516 & 2.183 & 2.137 & 2.107 \\
\midrule
pooled        & 3.812 & 3.844 & 3.798 & 2.695 & 2.692 & 2.654 & 3.620 & 3.483 & 3.484 & 2.546 & 2.505 & 2.478 \\
\bottomrule
\end{tabular}
\end{widetable}
}
\sharedtablehere{\begin{widetable}
\centering
\footnotesize
\setlength{\tabcolsep}{4pt}
\caption{\textbf{$\TPOT$ speedup over full width on each of the eight
benchmarks\arxivgpu{ on B200}}, from the same runs as Figure~\ref{fig:bench8} and
Table~\ref{tab:bench8t1}. Pooled is over every decode step of the
suite. Every benchmark is faster at both temperatures. The smallest
speedup is 1.02$\times$ at temperature 1 (DFlash-GPT-OSS half width on aime25)
and 1.05$\times$ at temperature 0.}
\label{tab:bench8tpot}
\begin{tabular}{lrr@{\hskip 1.2em}rr@{\hskip 1.2em}rr@{\hskip 1.2em}rr}
\toprule
 & \multicolumn{4}{c}{\textbf{DFlash}} & \multicolumn{4}{c}{\textbf{EAGLE-3}} \\
\cmidrule(lr){2-5}\cmidrule(lr){6-9}
 & \multicolumn{2}{c}{Qwen3-8B} & \multicolumn{2}{c}{\gptosshead} & \multicolumn{2}{c}{Qwen3-8B} & \multicolumn{2}{c}{\gptosshead} \\
\cmidrule(lr){2-3}\cmidrule(lr){4-5}\cmidrule(lr){6-7}\cmidrule(lr){8-9}
Benchmark & half & qtr & half & qtr & half & qtr & half & qtr \\
\midrule
\multicolumn{9}{l}{\emph{Temperature 0}} \\
gsm8k         & 1.10 & 1.11 & 1.06 & 1.06 & 1.05 & 1.11 & 1.06 & 1.10 \\
math500       & 1.10 & 1.12 & 1.06 & 1.06 & 1.05 & 1.10 & 1.07 & 1.09 \\
aime25        & 1.10 & 1.11 & 1.06 & 1.06 & 1.06 & 1.10 & 1.07 & 1.11 \\
humaneval     & 1.08 & 1.05 & 1.06 & 1.05 & 1.07 & 1.12 & 1.05 & 1.08 \\
mbpp          & 1.09 & 1.06 & 1.07 & 1.07 & 1.06 & 1.12 & 1.07 & 1.09 \\
livecodebench & 1.09 & 1.10 & 1.05 & 1.07 & 1.07 & 1.12 & 1.05 & 1.06 \\
mtbench       & 1.09 & 1.10 & 1.06 & 1.06 & 1.06 & 1.09 & 1.05 & 1.07 \\
alpaca        & 1.09 & 1.07 & 1.06 & 1.05 & 1.05 & 1.08 & 1.06 & 1.08 \\
\cmidrule(lr){2-9}
pooled        & 1.09 & 1.09 & 1.06 & 1.06 & 1.06 & 1.10 & 1.06 & 1.08 \\
\midrule
\multicolumn{9}{l}{\emph{Temperature 1}} \\
gsm8k         & 1.13 & 1.12 & 1.10 & 1.11 & 1.05 & 1.11 & 1.10 & 1.12 \\
math500       & 1.09 & 1.12 & 1.05 & 1.07 & 1.04 & 1.09 & 1.09 & 1.12 \\
aime25        & 1.04 & 1.11 & 1.02 & 1.06 & 1.12 & 1.19 & 1.06 & 1.13 \\
humaneval     & 1.09 & 1.08 & 1.09 & 1.09 & 1.08 & 1.12 & 1.08 & 1.13 \\
mbpp          & 1.15 & 1.12 & 1.05 & 1.09 & 1.05 & 1.13 & 1.09 & 1.10 \\
livecodebench & 1.09 & 1.10 & 1.07 & 1.10 & 1.08 & 1.12 & 1.07 & 1.12 \\
mtbench       & 1.11 & 1.12 & 1.04 & 1.05 & 1.04 & 1.08 & 1.07 & 1.11 \\
alpaca        & 1.07 & 1.09 & 1.06 & 1.07 & 1.05 & 1.10 & 1.07 & 1.11 \\
\cmidrule(lr){2-9}
pooled        & 1.09 & 1.11 & 1.06 & 1.08 & 1.07 & 1.12 & 1.08 & 1.12 \\
\bottomrule
\end{tabular}
\end{widetable}
}
\sharedtablehere{\begin{widetable}
\centering
\footnotesize
\setlength{\tabcolsep}{4pt}
\caption{Released full-width drafters on the same panel and contract\arxivgpu{, on B200}. Each
percentage is our full-width draft relative to the release, so positive means
ours is ahead. The releases share our full-width architecture
(App.~\ref{app:stockarch}) but not our data or budget, so a gap reflects
training rather than width. Bold marks the two cells where the release accepts
more than ours by over 1\%.}
\label{tab:stock}
\begin{tabular}{lllrrrr}
\toprule
 & & & \multicolumn{2}{c}{$T{=}0$} & \multicolumn{2}{c}{$T{=}1$} \\
\cmidrule(lr){4-5}\cmidrule(lr){6-7}
Family & Target & Reference & ref.\ $\AL$ & \textbf{ours vs ref.} & ref.\ $\AL$ & \textbf{ours vs ref.} \\
\midrule
DFlash  & Qwen3-8B    & \texttt{z-lab}    & 4.304 & $-0.1\%$ & 2.835 & $-0.6\%$ \\
DFlash  & GPT-OSS-20B & \texttt{z-lab}    & 4.309 & $+1.6\%$ & 2.332 & \textbf{\textminus2.0\%} \\
EAGLE-3 & Qwen3-8B    & \texttt{RedHatAI} & 2.846 & $+25.9\%$ & 2.186 & $+21.2\%$ \\
EAGLE-3 & GPT-OSS-20B & \texttt{RedHatAI} & 3.413 & $+8.6\%$  & 2.115 & \textbf{\textminus1.4\%} \\
\bottomrule
\end{tabular}
\end{widetable}
}

The largest category cost, quarter-width DFlash-Qwen on RAG ($-14.8\%$, with a
95\% paired-bootstrap interval of $-22.2$ to $-6.3\%$), comes from 7 of the
category's 80 prompts, those whose prompt and answer together exceed the
3{,}072-token training sequences (App.~\ref{app:train}). Quarter width accepts
38\% fewer tokens than full width on those 7 and 2.5\% fewer on the other 73.
Pooled $\AL$ is total generated tokens divided by total verification steps.
These 7 prompts contribute disproportionately to the aggregate because they
generate long responses and require more verification steps.
Table~\ref{tab:percattpot} gives each category's $\TPOT$ speedup from the same
runs. Half width is faster than full width in every category of every cell, by
1.03--1.11$\times$. Quarter width is too, by at least 1.02$\times$, except on
DFlash-Qwen, where it is at parity on multilingual (1.00$\times$) and
0.95$\times$ on RAG, whose 14.8\% acceptance cost outweighs its
1.11$\times$ faster iteration.

\paragraph{Released drafters.}
\label{app:stockarch}
Each full-width draft uses the architecture of the released drafter for its
cell (Table~\ref{tab:stock}), and the narrow drafts change only the residual
width and add the adapters. Our drafts differ from the released drafters in
training data and recipe and, for EAGLE-3, in the draft-vocabulary selection.
The releases document at most their training data, not their recipe or
budget, so we do not try to reproduce them. Every variant of a cell instead trains
on the same data and budget (Table~\ref{tab:setup}), and the releases show only
that our full-width drafts are competitive baselines.
The model card of \texttt{RedHatAI}'s Qwen3-8B drafter serves it with three
speculative tokens in its example command and reports acceptance for
$\blk{=}1$ through 7, so our $\blk{=}7$ lies within its evaluated range. We
report it because it is the only released EAGLE-3 drafter for both targets.
Every width of a cell uses the same $\blk$, so variants are compared at equal
draft length. For DFlash this is the released block size, and for EAGLE-3 it
is the seven steps our drafts are trained to unroll. The best $\blk$ may shift with
width, since a cheaper draft step makes a longer draft cheaper. For EAGLE-3
$\blk$ is a serving setting that can be swept without retraining, while
DFlash fixes its block size in training. In a separate run on EAGLE-3-Qwen at $B{=}1$ on
the batch-sweep prompts, we swept $\blk\in\{1,3,5,7\}$ at each width and scored
every setting on the same prompts. The best $\blk$ is 3 for full width and 5
for both narrow widths. This is a best case, not a setting a server can count
on. The best $\blk$ depends on acceptance, which varies by request and is not
known before decoding, so choosing it on the prompts it is scored on assumes
acceptance is known in advance. Under that assumption narrowing gains less,
1.03$\times$ at half width and 1.05$\times$ at quarter, against 1.08$\times$
and 1.11$\times$ at $\blk{=}7$ in the same run. We did not sweep $\blk$ on
GPT-OSS-20B or at larger batch.

\FloatBarrier
\section{Batch size and context length}
\label{app:scope}

\begin{widetable}
\centering
\footnotesize
\setlength{\tabcolsep}{4pt}
\caption{\textbf{Batch sweep detail\arxivgpu{ on B200}.} The first row of each cell is the full-width draft's
unprofiled $\Titer$. The next five rows are the Nsight stage split, in ms per iteration
except the draft share.
Verify does the same work in every variant, so only the full-width verify is shown. Draft
share is the full-width draft over draft plus verify. The last four rows are the narrow
drafts' unprofiled $\Titer$ speedup and $\AL$ ratio against full width, whose product is
the $\TPOT$ speedup of Table~\ref{tab:batch}. $\pm$ is the wider side of the
$\AL$ ratio's 95\% paired-bootstrap interval over the 512 prompts. $\Titer$
comes from one run per cell.}
\label{tab:batchstages}
\begin{tabular}{lllrrrrrr}
\toprule
Family & Target & Row & $B{=}1$ & $32$ & $64$ & $128$ & $256$ & $512$ \\
\midrule
\multirow{20}{*}{DFlash} & \multirow{10}{*}{Qwen3-8B} & full-width $\Titer$ (ms) & 3.50 & 7.42 & 12.30 & 22.33 & 42.58 & 83.69 \\
 &  & full-width draft & 0.895 & 1.831 & 3.386 & 6.516 & 12.683 & 26.556 \\
 &  & full-width verify & 2.629 & 5.070 & 8.360 & 15.297 & 29.310 & 57.915 \\
 &  & draft share & 25.4\% & 26.5\% & 28.8\% & 29.9\% & 30.2\% & 31.4\% \\
 &  & half draft & 0.643 & 1.320 & 2.213 & 4.113 & 8.414 & 17.905 \\
 &  & quarter draft & 0.525 & 1.051 & 1.731 & 3.024 & 6.114 & 13.357 \\
 &  & half $\Titer$ speedup & 1.08$\times$ & 1.12$\times$ & 1.14$\times$ & 1.15$\times$ & 1.13$\times$ & 1.12$\times$ \\
 &  & quarter $\Titer$ speedup & 1.14$\times$ & 1.19$\times$ & 1.17$\times$ & 1.23$\times$ & 1.20$\times$ & 1.21$\times$ \\
 &  & half $\AL$ ratio & 1.008{\tiny$\pm$.012} & 1.005{\tiny$\pm$.004} & 1.001{\tiny$\pm$.005} & 1.005{\tiny$\pm$.003} & 1.004{\tiny$\pm$.003} & 1.004{\tiny$\pm$.003} \\
 &  & quarter $\AL$ ratio & 0.987{\tiny$\pm$.008} & 0.987{\tiny$\pm$.004} & 0.986{\tiny$\pm$.004} & 0.987{\tiny$\pm$.004} & 0.985{\tiny$\pm$.005} & 0.985{\tiny$\pm$.005} \\
\cmidrule{2-9}
 & \multirow{10}{*}{GPT-OSS-20B} & full-width $\Titer$ (ms) & 3.29 & 6.29 & 8.67 & 13.75 & 24.54 & 45.23 \\
 &  & full-width draft & 0.941 & 1.471 & 2.248 & 4.088 & 7.682 & 15.771 \\
 &  & full-width verify & 2.126 & 4.401 & 5.573 & 8.483 & 16.103 & 28.446 \\
 &  & draft share & 30.7\% & 25.0\% & 28.7\% & 32.5\% & 32.3\% & 35.7\% \\
 &  & half draft & 0.753 & 1.202 & 1.796 & 3.273 & 5.668 & 12.305 \\
 &  & quarter draft & 0.628 & 1.033 & 1.571 & 2.653 & 4.986 & 10.909 \\
 &  & half $\Titer$ speedup & 1.07$\times$ & 1.07$\times$ & 1.07$\times$ & 1.11$\times$ & 1.09$\times$ & 1.10$\times$ \\
 &  & quarter $\Titer$ speedup & 1.11$\times$ & 1.08$\times$ & 1.10$\times$ & 1.14$\times$ & 1.13$\times$ & 1.13$\times$ \\
 &  & half $\AL$ ratio & 0.988{\tiny$\pm$.003} & 0.988{\tiny$\pm$.009} & 0.987{\tiny$\pm$.009} & 0.986{\tiny$\pm$.008} & 0.987{\tiny$\pm$.008} & 0.987{\tiny$\pm$.008} \\
 &  & quarter $\AL$ ratio & 0.965{\tiny$\pm$.002} & 0.963{\tiny$\pm$.009} & 0.960{\tiny$\pm$.008} & 0.960{\tiny$\pm$.008} & 0.960{\tiny$\pm$.008} & 0.960{\tiny$\pm$.007} \\
\midrule
\multirow{20}{*}{EAGLE-3} & \multirow{10}{*}{Qwen3-8B} & full-width $\Titer$ (ms) & 4.13 & 5.91 & 8.14 & 12.81 & 22.88 & 43.21 \\
 &  & full-width draft & 1.432 & 1.764 & 2.214 & 3.136 & 5.351 & 10.321 \\
 &  & full-width verify & 2.756 & 4.043 & 5.767 & 9.308 & 17.111 & 32.013 \\
 &  & draft share & 34.2\% & 30.4\% & 27.7\% & 25.2\% & 23.8\% & 24.4\% \\
 &  & half draft & 1.021 & 1.345 & 1.733 & 2.491 & 4.308 & 8.374 \\
 &  & quarter draft & 0.848 & 1.134 & 1.492 & 2.152 & 3.806 & 7.834 \\
 &  & half $\Titer$ speedup & 1.11$\times$ & 1.08$\times$ & 1.07$\times$ & 1.06$\times$ & 1.06$\times$ & 1.05$\times$ \\
 &  & quarter $\Titer$ speedup & 1.18$\times$ & 1.12$\times$ & 1.09$\times$ & 1.08$\times$ & 1.10$\times$ & 1.08$\times$ \\
 &  & half $\AL$ ratio & 0.971{\tiny$\pm$.005} & 0.969{\tiny$\pm$.004} & 0.968{\tiny$\pm$.005} & 0.969{\tiny$\pm$.004} & 0.970{\tiny$\pm$.004} & 0.970{\tiny$\pm$.004} \\
 &  & quarter $\AL$ ratio & 0.955{\tiny$\pm$.005} & 0.956{\tiny$\pm$.004} & 0.956{\tiny$\pm$.004} & 0.955{\tiny$\pm$.004} & 0.956{\tiny$\pm$.004} & 0.956{\tiny$\pm$.004} \\
\cmidrule{2-9}
 & \multirow{10}{*}{GPT-OSS-20B} & full-width $\Titer$ (ms) & 3.62 & 6.79 & 8.61 & 12.83 & 22.18 & 39.51 \\
 &  & full-width draft & 1.136 & 1.500 & 1.879 & 2.714 & 4.606 & 9.254 \\
 &  & full-width verify & 2.501 & 4.544 & 5.823 & 8.601 & 15.787 & 28.426 \\
 &  & draft share & 31.2\% & 24.8\% & 24.4\% & 24.0\% & 22.6\% & 24.6\% \\
 &  & half draft & 0.869 & 1.211 & 1.540 & 2.265 & 4.002 & 8.025 \\
 &  & quarter draft & 0.769 & 1.058 & 1.393 & 2.086 & 3.682 & 7.319 \\
 &  & half $\Titer$ speedup & 1.09$\times$ & 1.04$\times$ & 1.02$\times$ & 1.07$\times$ & 1.05$\times$ & 1.04$\times$ \\
 &  & quarter $\Titer$ speedup & 1.14$\times$ & 1.07$\times$ & 1.05$\times$ & 1.07$\times$ & 1.06$\times$ & 1.08$\times$ \\
 &  & half $\AL$ ratio & 0.980{\tiny$\pm$.003} & 0.982{\tiny$\pm$.008} & 0.984{\tiny$\pm$.009} & 0.979{\tiny$\pm$.007} & 0.976{\tiny$\pm$.007} & 0.976{\tiny$\pm$.007} \\
 &  & quarter $\AL$ ratio & 0.970{\tiny$\pm$.003} & 0.968{\tiny$\pm$.008} & 0.968{\tiny$\pm$.008} & 0.967{\tiny$\pm$.007} & 0.967{\tiny$\pm$.007} & 0.967{\tiny$\pm$.007} \\
\bottomrule
\end{tabular}
\end{widetable}

\begin{widetable}
\centering
\footnotesize
\setlength{\tabcolsep}{4pt}
\caption{\textbf{Context-length sweep at batch 1\arxivgpu{ on B200}.} Unprofiled $\Titer$ in ms, with
the $\Titer$ speedup over full width at the same context in brackets. The last
two columns give the change from 8k to 32k in full width's $\Titer$ and in
each narrow draft's speedup.}
\label{tab:ctx}
\begin{tabular}{lllrrrrr}
\toprule
 & & & & & & \multicolumn{2}{c}{8k $\to$ 32k} \\
\cmidrule(lr){7-8}
Family & Target & Draft & 8k & 16k & 32k & full $\Titer$ & speedup \\
\midrule
\multirow{6}{*}{DFlash}
 & \multirow{3}{*}{Qwen3-8B}
   & full   & 3.770 & 3.902 & 4.098 & $+$8.7\% & --- \\
 & & half    & 3.482 (1.08$\times$) & 3.620 (1.08$\times$) & 3.806 (1.08$\times$) & --- & $-$0.6\% \\
 & & quarter & 3.340 (1.13$\times$) & 3.476 (1.12$\times$) & 3.675 (1.12$\times$) & --- & $-$1.2\% \\
\cmidrule{2-8}
 & \multirow{3}{*}{GPT-OSS-20B}
   & full   & 3.410 & 3.491 & 3.547 & $+$4.0\% & --- \\
 & & half    & 3.206 (1.06$\times$) & 3.303 (1.06$\times$) & 3.364 (1.05$\times$) & --- & $-$0.9\% \\
 & & quarter & 3.102 (1.10$\times$) & 3.204 (1.09$\times$) & 3.275 (1.08$\times$) & --- & $-$1.5\% \\
\midrule
\multirow{6}{*}{EAGLE-3}
 & \multirow{3}{*}{Qwen3-8B}
   & full   & 4.364 & 4.502 & 4.716 & $+$8.1\% & --- \\
 & & half    & 3.944 (1.11$\times$) & 4.071 (1.11$\times$) & 4.282 (1.10$\times$) & --- & $-$0.5\% \\
 & & quarter & 3.817 (1.14$\times$) & 3.948 (1.14$\times$) & 4.177 (1.13$\times$) & --- & $-$1.2\% \\
\cmidrule{2-8}
 & \multirow{3}{*}{GPT-OSS-20B}
   & full   & 3.742 & 3.780 & 3.844 & $+$2.7\% & --- \\
 & & half    & 3.459 (1.08$\times$) & 3.492 (1.08$\times$) & 3.536 (1.09$\times$) & --- & $+$0.5\% \\
 & & quarter & 3.299 (1.13$\times$) & 3.366 (1.12$\times$) & 3.443 (1.12$\times$) & --- & $-$1.6\% \\
\bottomrule
\end{tabular}
\end{widetable}

\paragraph{Batch sweep.} Every batch size serves the same 512 single-turn
SPEED-Bench \texttt{throughput\_1k} prompts. They are the first 256
non-placeholder prompts by question ID in each of the high-entropy split (book
continuation and WritingBench) and the low-entropy split (repository code
completion). After the chat
template they are 928--2{,}517 tokens long, mean 1{,}039, in the Qwen3-8B
tokenizer, and 1{,}067--2{,}597 tokens, mean 1{,}072, in GPT-OSS-20B's, where
510 of the 512 render to exactly 1{,}067. Each $B$ serves them as $512/B$
sequential chunks of $B$ distinct prompts, at temperature 0 with up to
1{,}024 new tokens, stopping at end of sequence, with a 4{,}096-token context
and radix cache off. Every cell reaches its full $B$, and no request is
preempted.

$\AL$ is total completion tokens over total verify steps, each counted per
request, across all 512 prompts. $\Titer$ is the mean step time over the decode steps of each chunk
at which all $B$ requests are running, dropping the first two and last two of
each chunk and any step longer than twice the cell's median. Within a cell a
step's GPU work cannot double, so such steps are host or node stalls. They are
at most 0.9\% of a cell's steps and occur almost only at $B{=}1$. We report
$\TPOT$ as this full-batch $\Titer$ over the whole-run $\AL$ of the same
run, which approximates a server kept full at $B$ and differs from the per-turn
estimator of Table~\ref{tab:headline}. We use whole-run $\AL$ because at large
$B$ the full-batch steps cover only the start of each response, whose
acceptance is not representative of whole responses, 13.0\% higher at
$B{=}512$ on DFlash-Qwen.
This assumes each variant's whole-run $\AL$ also holds
during the full-batch steps. The early-response bias is shared across widths,
so taking tokens and time from the same full-batch steps instead changes no
$\TPOT$ speedup by more than 0.017$\times$ (at most from 1.091$\times$ to
1.074$\times$), and every speedup stays above 1.00$\times$. Keeping the
steps longer than twice the median changes none by more than 0.025$\times$ (at
most from 1.086$\times$ to 1.061$\times$, at $B{=}1$) and leaves the smallest
unchanged. Greedy
output depends on $B$ through kernel numerics, so we compare variants only at the
same $B$, each $\AL$ ratio with a 95\% paired-bootstrap interval over prompts
(Table~\ref{tab:batchstages}).

Stage times come from one Nsight capture per variant and $B$, ten decode steps
after five full-batch warmup steps, all with $B$ requests running. As at batch 1
(App.~\ref{app:profiling}), the profiler leaves out time a step spends off the
GPU in three cells, so their stage sums can fall up to 13\% below the
unprofiled $\Titer$. Nsight therefore supplies
proportions only, and every speedup comes from the unprofiled runs.

\paragraph{Context sweep.} Batch 1, six SPEED-Bench \texttt{throughput\_8k},
\texttt{16k} and \texttt{32k} prompts per context, run back to back in one job
per variant with a 34{,}816-token window, \texttt{max\_new\_tokens} 2{,}048 and no
profiling. The per-prompt standard error of the speedup is 0.12--0.2\% on
DFlash-Qwen and at most 0.85\% on DFlash-GPT-OSS. On EAGLE-3-GPT-OSS quarter
width's speedup falls 1.6\% from 8k to 32k, about four standard errors, while
half width's is flat.

\FloatBarrier
\section{Per-operation attribution in full}
\label{app:l3full}

Table~\ref{tab:l3full} gives the time of every draft operation.

\sharedtablehere{\begin{widetable}
\centering
\scriptsize
\setlength{\tabcolsep}{3pt}
\caption{\textbf{Every draft operation, all four cells\arxivgpu{, on B200}}, $\mu$s per iteration and
share of the variant's draft-body saving. Totals are draft-body kernel time,
without inter-kernel gaps, the DFlash context pre-pass (Table~\ref{tab:l3}, outside) or the EAGLE-3 carry-in
GEMM. GEMMs are identified by sequential position (App.~\ref{app:profiling}).
Negative entries are costs of narrowing. The DFlash input adapter, the only
operation narrowing adds, takes 4.2--7.2\,$\mu$s per iteration and is counted in \emph{other}.}
\label{tab:l3full}
\begin{tabular}{lrrrrr@{\hskip 1.2em}rrrrr}
\toprule
& \multicolumn{5}{c}{\textbf{Qwen3-8B}} & \multicolumn{5}{c}{\textbf{GPT-OSS-20B}} \\
\cmidrule(lr){2-6}\cmidrule(lr){7-11}
Operation & full & half & qtr & half save & qtr save & full & half & qtr & half save & qtr save \\
\midrule
\multicolumn{11}{l}{\emph{DFlash} ($\mu$s per iteration)}\\
\textbf{folded LM head} & \textbf{206.2} & 99.4 & 53.3 & \textbf{42.4\%} & \textbf{43.6\%} & \textbf{166.5} & 86.9 & 46.1 & \textbf{37.6\%} & \textbf{40.6\%} \\
FC1 (\texttt{gate\_up})  & 180.7 & 95.3 & 56.0 & 33.9\% & 35.5\% & 134.1 & 76.9 & 50.0 & 27.0\% & 28.3\% \\
FC2 (\texttt{down})      &  90.7 & 53.2 & 47.2 & 14.9\% & 12.4\% &  77.5 & 54.9 & 51.7 & 10.7\% &  8.7\% \\
QKV projection           &  52.3 & 34.2 & 26.1 &  7.2\% &  7.5\% &  60.5 & 41.7 & 32.2 &  8.9\% &  9.5\% \\
O projection             &  44.0 & 35.9 & 27.3 &  3.2\% &  4.8\% &  53.2 & 44.2 & 42.6 &  4.2\% &  3.6\% \\
\emph{attention}         &  55.2 & 54.5 & 55.3 &  0.3\% & $-0.0$\% &  84.2 & 80.7 & 81.1 & 1.7\% & 1.1\% \\
\emph{other}             & 144.2 & 149.0 & 157.4 & $-1.9\%$ & $-3.8\%$ & 241.1 & 220.0 & 216.7 & 10.0\% & 8.2\% \\
\textbf{body total}     & \textbf{773.2} & \textbf{521.5} & \textbf{422.6} & \textbf{251.7} & \textbf{350.6} & \textbf{817.1} & \textbf{605.3} & \textbf{520.5} & \textbf{211.7} & \textbf{296.6} \\
\midrule
\multicolumn{11}{l}{\emph{EAGLE-3} ($\mu$s per iteration, seven speculative steps)}\\
\textbf{LM head}         & \textbf{304.9} & 174.2 & 96.4 & \textbf{29.6\%} & \textbf{34.5\%} & \textbf{403.2} & 226.0 & 130.7 & \textbf{80.6\%} & \textbf{80.6\%} \\
FC1 (\texttt{gate\_up})  & 239.4 & 131.0 & 78.7 & 24.5\% & 26.6\% &  58.0 & 39.0 & 29.6 &  8.6\% &  8.4\% \\
QKV projection           & 131.0 &  73.6 & 48.5 & 13.0\% & 13.7\% &  85.1 & 53.5 & 37.5 & 14.4\% & 14.1\% \\
FC2 (\texttt{down})      & 127.5 &  76.1 & 70.9 & 11.6\% &  9.4\% &  42.6 & 46.1 & 32.3 & $-1.6\%$ & 3.0\% \\
O projection             &  62.1 &  48.1 & 37.7 &  3.2\% &  4.0\% &  53.5 & 45.5 & 37.2 &  3.6\% &  4.8\% \\
norms                    & \multicolumn{5}{c}{\emph{(inside \texttt{other})}} &  86.4 & 75.4 & 60.4 & 5.0\% & 7.7\% \\
\emph{attention}         &  83.5 &  82.7 & 82.5 &  0.2\% &  0.2\% &  76.4 & 79.8 & 79.6 & $-1.5\%$ & $-0.9\%$ \\
\emph{other}             & 472.0 & 392.7 & 402.1 & 17.9\% & 11.6\% & 318.6 & 338.7 & 378.4 & $-9.1\%$ & $-17.7\%$ \\
\textbf{body total}     & \textbf{1420.4} & \textbf{978.4} & \textbf{816.8} & \textbf{442.0} & \textbf{603.6} & \textbf{1123.8} & \textbf{904.0} & \textbf{785.7} & \textbf{219.8} & \textbf{338.1} \\
\bottomrule
\end{tabular}
\end{widetable}
}

\section{Profiling method}
\label{app:profiling}

\paragraph{Timing at batch 1.} The $\Titer$ and $\TPOT$ of
Table~\ref{tab:headline} use every decode step of turns with at least 8 steps
except the first, whose log line includes prefill time. $\Titer$ is their mean
time and $\TPOT$ their total time over the tokens they accept, so $\TPOT$
equals $\Titer$ over those steps' mean acceptance, which differs from $\AL$ by
at most 0.4\%.

Every profile uses one summarization prompt, 866 tokens in the Qwen3-8B
tokenizer, with a 4{,}096-token context and 128 new tokens, and runs in production mode, with CUDA graphs and overlap scheduling
on, and discards graph warmup. Profiling changes absolute times by a few
percent, but width ratios agree with an unprofiled run of the same prompt to
0.2--0.9\% in every cell, so we read profiled stage times as proportions. Per-operation NVTX ranges do not survive
CUDA graphs, so we identify each draft GEMM by its position in the kernel
stream rather than by name.

\FloatBarrier
\section{DGX Spark (GB10) setup and results}
\label{app:gb10}

\sharedtablehere{\begin{widetable}
\centering
\footnotesize
\setlength{\tabcolsep}{3pt}
\caption{\textbf{Serving on B200 and GB10.} Every setting matches B200 unless
GB10 cannot run it. Within a GPU, all three widths of a cell share one
configuration.}
\label{tab:gb10setup}
\begin{tabular}{lll}
\toprule
Setting & B200 & GB10 \\
\midrule
GPU & SM100, HBM3e, 8\,TB/s & SM121, unified LPDDR5x, 273\,GB/s \\
Target attention, Qwen3-8B & TRT-LLM MHA & \textbf{FlashInfer} \\
Target attention, GPT-OSS-20B & TRT-LLM MHA & \textbf{FA4} \\
KV page size (follows the target attention backend) & 64 & \textbf{1} \\
Static memory fraction & 0.75--0.9 & \textbf{0.5} \\
\midrule
Draft attention & FA4 & FA4 \\
MoE runner, GPT-OSS-20B & FlashInfer MXFP4 & FlashInfer MXFP4 \\
KV cache & target FP8 on Qwen3-8B, else BF16 & same \\
CUDA graphs, overlap, radix cache & on, on, off & same \\
Speculative settings, drafts, targets & Table~\ref{tab:setup} & same \\
SGLang & 0.5.18 nightly & same \\
\bottomrule
\end{tabular}
\end{widetable}
}
\sharedtablehere{\begin{widetable}
\centering
\footnotesize
\setlength{\tabcolsep}{4pt}
\caption{\textbf{Main result on DGX Spark (GB10) at temperature 0.} The same
drafts and 880-prompt contract as Table~\ref{tab:headline}, batch 1, one GB10.
$\TPOT$ is $\Titer$ over pooled $\AL$. $\Delta\AL$ is relative to the
full-width draft. Bold marks each cell's best $\AL$ and $\TPOT$
speedup. Temperature 1 is in Table~\ref{tab:gb10t1}.}
\label{tab:gb10headline}
\begin{tabular}{llrrrrrr}
\toprule
Target & Draft & $\AL$ & $\Delta\AL$ & $\Titer$ (ms) & $\TPOT$ (ms) & $\Titer$ speedup & \textbf{\boldmath$\TPOT$ speedup} \\
\midrule
\multicolumn{8}{l}{\textbf{DFlash}}\\
\midrule
\multirow{3}{*}{Qwen3-8B}
 & full 4{,}096    & \textbf{4.156} & ---       & 41.810 & 10.060 & 1.00$\times$ & 1.00$\times$ \\
 & half 2{,}048    & 4.149 & $-0.17\%$ & 34.718 & 8.368 & 1.20$\times$ & 1.20$\times$ \\
 & quarter 1{,}024 & 3.899 & $-6.17\%$ & 31.025 & 7.956 & 1.35$\times$ & \textbf{1.26$\times$} \\
\midrule
\multirow{3}{*}{GPT-OSS-20B}
 & full 2{,}880    & \textbf{4.395} & ---       & 46.232 & 10.520 & 1.00$\times$ & 1.00$\times$ \\
 & half 1{,}440    & 4.306 & $-2.02\%$ & 40.142 & 9.323 & 1.15$\times$ & 1.13$\times$ \\
 & quarter 720     & 4.224 & $-3.89\%$ & 37.143 & 8.794 & 1.24$\times$ & \textbf{1.20$\times$} \\
\midrule
\multicolumn{8}{l}{\textbf{EAGLE-3}}\\
\midrule
\multirow{3}{*}{Qwen3-8B}
 & full 4{,}096    & \textbf{3.487} & ---       & 50.923 & 14.604 & 1.00$\times$ & 1.00$\times$ \\
 & half 2{,}048    & 3.381 & $-3.02\%$ & 40.761 & 12.054 & 1.25$\times$ & 1.21$\times$ \\
 & quarter 1{,}024 & 3.389 & $-2.81\%$ & 34.440 & 10.162 & 1.48$\times$ & \textbf{1.44$\times$} \\
\midrule
\multirow{3}{*}{GPT-OSS-20B}
 & full 2{,}880    & 3.733 & ---       & 52.660 & 14.108 & 1.00$\times$ & 1.00$\times$ \\
 & half 1{,}440    & \textbf{3.745} & $+0.33\%$ & 42.894 & 11.453 & 1.23$\times$ & 1.23$\times$ \\
 & quarter 720     & 3.634 & $-2.65\%$ & 38.168 & 10.504 & 1.38$\times$ & \textbf{1.34$\times$} \\
\bottomrule
\end{tabular}
\end{widetable}
}
\sharedtablehere{\begin{widetable}
\centering
\footnotesize
\setlength{\tabcolsep}{4pt}
\caption{\textbf{GB10 at temperature 1.} As Table~\ref{tab:gb10headline},
at temperature 1 with seed 7.}
\label{tab:gb10t1}
\begin{tabular}{llrrrrrr}
\toprule
Target & Draft & $\AL$ & $\Delta\AL$ & $\Titer$ (ms) & $\TPOT$ (ms) & $\Titer$ speedup & \textbf{\boldmath$\TPOT$ speedup} \\
\midrule
\multicolumn{8}{l}{\textbf{DFlash}}\\
\midrule
\multirow{3}{*}{Qwen3-8B}
 & full 4{,}096    & \textbf{2.848} & ---       & 42.769 & 15.016 & 1.00$\times$ & 1.00$\times$ \\
 & half 2{,}048    & 2.828 & $-0.70\%$ & 35.587 & 12.582 & 1.20$\times$ & 1.19$\times$ \\
 & quarter 1{,}024 & 2.738 & $-3.86\%$ & 32.004 & 11.688 & 1.34$\times$ & \textbf{1.28$\times$} \\
\midrule
\multirow{3}{*}{GPT-OSS-20B}
 & full 2{,}880    & \textbf{2.275} & ---       & 45.786 & 20.128 & 1.00$\times$ & 1.00$\times$ \\
 & half 1{,}440    & 2.249 & $-1.16\%$ & 39.828 & 17.713 & 1.15$\times$ & 1.14$\times$ \\
 & quarter 720     & 2.226 & $-2.14\%$ & 36.719 & 16.494 & 1.25$\times$ & \textbf{1.22$\times$} \\
\midrule
\multicolumn{8}{l}{\textbf{EAGLE-3}}\\
\midrule
\multirow{3}{*}{Qwen3-8B}
 & full 4{,}096    & \textbf{2.637} & ---       & 50.703 & 19.228 & 1.00$\times$ & 1.00$\times$ \\
 & half 2{,}048    & 2.542 & $-3.62\%$ & 40.348 & 15.876 & 1.26$\times$ & 1.21$\times$ \\
 & quarter 1{,}024 & 2.530 & $-4.04\%$ & 33.787 & 13.353 & 1.50$\times$ & \textbf{1.44$\times$} \\
\midrule
\multirow{3}{*}{GPT-OSS-20B}
 & full 2{,}880    & 2.076 & ---       & 52.685 & 25.377 & 1.00$\times$ & 1.00$\times$ \\
 & half 1{,}440    & \textbf{2.093} & $+0.80\%$ & 42.865 & 20.483 & 1.23$\times$ & 1.24$\times$ \\
 & quarter 720     & 2.046 & $-1.46\%$ & 38.031 & 18.590 & 1.39$\times$ & \textbf{1.37$\times$} \\
\bottomrule
\end{tabular}
\end{widetable}
}

\paragraph{Setup.} GB10 runs the same drafts, targets, SGLang build and
evaluation as B200, and keeps every B200 serving setting it can
(Table~\ref{tab:gb10setup}). Two things differ. The B200 target attention
kernel, TRT-LLM MHA, does not run on GB10, so the Qwen3-8B target uses
FlashInfer and GPT-OSS-20B uses FA4, the one option SGLang supports for its
attention sinks, and the KV page size follows each backend's default. GB10's
128\,GB of memory is also shared with the operating system, so SGLang gets half
of it. At that budget the Qwen3-8B KV pool holds 542K tokens with DFlash and
678K with EAGLE-3, which seats only 305 and 385 of the 512 requests at
$B{=}512$, so its sweeps stop at $B{=}256$. GPT-OSS-20B, with a 128-token
window in half its layers, seats all 512.

\paragraph{Measurement.} All three widths of a cell run back to back on one
node, and repeated full-width runs on different nodes agree within 0.7\% on
$\Titer$. $\TPOT$ on the panel is mean $\Titer$ over pooled $\AL$.

\paragraph{Results.} At temperature 0 halving gives 1.13--1.23$\times$ end
to end and quartering 1.20--1.44$\times$ (Table~\ref{tab:gb10headline}), and
at temperature 1 1.14--1.24$\times$ and 1.22--1.44$\times$
(Table~\ref{tab:gb10t1}). Quarter width is the fastest variant in every cell
at both temperatures. Full-width $\AL$ is within 3.3\% of B200's in every
cell, and narrowing costs at most 3.0\% acceptance at half width and
2.7--6.2\% at quarter, within 1.8 points of B200. On $\Titer$ alone half width
is 1.15--1.25$\times$ faster and quarter width 1.24--1.48$\times$, against
1.07--1.12$\times$ and 1.11--1.16$\times$ on B200.

\sharedtablehere{\begin{widetable}
\centering
\footnotesize
\setlength{\tabcolsep}{4pt}
\caption{\textbf{GB10 stage split.} Per-iteration stage times on GB10 at batch 1 on the
profiling prompt of Table~\ref{tab:l1}, with $\tdraft$ measured as there.
$\tverify$ is the unprofiled $\Titer$ minus $\tdraft$, so the last column is the
unprofiled $\Titer$ (App.~\ref{app:gb10}).}
\label{tab:gb10l1}
\begin{tabular}{llrrrr}
\toprule
Family / target & Draft & $\tdraft$ (ms) & vs full & $\tverify$ (ms) & sum, vs full \\
\midrule
\multirow{3}{*}{DFlash, Qwen3-8B}
 & full 4{,}096 & 15.38 & --- & 25.42 & 40.80 \\
 & half 2{,}048 & 8.10 & \textbf{\textminus47.3\%} & 25.32 ($-0.40\%$) & 33.42 (1.22$\times$) \\
 & quarter 1{,}024 & 4.30 & \textbf{\textminus72.0\%} & 25.42 ($-0.02\%$) & 29.72 (1.37$\times$) \\
\midrule
\multirow{3}{*}{DFlash, GPT-OSS-20B}
 & full 2{,}880 & 12.80 & --- & 31.45 & 44.25 \\
 & half 1{,}440 & 6.67 & \textbf{\textminus47.9\%} & 31.72 ($+0.87\%$) & 38.39 (1.15$\times$) \\
 & quarter 720 & 3.72 & \textbf{\textminus70.9\%} & 32.01 ($+1.78\%$) & 35.73 (1.24$\times$) \\
\midrule
\multirow{3}{*}{EAGLE-3, Qwen3-8B}
 & full 4{,}096 & 24.34 & --- & 29.21 & 53.55 \\
 & half 2{,}048 & 13.68 & \textbf{\textminus43.8\%} & 28.95 ($-0.90\%$) & 42.63 (1.26$\times$) \\
 & quarter 1{,}024 & 6.87 & \textbf{\textminus71.8\%} & 29.61 ($+1.38\%$) & 36.49 (1.47$\times$) \\
\midrule
\multirow{3}{*}{EAGLE-3, GPT-OSS-20B}
 & full 2{,}880 & 19.52 & --- & 32.20 & 51.72 \\
 & half 1{,}440 & 10.09 & \textbf{\textminus48.3\%} & 31.95 ($-0.77\%$) & 42.04 (1.23$\times$) \\
 & quarter 720 & 5.04 & \textbf{\textminus74.2\%} & 32.11 ($-0.26\%$) & 37.15 (1.39$\times$) \\
\bottomrule
\end{tabular}
\end{widetable}
}
\sharedtablehere{\begin{widetable}
\centering
\footnotesize
\setlength{\tabcolsep}{3pt}
\caption{\textbf{Where the draft-side saving comes from on GB10.} ms per
iteration, ratio to full width in brackets, with columns as in
Table~\ref{tab:l3}. GEMMs are assigned to operations by their position in the
kernel stream. \emph{Outside} holds the DFlash context pre-pass and input
adapter and the EAGLE-3 carry-in GEMM, plus inter-kernel gaps on DFlash-Qwen, so
each row sums to $\tdraft$ (Table~\ref{tab:gb10l1}). Unlike Table~\ref{tab:l3}, the DFlash
input adapter is in \emph{outside}, because the position-based assignment does not
separate its GEMM from the context projections. Attention, inside
\emph{rest of body}, stays at 0.08--0.19\,ms.}
\label{tab:gb10l2}
\begin{tabular}{llrrrrrr}
\toprule
Family / target & Draft & attn.\ proj. & MLP & head & rest of body & outside & $\tdraft$ \\
\midrule
\multirow{3}{*}{DFlash, Qwen3-8B}
 & full 4{,}096 & 1.92 & 6.70 & 5.39 & 0.34 & 1.04 & 15.38 \\
 & half & 0.96 (2.00$\times$) & 3.39 (1.97$\times$) & 2.68 (2.01$\times$) & 0.31 (1.12$\times$) & 0.76 & 8.10 (1.90$\times$) \\
 & quarter & 0.52 (3.69$\times$) & 1.71 (3.91$\times$) & 1.36 (3.95$\times$) & 0.28 (1.20$\times$) & 0.43 & 4.30 (3.57$\times$) \\
\midrule
\multirow{3}{*}{DFlash, GPT-OSS-20B}
 & full 2{,}880 & 1.93 & 4.73 & 5.02 & 0.43 & 0.69 & 12.80 \\
 & half & 1.02 (1.90$\times$) & 2.36 (2.00$\times$) & 2.53 (1.99$\times$) & 0.36 (1.20$\times$) & 0.41 & 6.67 (1.92$\times$) \\
 & quarter & 0.52 (3.70$\times$) & 1.26 (3.76$\times$) & 1.38 (3.63$\times$) & 0.35 (1.23$\times$) & 0.21 & 3.72 (3.44$\times$) \\
\midrule
\multirow{3}{*}{EAGLE-3, Qwen3-8B}
 & full 4{,}096 & 4.43 & 9.02 & 9.99 & 0.41 & 0.48 & 24.34 \\
 & half & 2.70 (1.64$\times$) & 5.41 (1.67$\times$) & 4.92 (2.03$\times$) & 0.39 (1.05$\times$) & 0.27 & 13.68 (1.78$\times$) \\
 & quarter & 1.34 (3.32$\times$) & 2.66 (3.39$\times$) & 2.39 (4.18$\times$) & 0.36 (1.14$\times$) & 0.13 & 6.87 (3.54$\times$) \\
\midrule
\multirow{3}{*}{EAGLE-3, GPT-OSS-20B}
 & full 2{,}880 & 2.75 & 1.96 & 14.16 & 0.39 & 0.25 & 19.52 \\
 & half & 1.69 (1.63$\times$) & 1.00 (1.96$\times$) & 6.89 (2.06$\times$) & 0.37 (1.06$\times$) & 0.14 & 10.09 (1.93$\times$) \\
 & quarter & 0.90 (3.06$\times$) & 0.46 (4.22$\times$) & 3.12 (4.53$\times$) & 0.48 (0.81$\times$) & 0.08 & 5.04 (3.87$\times$) \\
\bottomrule
\end{tabular}
\end{widetable}
}
\sharedtablehere{\begin{widetable}
\centering
\footnotesize
\setlength{\tabcolsep}{3pt}
\caption{\textbf{Draft GEMM time and bandwidth on B200 and GB10.} DFlash-Qwen,
batch 1. The vocabulary projection is one kernel per iteration, and each other
projection is one kernel in each of the five draft layers. Weight MB is one
kernel's BF16 weights at full width, and halves with each halving of width.
$\mu$s is one kernel's measured time. \% is weight bytes over time as a
share of peak memory bandwidth (8\,TB/s on B200, 273\,GB/s on GB10), which no
kernel reaches in practice, so the trend with width matters more than the
absolute level. Activations are small at batch 1, so weight bytes are close to
all the bytes a kernel moves. \emph{All five}
pools the 21 kernels.}
\label{tab:opbw}
\begin{tabular}{lrrrrrrrrrrrrrr}
\toprule
 & & & \multicolumn{6}{c}{B200} & \multicolumn{6}{c}{GB10} \\
\cmidrule(lr){4-9}\cmidrule(lr){10-15}
 & & weight & \multicolumn{3}{c}{$\mu$s per kernel} & \multicolumn{3}{c}{\% of peak} & \multicolumn{3}{c}{$\mu$s per kernel} & \multicolumn{3}{c}{\% of peak} \\
\cmidrule(lr){4-6}\cmidrule(lr){7-9}\cmidrule(lr){10-12}\cmidrule(lr){13-15}
Operation & kernels & MB & full & half & qtr & full & half & qtr & full & half & qtr & full & half & qtr \\
\midrule
Vocabulary projection & 1 & 1{,}245 & 206.2 & 99.4 & 53.3 & 75 & 78 & 73 & 5{,}388 & 2{,}679 & 1{,}363 & 85 & 85 & 84 \\
FC1 (gate/up)         & 5 & 201     & 36.1 & 19.1 & 11.2 & 70 & 66 & 56 & 882 & 445 & 226 & 84 & 83 & 82 \\
FC2 (down)            & 5 & 101     & 18.1 & 10.6 & 9.4  & 69 & 59 & 33 & 457 & 234 & 116 & 81 & 79 & 79 \\
QKV projection        & 5 & 50      & 10.5 & 6.8  & 5.2  & 60 & 46 & 30 & 228 & 114 & 58  & 81 & 81 & 80 \\
O projection          & 5 & 34      & 8.8  & 7.2  & 5.5  & 48 & 29 & 19 & 155 & 78  & 46  & 79 & 79 & 67 \\
\midrule
All five              & 21 & ---    & ---  & ---  & ---  & 69 & 62 & 47 & --- & --- & --- & 83 & 83 & 81 \\
\bottomrule
\end{tabular}
\end{widetable}
}

\paragraph{Profiles.} Stage times come from one Nsight capture per variant on
the same prompt and settings as the B200 profiles (App.~\ref{app:profiling}),
with $\tdraft$ measured as on B200 in each cell and GEMMs assigned to operations
by their position in the kernel stream. The GB10 entries of all three tables
average the same nine decode windows per variant. $\tverify$ is taken from the unprofiled $\Titer$
because in DFlash-GPT-OSS the profiled verify window shifts by up to 6\% across
widths. It moves by at most 1.8\% with width in every cell.
Table~\ref{tab:gb10l1} gives the stage split, Table~\ref{tab:gb10l2} the
draft time by operation and Table~\ref{tab:opbw} the time and fraction
of peak bandwidth of each DFlash-Qwen draft GEMM kernel on both GPUs.

\sharedtablehere{\begin{widetable}
\centering
\footnotesize
\setlength{\tabcolsep}{4pt}
\caption{\textbf{End-to-end speedup over full width across batch size on GB10.}
$\TPOT$ speedup on the 512 SPEED-Bench throughput prompts of
Table~\ref{tab:batch}, under the same estimator. The first row of each cell is
the full-width draft's unprofiled $\Titer$. Draft share is full-width draft time over draft plus
verify, from a separate Nsight capture of about ten full-batch steps on the first $B$ of
the same prompts, measured as in Table~\ref{tab:batch}. There is no capture at $B{=}512$. Qwen3-8B does not fit at $B{=}512$ in the
memory GB10 leaves for SGLang (App.~\ref{app:gb10}).}
\label{tab:gb10batch}
\begin{tabular}{lllrrrrrr}
\toprule
Family & Target & Row & $B{=}1$ & $32$ & $64$ & $128$ & $256$ & $512$ \\
\midrule
\multirow{8}{*}{DFlash} & \multirow{4}{*}{Qwen3-8B} & full-width $\Titer$ (ms) & 45.4 & 110.3 & 187.0 & 383.3 & 804.4 & --- \\
 &  & half & 1.18$\times$ & 1.13$\times$ & 1.12$\times$ & 1.14$\times$ & 1.12$\times$ & --- \\
 &  & quarter & 1.29$\times$ & 1.17$\times$ & 1.16$\times$ & 1.19$\times$ & 1.16$\times$ & --- \\
 &  & draft share & 36.9\% & 30.9\% & 27.6\% & 28.1\% & 26.1\% & --- \\
 & \multirow{4}{*}{GPT-OSS-20B} & full-width $\Titer$ (ms) & 45.7 & 96.5 & 126.8 & 202.0 & 382.1 & 709.9 \\
 &  & half & 1.13$\times$ & 1.06$\times$ & 1.06$\times$ & 1.06$\times$ & 1.09$\times$ & 1.09$\times$ \\
 &  & quarter & 1.20$\times$ & 1.11$\times$ & 1.07$\times$ & 1.08$\times$ & 1.11$\times$ & 1.11$\times$ \\
 &  & draft share & 28.6\% & 21.7\% & 22.4\% & 25.0\% & 26.6\% & --- \\
\midrule
\multirow{8}{*}{EAGLE-3} & \multirow{4}{*}{Qwen3-8B} & full-width $\Titer$ (ms) & 53.6 & 86.9 & 132.5 & 228.4 & 439.2 & --- \\
 &  & half & 1.20$\times$ & 1.12$\times$ & 1.09$\times$ & 1.07$\times$ & 1.04$\times$ & --- \\
 &  & quarter & 1.38$\times$ & 1.21$\times$ & 1.14$\times$ & 1.10$\times$ & 1.05$\times$ & --- \\
 &  & draft share & 48.9\% & 34.2\% & 28.5\% & 25.8\% & 20.9\% & --- \\
 & \multirow{4}{*}{GPT-OSS-20B} & full-width $\Titer$ (ms) & 52.4 & 98.4 & 125.7 & 188.4 & 338.2 & 627.1 \\
 &  & half & 1.20$\times$ & 1.07$\times$ & 1.05$\times$ & 1.02$\times$ & 1.01$\times$ & 1.02$\times$ \\
 &  & quarter & 1.31$\times$ & 1.12$\times$ & 1.08$\times$ & 1.04$\times$ & 1.03$\times$ & 1.03$\times$ \\
 &  & draft share & 38.8\% & 21.7\% & 21.0\% & 19.0\% & 17.8\% & --- \\
\bottomrule
\end{tabular}
\end{widetable}
}

\paragraph{Batch sweep.} Table~\ref{tab:gb10batch} uses the prompts,
estimator and settings of the B200 sweep (App.~\ref{app:scope}).

\FloatBarrier
\section{H100 reproduction}
\label{app:h100}

\paragraph{Setup.} Serving differs from B200 in the target (FP8 instead of
NVFP4), attention (FA3 for target and draft instead of TRT-LLM MHA and FA4) and
KV cache (BF16 instead of FP8), the fastest configuration we timed on the
full-width draft. Width comparisons are controlled within this setup. Absolute
acceptance is 8.6--9.7\% lower than on B200 in every variant, and because NVFP4
does not run on H100 we cannot separate the effect of FP8 serving from that of
attention numerics.

\paragraph{Results.} Half width stays at acceptance parity
(Table~\ref{tab:h100}), $+0.70\%$ against
$+0.84\%$ on B200, and quarter width costs 4.42\% against 5.44\%. Both narrow
drafts decode 1.11$\times$ faster, so the two widths tie, where on B200 halving
led, 1.09$\times$ against 1.06$\times$. In the profiled prompt
(Table~\ref{tab:h100l3}) the quarter draft runs 2.37$\times$ faster, against 1.75$\times$ on B200 (Table~\ref{tab:l3}), with its MLP and
attention projections at 3.37$\times$ and 2.67$\times$ against 2.63$\times$
and 1.80$\times$. Of the quarter-width $\tdraft$ saving the head supplies 35\%, the MLP
43\% and the attention projections 13\%.

\paragraph{Measurement.} $\TPOT$ is mean $\Titer$ over pooled $\AL$, over
about 190K--200K iterations per variant. Each SGLang decode log line is timed from
the previous line, so the two lines after each prefill include prefill time
and are dropped. Dropping one, two or four lines moves no speedup by more than
0.001$\times$.
Stage times come from one Nsight capture per variant on the B200 profiling
prompt (App.~\ref{app:profiling}), with a 2{,}048-token context and 1{,}024 new
tokens instead of 4{,}096 and 128. Nsight adds
1.1--1.6\% to $\Titer$, and profiled and unprofiled speedups agree within
0.5\%.

\sharedtablehere{\begin{widetable}
\centering
\footnotesize
\setlength{\tabcolsep}{4pt}
\caption{\textbf{H100 reproduction.} DFlash on Qwen3-8B with the same drafts and
880-prompt contract as Table~\ref{tab:headline} ($T{=}0$, $B{=}1$), on a
Qwen3-8B FP8 target with FA3 attention and a BF16 KV cache. Stage times come
from one profiled prompt. The last column repeats the B200 speedup.}
\label{tab:h100}
\begin{tabular}{lrrrrrrrr}
\toprule
Draft & $\AL$ & $\Delta\AL$ & $\Titer$ (ms) & $\TPOT$ (ms) & $\tdraft$ (ms) & $\tverify$ (ms) & $\TPOT$ speedup & B200 \\
\midrule
full 4{,}096           & 3.886 & ---       & 6.247 & 1.607 & 1.502 & 4.800 & 1.00$\times$ & 1.00$\times$ \\
half 2{,}048    & 3.914 & $+0.70\%$ & 5.682 & 1.452 & 0.930 ($-38.1\%$) & 4.807 & 1.11$\times$ & 1.09$\times$ \\
quarter 1{,}024 & 3.715 & $-4.42\%$ & 5.395 & 1.452 & 0.633 ($-57.8\%$) & 4.810 & 1.11$\times$ & 1.06$\times$ \\
\bottomrule
\end{tabular}
\end{widetable}
}
\sharedtablehere{\begin{widetable}
\centering
\footnotesize
\setlength{\tabcolsep}{4pt}
\caption{\textbf{Draft operations on H100.} DFlash on Qwen3-8B, $\mu$s per
iteration, with the ratio against full width in brackets. Columns as in
Table~\ref{tab:l3}. Only the ratios are comparable across GPUs.}
\label{tab:h100l3}
\begin{tabular}{lrrrrrr}
\toprule
Draft & attn.\ proj. & MLP & head & rest of body & outside & $\tdraft$ \\
\midrule
full 4{,}096           & 175.7 & 536.8 & 414.8 & 197.8 & 177.1 & 1502.3 \\
half 2{,}048    & 103.0 (1.71$\times$) & 285.3 (1.88$\times$) & 208.0 (1.99$\times$) & 201.5 (0.98$\times$) & 132.2 & 929.9 (1.62$\times$) \\
quarter 1{,}024 & 65.9 (2.67$\times$) & 159.4 (3.37$\times$) & 107.6 (3.86$\times$) & 193.6 (1.02$\times$) & 107.0 & 633.4 (2.37$\times$) \\
\bottomrule
\end{tabular}
\end{widetable}
}

\FloatBarrier
\section{Quantizing the draft}
\label{app:quant}

\paragraph{Setup.} We compare six variants of DFlash-Qwen, the full-, half-
and quarter-width BF16 drafts and the same three drafts quantized to NVFP4. We
quantize them after training, with no further training, using ModelOpt 0.47 (NVFP4 preset,
\texttt{max} algorithm). Every decoder-layer GEMM (QKV, O, gate/up and down) and
the draft's vocabulary projection run in NVFP4 with FP4 weights and FP4
activations (W4A4) on FlashInfer 0.6.17 FP4 GEMMs. NVFP4 uses 16-element blocks with
an FP8 scale per block and an FP32 scale per tensor. Weight scales come from
the weights, and fused QKV and gate/up shards share one per-tensor scale.
Activation block scales are computed at run time, and each GEMM's per-tensor
activation scale is fixed at the ModelOpt drafter default (amax 448), so no
calibration data is used. Calibrating it instead, to each GEMM's maximum over
64 SPEED-Bench throughput prompts disjoint from the panel, moved $\AL$ by at
most 0.4\% in a run that kept the vocabulary projection in BF16. At
full width we quantize a copy of the target's BF16 head to NVFP4, and the
target keeps its own BF16 head for verification. At half and quarter width we
fold first, computing $\Wfold = \Wlm\Wout$ from the BF16 head in FP32 and
rounding it to BF16, and then quantize $\Wfold$ once to NVFP4. Every width's
vocabulary projection is therefore quantized once, from BF16. \texttt{fc}, \texttt{input\_adapter}, the norms, attention and the draft KV
cache stay in BF16, and so does a copy of the context K/V weights, which
keeps the context pre-pass on its fused GEMM. The target and its verification
are unchanged, so decoding stays lossless. Each category's six variants ran in one
job on the same four GPUs of one node, each W4A4 variant at the same time as
its same-width BF16 variant, and all 11 categories ran twice, on different
nodes. The two passes agree within 0.002$\times$ on every speedup.
Table~\ref{tab:headline}'s DFlash-Qwen rows come from this run's BF16 variants.

\paragraph{Results.} Quantization gains less as the draft narrows
(Table~\ref{tab:quant}). W4A4 speeds up the iteration by 1.09$\times$ at full
width, 1.04$\times$ at half and 1.02$\times$ at quarter, and costs
2.3--3.4\% acceptance at each width, so it speeds up end-to-end decoding by
1.07$\times$, 1.02$\times$ and 0.99$\times$. Each FP4 projection runs a
separate activation-quantize kernel, about 48\,$\mu$s per step in all at both
full and half width, and FP4 GEMMs bottom out at about 5\,$\mu$s, so the small
GEMMs of a narrow draft gain little. Halving alone (1.09$\times$, $+0.84\%$
acceptance) beats quantizing the full-width draft (1.07$\times$,
$-2.36\%$). The two compose at half width, where the quantized half-width
draft is the fastest variant at 1.11$\times$, but not at quarter width. We leave
other drafter--target pairs, large batches and quantization-aware training to
future work.

\sharedtablehere{\begin{widetable}
\centering
\footnotesize
\setlength{\tabcolsep}{4pt}
\caption{\textbf{Quantizing the draft at each width.} DFlash on Qwen3-8B with the
880-prompt contract of Table~\ref{tab:headline} ($T{=}0$, $B{=}1$, one B200).
W4A4 is NVFP4 post-training quantization of the draft's decoder GEMMs and
vocabulary projection. All six variants of a category ran in one job on the same GPUs,
and the BF16 rows are those of Table~\ref{tab:headline}. $\Delta\AL$ and
the same-width speedups are against the BF16 draft of the same width, and the
last column is the $\TPOT$ speedup over the full-width BF16 draft. Bold marks
the best.}
\label{tab:quant}
\begin{tabular}{llrrrrrrr}
\toprule
Draft & Precision & $\AL$ & $\Delta\AL$ & $\Titer$ (ms) & $\TPOT$ (ms) & $\Titer$ speedup & $\TPOT$ speedup & vs full BF16 \\
\midrule
full 4{,}096    & BF16 & 4.298 & ---       & 3.585 & 0.834 & ---           & ---           & 1.00$\times$ \\
half 2{,}048    & BF16 & 4.334 & ---       & 3.313 & 0.764 & ---           & ---           & 1.09$\times$ \\
quarter 1{,}024 & BF16 & 4.064 & ---       & 3.197 & 0.787 & ---           & ---           & 1.06$\times$ \\
\midrule
full 4{,}096    & W4A4 & 4.196 & $-2.36\%$ & 3.284 & 0.782 & 1.09$\times$ & 1.07$\times$ & 1.07$\times$ \\
half 2{,}048    & W4A4 & 4.232 & $-2.33\%$ & 3.180 & 0.751 & 1.04$\times$ & 1.02$\times$ & \textbf{1.11$\times$} \\
quarter 1{,}024 & W4A4 & 3.925 & $-3.42\%$ & 3.130 & 0.797 & 1.02$\times$ & 0.99$\times$ & 1.05$\times$ \\
\bottomrule
\end{tabular}
\end{widetable}
}

\end{document}